\documentclass[a4paper, 11pt]{article}

\usepackage{fullpage}

\usepackage[english]{babel}
\usepackage[utf8]{inputenc}
\usepackage{amsmath, amsfonts, amssymb}
\usepackage{mathtools}
\usepackage{stmaryrd}
\usepackage{bm}
\usepackage{authblk}
\usepackage{cite}

\usepackage{hyperref}
\usepackage{booktabs}
\usepackage{graphicx}
\usepackage{subcaption}

\usepackage[linesnumbered,ruled,vlined]{algorithm2e}
\SetKwInput{KwInput}{input}
\SetAlFnt{\small}

\newcommand{\Real}{\mathbb{R}}

\newcommand{\E}{\mathbb{E}}

\makeatletter%
\@ifclassloaded{beamer}%
    {}%
    {

    }%
\makeatother%

\title{Variational Bayesian inference for CP tensor completion with side information}
\author[1]{Stanislav Budzinskiy}
\author[1]{Nikolai Zamarashkin}
\affil[1]{Marchuk Institute of Numerical Mathematics RAS}
\date{}

\begin{document}
\maketitle

\begin{abstract}
We propose a message passing algorithm, based on variational Bayesian inference, for low-rank tensor completion with automatic rank determination in the canonical polyadic format when additional side information (SI) is given. The SI comes in the form of low-dimensional subspaces the contain the fiber spans of the tensor (columns, rows, tubes, etc.). We validate the regularization properties induced by SI with extensive numerical experiments on synthetic and real-world data and present the results about tensor recovery and rank determination. The results show that the number of samples required for successful completion is significantly reduced in the presence of SI. We also discuss the origin of a bump in the phase transition curves that exists when the dimensionality of SI is comparable with that of the tensor.
\end{abstract}

\tableofcontents

\clearpage

\section{Introduction}
A big part of modern signal processing is based on working with multi-dimensional signals that exhibit some kind of hidden structure: this typically means sparsity in a certain basis/frame or low-rankedness. One of the principal ideas is to leverage the signal's structure to reconstruct it from a series of measurements that can be, on the one hand, scant and, on the other hand, corrupted by noise and outliers. The advent of computationally feasible algorithms for compressed sensing \cite{DonohoCompressed2006}, matrix completion \cite{CandesRechtExact2009a, CandesTaoPower2010}, and robust principal component analysis \cite{CandesEtAlRobust2011} marked the beginning of a new era in signal processing.

In certain applications, knowing that the signal is structured is not the only piece of information we have at our disposal. Some auxiliary information might be available as well, which can come in a variety of different forms: as an approximate solution to the problem \cite{MotaEtAlCompressed2017, XueEtAlSide2017}; as a graph describing the relations between the elements of the data \cite{RaoEtAlCollaborative2015, ShahidEtAlFast2016}; as an accompanying dataset that shares some of the latent factors \cite{SinghGordonRelational2008, ShanBanerjeeGeneralized2010a}; as low-dimensional subspaces containing row and column spans \cite{JainDhillonProvable2013, XuEtAlSpeedup2013, KimChoiScalable2014, ChiangEtAlMatrix2015, ChiangEtAlRobust2016}. By incorporating auxiliary information in an algorithm, we aim to help it recover signals using fewer measurements and in the presence of more severe corruption. The possible applications of auxiliary information include, for example, recommender systems \cite{MaEtAlSoRec2008}, video processing \cite{XieEtAlRobust2019}, and bioinformatics \cite{NatarajanDhillonInductive2014, ZakeriEtAlGene2018, GuvencPaltunEtAlImproving2021}. In what follows, we fix the name \textit{side information} for the auxiliary information in the form of low-dimensional subspaces.

Our focus is on tensor completion, or tensor factorization from incomplete data: given a multi-dimensional tensor, we want to reconstruct it from a small fraction of its elements. The problem is obviously ill-posed unless the tensor in question can be described with a small number of parameters and unless it is incoherent vis-\`a-vis the point measurements. For tensors, there exist several widely used low-parametric representations: the canonical polyadic (CP) decomposition (also known as CANDECOMP/PARAFAC), the Tucker decomposition, and the tensor train (TT) decomposition. See \cite{KoldaBaderTensor2009, OseledetsTensorTrain2011} for an introduction to these tensor representations and \cite{CichockiEtAlTensor2015, PapalexakisEtAlTensors2016, SidiropoulosEtAlTensor2017} for a review of their applications.

The properties of Tucker and TT decompositions are closely related to those of low-rank matrix decompositions, which stems from the fact that Tucker and TT ranks are simply the ranks of certain tensor unfoldings or flattenings. This explains why similar techniques have been applied to solve low-rank matrix, Tucker, and TT completion problems: convex optimization based on nuclear norm minimization \cite{CandesRechtExact2009a, CandesTaoPower2010, SignorettoEtAlNuclear2010, GandyEtAlTensor2011, BenguaEtAlEfficient2017}; non-convex optimization approaches that minimize the residual based on low-rank projections \cite{JainDhillonProvable2013, TannerWeiNormalized2013, LebedevaEtAlLowRank2021, RauhutEtAlTensor2015} or explicit factorization \cite{HaldarHernandoRankConstrained2009, WenEtAlSolving2012a, JainEtAlLowrank2013a, HastieEtAlMatrix2015, TannerWeiLow2016, GrasedyckEtAlVariants2015, GrasedyckKramerStable2019}, including their Riemannian variants \cite{KeshavanEtAlMatrix2010, BoumalAbsilRTRMC2011, VandereyckenLowRank2013, MishraEtAlFixedrank2014, KressnerEtAlLowrank2014, SteinlechnerRiemannian2016}.

CP decomposition inherits less from the matrix case; for instance, the best low-rank approximation problem becomes ill-posed \cite{deSilvaLimTensor2008}. Consequently, CP completion is carried out by updating the canonical factors in an alternating fashion \cite{TomasiBroPARAFAC2005, AcarEtAlScalable2011, YokotaEtAlSmooth2016}, and the canonical rank of the data is typically  determined by explicitly fitting several CP models of different ranks.

An important group of methods we have not mentioned yet are probabilistic ones that employ the Bayesian inference machinery to solve matrix and tensor completion problems. The general approach consists in setting up a probabilistic model for the data and estimating its parameters and hyperparameters. In the seminal paper \cite{SalakhutdinovMnihProbabilistic2007}, a maximum a posteriori estimate was found for the latent factors of a low-rank matrix and it was noted that a fully Bayesian approach would give higher predictive accuracy. Computing the exact posterior distribution of the parameters conditioned on the data is typically impossible, so one has to adopt an approximate Bayesian inference strategy. The two dominant ones are Markov chain Monte Carlo (MCMC) sampling \cite{NealProbabilistic1993} and variational inference \cite{BealVariational2003, WinnBishopVariational2005}, and both of them have been applied to matrix completion problems; see \cite{SalakhutdinovMnihBayesian2008, GilbertWellsTuning2019} and  \cite{LakshminarayananEtAlRobust2011, BabacanEtAlSparse2012, YangEtAlFast2018}, respectively.

Tensor completion has also been addressed with probabilistic and Bayesian methods for Tucker \cite{ChuGhahramaniProbabilistic2009, ZhaoEtAlBayesian2015a}, TT \cite{XuEtAlLearning2020}, and CP \cite{RaiEtAlScalable2014, ZhaoEtAlBayesian2015, ZhaoEtAlBayesian2016, ChengEtAlProbabilistic2020} decompositions. The Bayesian framework is especially promising for CP factorization since inference of a probabilistic model based on Gaussian-Gamma priors achieves automatic rank determination \cite{ZhaoEtAlBayesian2015}. These priors were first introduced for matrix completion \cite{BabacanEtAlSparse2012} and have recently been extended to generalized hyperbolic priors \cite{ChengEtAlProbabilistic2020}; other choices are possible too \cite{AlquierEtAlBayesian2014, GilbertWellsTuning2019}.

In this article, we deal with what lies at the intersection of the topics mentioned above: Bayesian CP completion in the presence of side information. In fact, this amounts to finding a CP factorization of a Tucker core when the corresponding Tucker factors are known. Completion and factorization problems with different kinds of auxiliary information have been studied in the literature for CP \cite{AcarEtAlAllatonce2011, NaritaEtAlTensor2012, YokotaEtAlLinked2012, BazerqueEtAlRank2013a,WuEtAlFused2019, GuanEtAlAlternating2020, IoannidisEtAlCoupled2021, YangEtAlMTC2021} and Tucker \cite{ErmisEtAlLink2015} decompositions. Side information, as we defined it, received less attention: it was used for TT completion \cite{BudzinskiyZamarashkinNote2020, BudzinskiyZamarashkinTensor2021}, Tucker completion \cite{LongEtAlTrainable2022}, and we have not seen such papers for the CP decomposition. Similar formulations appear in kernelized matrix completion \cite{GonenEtAlKernelized2013, LiEtAlKernelized2021}.

Our goal is to develop a tensor completion method for the CP decomposition using variational Bayesian inference and incorporating side information. To our knowledge, this is the first time side information is incorporated in CP tensor completion.

The paper begins with the matrix case: in Section \ref{section:matrix} we formulate a probabilistic model, introduce the variational Bayesian approach, and present formulas for an iterative message passing inference algorithm. In Section \ref{section:tensor} we turn to tensors and describe the corresponding CP completion approach. Section \ref{section:numerical} is devoted to numerical experiments that give insight into the regularization properties of side information. All the details of the derivations can be found in the Appendix.

\section{Notation}
Matrices are denoted by uppercase letters (e.g. $X$), and we use bold uppercase letters (e.g. $\bm{X}$) for tensors. For a $d$-dimensional tensor $\bm{X}$ of size $n_1 \times \ldots \times n_d$, we write its element in position $(i_1, \ldots, i_d) \in [n_1] \times \ldots \times [n_d]$ as $x_{i_1 \ldots i_d}$, where $[n] = \{ 1, \ldots, n \}$. If a tensor $\bm{X}$ admits a rank-$r$ canonical decomposition with factor matrices $\{ A_l \}$ of sizes $n_l \times r$, we can write it as
\begin{equation*}
    x_{i_1 \ldots i_d} = \sum_{j = 1}^{r} a_{1, i_1 j} \ldots a_{d, i_d j}, \quad \bm{X} = \llbracket A_1, \ldots, A_d \rrbracket.
\end{equation*}
The trace of a matrix is denoted by $\text{Tr}$. For a matrix $X$ of size $mk \times nk$, we denote its $j$-th diagonal $m \times n$ block by $\text{block}_j X$. For a subset of indices $\Omega \subseteq [n_1] \times \ldots \times [n_d]$ product and sum over $\Omega$ will be denoted by $\prod_\Omega$ and $\sum_\Omega$, respectively.

The Kronecker and Hadamard (elementwise) products of matrices are denoted by $\otimes$ and $\odot$, respectively. For $d$ matrices $\{ X_l \}$ of size $m \times n$, we define their multi-linear product as
\begin{equation*}
    \langle X_1, \ldots, X_d \rangle = \sum_{i = 1}^{m} \sum_{j = 1}^{n} x_{1, ij} \ldots x_{d, ij}.
\end{equation*}
This is a generalization of the Frobenius inner product
\begin{equation*}
    \langle X_1, X_2 \rangle_F = \text{Tr}\{ X_1^T X_2 \}, \quad \| X \|_F = \sqrt{\langle X, X \rangle_F}.
\end{equation*}

The Gaussian, Gamma, and Student's distributions are written as $\mathcal{N}$, $\mathcal{G}$, and $\text{St}$, respectively. The expectation of a random variable is denoted by $\E$.

\section{Matrix completion with side information}\label{section:matrix}
\subsection{Probabilistic model: priors}
Let $X \in \Real^{n_1 \times n_2}$ be a rank-$r$ matrix and $Y = X + E$ be the same matrix but corrupted by i.i.d. random Gaussian noise
\begin{align*}
    &p(E) = \prod_{i_1=1}^{n_1} \prod_{i_2=1}^{n_2} \mathcal{N}(\varepsilon_{i_1 i_2} | 0, \tau^{-1})
\end{align*}
with zero mean and precision $\tau > 0$. In matrix completion we have access only to those entries of $Y$ that belong to a given collection of indices $\Omega \subseteq [n_1] \times [n_2]$:
\begin{equation*}
    Y_{\Omega} = \mathcal{P}_{\Omega}(X + E) \in \Real^{n_1 \times n_2}.
\end{equation*}
The operator $\mathcal{P}_{\Omega}: \Real^{n_1 \times n_2} \to \Real^{n_1 \times n_2}$ keeps intact the elements of a matrix that lie in $\Omega$ and sets to zero all the remaining ones.

In the setting of completion with side information, we are additionally given a pair of subspaces spanned by the columns of full-rank matrices 
\begin{equation*}
    G = 
    \begin{bmatrix}
    g_1^T \\
    g_2^T \\
    \vdots \\
    g_{n_1}^T
    \end{bmatrix} \in \Real^{n_1 \times m_1},
    \quad 
    H = 
    \begin{bmatrix}
    h_1^T \\
    h_2^T \\
    \vdots \\
    h_{n_2}^T
    \end{bmatrix} \in \Real^{n_2 \times m_2}
\end{equation*}
with $m_1 < n_1$ and $m_2 < n_2$ and it is known that these subspaces contain the column and row spaces of $X$, respectively:
\begin{equation*}
    \mathrm{col} X \subseteq \mathrm{col} G, \quad \mathrm{col} X^T \subseteq \mathrm{col} H.
\end{equation*}
We can express the a priori information about $X$ in a compact form by writing it as a product $X = G U (HV)^T$, where 
\[
    U = 
    \begin{bmatrix}
    u_1^T \\
    u_2^T \\
    \vdots \\
    u_{m_1}^T
    \end{bmatrix} \in \Real^{m_1 \times k},
    \quad 
    V = 
    \begin{bmatrix}
    v_1^T \\
    v_2^T \\
    \vdots \\
    v_{m_2}^T
    \end{bmatrix} \in \Real^{m_2 \times k}
\]
are the unknown factors that we need to recover; $k \geq r$ serves as a possibly overestimated prediction of the rank. As a result, the conditional distribution of $Y_{\Omega}$ becomes
\[
    p(Y_{\Omega} | U, V, \tau) = \prod_{\Omega} \mathcal{N}(y_{i_1 i_2} | g_{i_1}^T U V^T h_{i_2}, \tau^{-1}).
\]
Note that this model is different from \cite{KimChoiScalable2014}, where the interaction between $G$ and $U$, and between $H$ and $V$, is subject to Gaussian noise.

Following \cite{BabacanEtAlSparse2012}, assume that the rows of $U$ and $V$ are i.i.d. random Gaussian vectors 
\begin{align*}
    &p(U | \Lambda) = \prod_{i_1 = 1}^{n_1} \mathcal{N}( u_{i_1} | 0, \Lambda^{-1} ), \\
    &p(V | \Lambda) = \prod_{i_2 = 1}^{n_2} \mathcal{N}( v_{i_2} | 0, \Lambda^{-1} ),
\end{align*}
with zero mean and precision matrix $\Lambda = \text{diag}(\lambda_j) \in \Real^{k \times k}$. The idea behind this prior is twofold. First, it enforces the columns of $U$ and $V$ to be balanced in terms of their norms. Second, if some of the $\lambda_j$ are large, the corresponding columns of $U$ and $V$ have little impact and can be removed to reduce the rank prediction $k$. We fix a Gamma hyperprior, parametrized with shape and rate parameters, for the precision matrix:
\begin{equation*}
    p(\Lambda) = \prod_{j = 1}^{k} \mathcal{G}(\lambda_j | a_j, b_j).
\end{equation*}
Gaussian random variables with Gamma-distributed precision are ubiquitous in Bayesian inference since they form an exponentially conjugate pair (see Appendix). Finally, we choose a Gamma hyperprior for the noise precision as well,
\begin{equation*}
    p(\tau) = \mathcal{G}(\tau | a_0, b_0),
\end{equation*}
which gives us the following joint distribution:
\begin{equation*}
    p(Y_{\Omega}, U, V, \Lambda, \tau) = p(Y_{\Omega} | U, V, \tau) p(U | \Lambda) p(V | \Lambda) p(\Lambda) p(\tau).
\end{equation*}

\subsection{Variational Bayesian inference}
We now turn to the posterior distribution of the model parameters conditioned on the observed data:
\begin{equation*}
    p(U, V, \Lambda, \tau | Y_{\Omega}) = \frac{p(Y_{\Omega}, U, V, \Lambda, \tau)}{\int p(Y_{\Omega}, U, V, \Lambda, \tau) dU dV d\Lambda d\tau}.
\end{equation*}
Exact Bayesian inference consists in evaluating $p(U, V, \Lambda, \tau | Y_{\Omega})$, which, however, is not an option, since the evidence of the model, the denominator of the right hand side, is an intractable integral. So approximate inference methods need to be used that seek a distribution $q(U, V, \Lambda, \tau)$ such that
\begin{equation*}
    q(U, V, \Lambda, \tau) \approx p(U, V, \Lambda, \tau | Y_{\Omega}).
\end{equation*}
To this end, we use variational Bayesian inference.

Denote by $\Theta = (\theta_1, \theta_2, \theta_3, \theta_4) = (U, V, \Lambda, \tau)$ the model parameters. We will look for a factorized variational distribution
\begin{equation*}
    q(\Theta) = q(U) q(V) q(\Lambda) q(\tau)
\end{equation*}
that minimizes the Kullback--Leibler divergence 
\begin{equation*}
    \mathcal{KL}\big[q(\Theta) ~||~ p(\Theta | Y_{\Omega})\big] = \int q(\Theta) \log \frac{q(\Theta)}{p(\Theta | Y_{\Omega})} d\Theta.
\end{equation*}
Simple algebra shows that this is equivalent to minimizing 
\begin{equation*}
    \int q(\Theta) \log \frac{q(\Theta)}{p(Y_{\Omega}, \Theta)} d\Theta \to \min_{q(\Theta)}.
\end{equation*}
If we now substitute the factorized form of $q(\Theta)$ and attempt to minimize over $q(\theta_i)$ with all the remaining $q(\theta_j)$, $j \neq i$, fixed, we will see that the minimum is attained at the optimal distribution $q^*(\theta_i)$, whose logarithm is
\begin{equation*}
    \log q^*(\theta_i) = \E_{{\sim} \theta_i} \{ \log p(Y_{\Omega}, \Theta) \} + \mathrm{const}.
\end{equation*}
The notation $\E_{{\sim} \theta_i}$ stands for the expectation with respect to the distribution $\prod_{j \neq i} q(\theta_j)$, and the constant contains the logarithm of the normalization factor. This gives rise to a message-passing algorithm for variational Bayesian inference, where we update the distributions iteratively for each $\theta_i$ one by one and which converges to a local minimum \cite{WinnBishopVariational2005}.

What is particularly appealing in the variational Bayesian inference approach for our model is that the optimal distributions $q^*(\theta_i)$ are of the same form as the corresponding prior distributions, owing to the exponential conjugacy. Below, we present the explicit formulas for the $q^*(\theta_i)$. Find their derivations in the Appendix. 

\subsection{Optimal posterior distributions}
\subsubsection{Factor matrices $U$ and $V$}
Denote by $\overline{u} \in \Real^{m_1 k}$ and $\overline{v} \in \Real^{m_2 k}$ the vectorizations of $U$ and $V$ obtained by stacking their columns as
\begin{gather*}
    \overline{u} = 
    \begin{bmatrix}
    u_{1 1} & \ldots & u_{m_1 1} & u_{1 2} & \ldots & u_{m_1 k}
    \end{bmatrix}^T, \\
    \overline{v} = 
    \begin{bmatrix}
    v_{1 1} & \ldots & v_{m_2 1} & v_{1 2} & \ldots & v_{m_2 k}
    \end{bmatrix}^T,
\end{gather*}
and let $\E$ with no subscript be the expectation with respect to the product of those $q(\theta_i)$, for which $\theta_i$ is included in the expression that is averaged. 

The optimal posterior distribution for the factor matrix $U$ is a Gaussian distribution of its vectorization
\begin{equation*}
    q^*(U) = \mathcal{N}(\overline{u} | \mu_{\overline{u}}, A_{{\overline{u}}})
\end{equation*}
with covariance
\begin{gather*}
    A_{\overline{u}} = \bigg[ \E\{\Lambda\} \otimes I_{m_1} + \E\{\tau\} B_{\overline{u}} \bigg]^{-1}, \\
    B_{\overline{u}} =\sum_{\Omega} (I_k \otimes h_{i_2}^T) \E\{ \overline{v} \overline{v}^T \} (I_k \otimes h_{i_2}) \otimes g_{i_1} g_{i_1}^T
\end{gather*}
and mean
\begin{equation*}
    \mu_{\overline{u}} = \E\{\tau\} A_{\overline{u}} \sum_{\Omega} y_{i_1 i_2} (I_k \otimes h_{i_2}^T) \E\{ \overline{v} \} \otimes g_{i_1}.
\end{equation*}
If the side information is trivial, that is $G$ and $H$ are square identity matrices, we recognize block-diagonal structure in $A_{\overline{u}}$, and the rows of $U$ remain independent in the posterior just as in the prior (cf. \cite{BabacanEtAlSparse2012}). The non-trivial side information, on the contrary, intertwines the rows; if, however, the side information is incorporated as in \cite{KimChoiScalable2014}, the rows stay independent.

Similarly
\begin{equation*}
    q^*(V) = \mathcal{N}(\overline{v} | \mu_{\overline{v}}, A_{{\overline{v}}})
\end{equation*}
with
\begin{gather*}
    A_{\overline{v}} = \bigg[ \E\{\Lambda\} \otimes I_{m_2} + \E\{\tau\} B_{\overline{v}} \bigg]^{-1}, \\
    B_{\overline{v}} = \sum_{\Omega} (I_k \otimes g_{i_1}^T) \E\{ \overline{u} \overline{u}^T \} (I_k \otimes g_{i_1}) \otimes h_{i_2} h_{i_2}^T
\end{gather*}
and
\begin{equation*}
    \mu_{\overline{v}} = \E\{\tau\} A_{\overline{v}} \sum_{\Omega} y_{i_1 i_2} (I_k \otimes g_{i_1}^T) \E\{ \overline{u} \} \otimes h_{i_2}.
\end{equation*}

\subsubsection{Precision matrix $\Lambda$}
The optimal posterior distribution for the precision matrix $\Lambda$ of the factors $U$ and $V$ is again a product of Gamma distributions
\begin{equation*}
    q^*(\Lambda) = \prod_{j = 1}^{k} \mathcal{G}(\lambda_j | c_j, d_j)
\end{equation*}
but with shifted shape and rate parameters for $j = 1, \ldots, k$:
\begin{equation*}
    c_j = a_j + \frac{m_1 + m_2}{2}, \, d_j = b_j + \frac{1}{2} \E\{ U^T U \}_{jj} + \frac{1}{2} \E\{ V^T V \}_{jj}.
\end{equation*}
For instance, the parameters of the posterior distribution depend on the side information only via the dimensions of the subspaces and not the subspaces themselves.

\subsubsection{Noise precision $\tau$}
The optimal posterior distribution for the noise precision $\tau$ follows a Gamma distribution
\begin{equation*}
    q^*(\tau) = \mathcal{G}(\tau | c_0, d_0)
\end{equation*}
with
\begin{equation*}
    c_0 = a_0 + \frac{|\Omega|}{2}, \quad d_0 = b_0 + \frac{1}{2} \E\Big\{ \| Y_{\Omega} - \mathcal{P}_{\Omega} (G U V^T H^T) \|_F^2 \Big\}.
\end{equation*}
The rate parameter is updated by the averaged squared Frobenius norm of the residual.

\subsection{Message passing updates}
To turn the expressions for the optimal distributions $q^*(\theta_i)$ into an iterative algorithm, it remains to explicitly compute the expectations. An iteration of the message passing algorithm then proceeds as follows. At first, we update the covariance for the matrix factor $U$:
\begin{equation}
\label{m_vmp:eq:update_Au}
\begin{split}
    &B_{\overline{u}} \gets \sum_{\Omega} (I_k \otimes h_{i_2}^T) (\mu_{\overline{v}} \mu_{\overline{v}}^T + A_{\overline{v}}) (I_k \otimes h_{i_2}) \otimes g_{i_1} g_{i_1}^T, \\
    &A_{\overline{u}} \gets \bigg[ \text{diag}\left(\frac{c_j}{d_j}\right) \otimes I_{m_1} + \frac{c_0}{d_0} B_{\overline{u}} \bigg]^{-1}.
\end{split}
\end{equation}
This new $A_{\overline{u}}$ is used to calculate the corresponding mean:
\begin{equation}
\label{m_vmp:eq:update_mu}
    \mu_{\overline{u}} \gets \frac{c_0}{d_0} A_{\overline{u}} \sum_{\Omega} y_{i_1 i_2} (I_k \otimes h_{i_2}^T) \mu_{\overline{v}} \otimes g_{i_1}.
\end{equation}
Then, in a similar fashion, we evaluate the new covariance
\begin{equation}
\label{m_vmp:eq:update_Av}
\begin{split}
    &B_{\overline{v}} \gets \sum_{\Omega} (I_k \otimes g_{i_1}^T) (\mu_{\overline{u}} \mu_{\overline{u}}^T + A_{\overline{u}}) (I_k \otimes g_{i_1}) \otimes h_{i_2} h_{i_2}^T, \\
    &A_{\overline{v}} \gets \bigg[ \text{diag}\left(\frac{c_j}{d_j}\right) \otimes I_{m_2} + \frac{c_0}{d_0} B_{\overline{v}} \bigg]^{-1}
\end{split}
\end{equation}
and mean
\begin{equation}
\label{m_vmp:eq:update_mv}
    \mu_{\overline{v}} \gets \frac{c_0}{d_0} A_{\overline{v}} \sum_{\Omega} y_{i_1 i_2} (I_k \otimes g_{i_1}^T) \mu_{\overline{u}} \otimes h_{i_2}
\end{equation}
for the second factor matrix $V$. The shape parameters of $q^*(\Lambda)$ are updated once and for all as
\begin{equation}
\label{m_vmp:eq:update_cj}
    c_j \leftarrow a_j + \frac{m_1 + m_2}{2}, \quad j = 1, \ldots, k,
\end{equation}
while the rate parameters are recomputed on each iteration of the message passing procedure:
\begin{equation}
\label{m_vmp:eq:update_dj}
\begin{split}
    d_j \leftarrow b_j &+ \frac{1}{2} \text{Tr}\left\{\text{block}_j (\mu_{\overline{u}} \mu_{\overline{u}}^T + A_{\overline{u}}) \right\} \\
    &+ \frac{1}{2} \text{Tr}\left\{\text{block}_j (\mu_{\overline{v}} \mu_{\overline{v}}^T + A_{\overline{v}}) \right\}, \quad j = 1,\ldots,k,
\end{split}
\end{equation}
where $\text{block}_j A_{\overline{u}} \in \Real^{m_1 \times m_1}$ is the $j$-th diagonal block of $A_{\overline{u}}$. Likewise, the shape hyperparameter for noise precision is set only once
\begin{equation}
\label{m_vmp:eq:update_c0}
    c_0 \leftarrow a_0 + \frac{|\Omega|}{2},
\end{equation}
but the corresponding rate parameter is evaluated every time:
\begin{equation}
\label{m_vmp:eq:update_d0}
\begin{split}
    d_0 \leftarrow b_0 &+ \frac{1}{2} \sum_{\Omega} \Big[ (y_{i_1 i_2} - g_{i_1}^T M_{\overline{u}} M_{\overline{v}}^T h_{i_2})^2 \\
    &+ g_{i_1}^T M_{\overline{u}} (I_k \otimes h_{i_2}^T) A_{\overline{v}} (I_k \otimes h_{i_2}) M_{\overline{u}}^T g_{i_1} \\
    &+ h_{i_2}^T M_{\overline{v}} (I_k \otimes g_{i_1}^T) A_{\overline{u}} (I_k \otimes g_{i_1}) M_{\overline{v}}^T h_{i_2} \\
    &+ \text{Tr}\left\{ A_{\overline{v}} (I_k \otimes h_{i_2} g_{i_1}^T) A_{\overline{u}} (I_k \otimes g_{i_1} h_{i_2}^T) \right\} \Big],
\end{split}
\end{equation}
where we denote by $M_{\overline{u}} \in \Real^{m_1 \times k}$ and $M_{\overline{v}} \in \Real^{m_2 \times k}$ the matricizations of $\mu_{\overline{u}}$ and $\mu_{\overline{v}}$, respectively.

Having computed the new posterior distribution $q(\Theta)$, we can reduce the rank prediction $k$ by removing those columns of $U$ and $V$, for which the mean $c_j / d_j$ of $q(\lambda_j)$ is large.

We can also approximately compute the distribution of the unknown elements of $Y$. Namely, for $(i_1, i_2) \not\in \Omega$ the distribution of $y_{i_1 i_2}$ conditioned on $Y_{\Omega}$ is close to a Student's $t$-distribution
\begin{equation*}
    p(y_{i_1 i_2} | Y_{\Omega}) \approx \text{St}\left(y_{i_1 i_2} | g_{i_1}^T M_{\overline{u}} M_{\overline{v}}^T h_{i_2}, \xi, 2 c_0 \right)
\end{equation*}
with
\begin{align*}
    \xi = \Big[ \frac{d_0}{c_0} &+ h_{i_2}^T M_{\overline{v}} (I_k \otimes g_{i_1}^T) A_{\overline{u}} (I_k \otimes g_{i_1} ) M_{\overline{v}} h_{i_2} \\
    &+ g_{i_1}^T M_{\overline{u}} (I_k \otimes h_{i_2}^T) A_{\overline{v}} (I_k \otimes h_{i_2}) M_{\overline{u}}^T h_{i_2} \Big]^{-1}.
\end{align*}
So its mean is $g_{i_1}^T M_{\overline{u}} M_{\overline{v}}^T h_{i_2}$ and its variance is $\dfrac{c_0}{\xi(c_0 - 1)}$.
\section{Tensor completion with side information}\label{section:tensor}
\subsection{Probabilistic model: priors}
Let $\bm{X} \in \Real^{n_1 \times \ldots \times n_d}$ be a $d$-dimensional tensor with canonical rank equal to $r$. Assume that for every dimension we have full-rank side information matrices
\begin{equation*}
    G_l = 
    \begin{bmatrix}
    g_{l, 1}^T \\
    g_{l, 2}^T \\
    \vdots \\
    g_{l, n_l}^T \\
    \end{bmatrix} \in \Real^{n_l \times m_l}, \quad l = 1, \ldots, d.
\end{equation*}
If $k \geq r$ is our prediction of the canonical rank, the latent factor matrices of the canonical decomposition are
\begin{equation*}
    U_l = 
    \begin{bmatrix}
    u_{l, 1}^T \\
    u_{l, 2}^T \\
    \vdots \\
    u_{l, n_l}^T \\
    \end{bmatrix} \in \Real^{m_l \times k}, \quad l = 1, \ldots, d.
\end{equation*}
This allows us to write every element of the tensor $\bm{X}$ as a multi-linear product of length-$k$ vectors
\begin{align*}
    x_{i_1 \ldots i_d} &= \langle U_1^T g_{1, i_1}, \ldots, U_d^T g_{d, i_d} \rangle = \sum_{j = 1}^{k} (U_1^T g_{1, i_1})_j \ldots (U_d^T g_{d, i_d})_j.
\end{align*}
As a shorthand for this, we will write $\bm{X} = \llbracket G_1 U_1, \ldots, G_d U_d \rrbracket$. Just as in the two-dimensional matrix case, we have access only to a subset of entries that are additionally corrupted by noise. Denote by $\Omega \subseteq [n_1] \times \ldots \times [n_d]$ the corresponding collection of multi-indices and let $\bm{E} \in \Real^{n_1 \times \ldots \times n_d}$ be a tensor with i.i.d random Gaussian components
\begin{equation*}
    p(\bm{E}) = \prod_{i_1=1}^{n_1} \ldots \prod_{i_d=1}^{n_d} \mathcal{N}(\varepsilon_{i_1 \ldots i_d} | 0, \tau^{-1}).
\end{equation*}
Then what we know is a sample $\bm{Y}_{\Omega} = \mathcal{P}_{\Omega}(\bm{X} + \bm{E})$ that is distributed according to
\begin{align*}
    p(\bm{Y}_{\Omega} &| U_1, \ldots, U_d, \tau) = \prod_{\Omega} \mathcal{N}(y_{i_1 \ldots i_d} | \langle U_1^T g_{1, i_1}, \ldots, U_d^T g_{d, i_d} \rangle, \tau^{-1}).
\end{align*}
We choose the same priors as before for the factor matrices
\begin{equation*}
    p(U_l | \Lambda) = \prod_{i_l = 1}^{n_l} \mathcal{N}(u_{l, i_l} | 0, \Lambda^{-1}), \quad l = 1, \ldots, d,
\end{equation*}
and the hyperparameters
\begin{equation*}
    p(\Lambda) = \prod_{j = 1}^{k} \mathcal{G}(\lambda_j | a_j, b_j), \quad p(\tau) = \mathcal{G}(\tau | a_0, b_0).
\end{equation*}

\subsection{Optimal posterior distributions}
The variational inference framework with a factorized distribution $q(\Theta)$ can be applied in the tensor case too. It provides optimal posterior distributions $q^*(\theta_i)$ that, due to exponential conjugacy, are of the same form as the corresponding priors. 

\subsubsection{Canonical factors $U_l$}
The optimal posterior distribution for each canonical factor $U_l$ is a Gaussian distribution of its vectorization
\begin{equation*}
    q^*(U_l) = \mathcal{N}(\overline{u}_l | \mu_{l}, A_{l}).
\end{equation*}
To present the formulas for the mean and covariance, it is convenient to express multi-linear products in terms of the Hadamard product. For every $l = 1, \ldots, d$ we have
\begin{equation*}
    \langle U_1^T g_{1, i_1}, \ldots, U_d^T g_{d, i_d} \rangle = (U_l^T g_{l, i_l})^T \bigodot_{s \neq l} U_s^T g_{s, i_s}.
\end{equation*}
Then the covariance matrix $A_l \in \Real^{m_l k \times m_l k}$ can be written as
\begin{gather*}
    A_l = \Big[ \E\{\Lambda\} \otimes I_{m_l} + \E\{\tau\} B_{l} \Big]^{-1}, \\
    B_{l} = \sum_{\Omega} \left( \bigodot_{s \neq l} (I_k \otimes g_{s, i_s}^T) \E\{\overline{u}_s \overline{u}_s^T\} (I_k \otimes g_{s, i_s}) \right) \otimes g_{l, i_l} g_{l, i_l}^T,
\end{gather*}
and the mean is
\begin{equation*}
    \mu_l = \E\{\tau\} A_l \sum_{\Omega} y_{i_1 \ldots i_d} \left( \bigodot_{s \neq l} \E\{U_s^T\} g_{s, i_s} \right) \otimes  g_{l, i_l}.
\end{equation*}

\subsubsection{Precision matrix $\Lambda$}
As previously, the components of the diagonal precision matrix $\Lambda$ are independent Gamma random variables in the posterior distribution:
\begin{equation*}
    q^*(\Lambda) = \prod_{j = 1}^{k} \mathcal{G}(\lambda_j | c_j, d_j).
\end{equation*}
The formulas for their parameters are simple multi-dimensional extensions of what we saw in the matrix case, that is for $j = 1, \ldots, k$ we have 
\begin{equation*}
    c_j = a_j + \frac{1}{2} \sum_{l = 1}^d m_l, \quad d_j = b_j + \frac{1}{2} \sum_{l = 1}^d \E\{ U_l^T U_l \}_{jj}.
\end{equation*}

\subsubsection{Noise precision $\tau$}
As for the precision parameter $\tau$ of the noise, it follows a posterior Gamma distribution
\begin{equation*}
    q^*(\tau) = \mathcal{G}(\tau | c_0, d_0)
\end{equation*}
with shape and rate given by
\begin{equation*}
    c_0 = a_0 + \frac{|\Omega|}{2}, \, d_0 = b_0 + \frac{1}{2} \E\Big\{ \| \bm{Y}_{\Omega} - \mathcal{P}_{\Omega} \llbracket G_1 U_1, \ldots, G_d U_d \rrbracket \|_F^2 \Big\}.
\end{equation*}

\subsection{Message passing updates}
It is now straightforward to turn the formulas for the optimal posterior distributions into a message passing algorithm. On each iteration, we will start by updating the posteriors of the canonical factors one by one: the covariance
\begin{align}
\label{t_vmp:eq:update_Al}
    B_l \leftarrow \sum_{\Omega} \Bigg( \bigodot_{s \neq l} &\bigg[ (I_k \otimes g_{s, i_s}^T) A_s (I_k \otimes g_{s, i_s}) + M_s^T g_{s, i_s} g_{s, i_s}^T M_s \bigg] \Bigg) \otimes g_{l, i_l} g_{l, i_l}^T,
\end{align}
\begin{equation*}
    A_l \leftarrow \left[ \text{diag}\left(\frac{c_j}{d_j}\right) \otimes I_{m_l} + \frac{c_0}{d_0} B_l \right]^{-1},
\end{equation*}
followed by the mean
\begin{equation}
\label{t_vmp:eq:update_ml}
    \mu_l \leftarrow \frac{c_0}{d_0} A_l \sum_{\Omega} y_{i_1 \ldots i_d} \left( \bigodot_{s \neq l} M_s^T g_{s, i_s} \right) \otimes g_{l, i_l},
\end{equation}
where $M_s$ stands for the matricization of $\mu_s$. We then update the rate parameters for the precision matrix
\begin{equation}
\label{t_vmp:eq:update_dj}
    d_j \leftarrow b_j + \frac{1}{2} \sum_{l = 1}^d \text{Tr}\left\{\text{block}_j (\mu_l \mu_l^T + A_l) \right\}
\end{equation}
for $j = 1, \ldots, k$, and for the noise precision
\begin{align}
\label{t_vmp:eq:update_d0}
    d_0 &\leftarrow b_0 + \frac{1}{2} \sum_{\Omega} \Big[ y_{i_1 \ldots i_d}^2 - 2 y_{i_1 \ldots i_d} \langle M_1^T g_{1, i_1}, \ldots, M_d^T g_{d, i_d} \rangle \notag\\ 
    + \Big\langle &(I_k \otimes g_{1, i_1}^T) A_1 (I_k \otimes g_{1, i_1}) + M_s^T g_{s, i_s} g_{s, i_s}^T M_s, \ldots, \notag\\
    &(I_k \otimes g_{d, i_d}^T) A_d (I_k \otimes g_{d, i_d}) + M_d^T g_{d, i_d} g_{d, i_d}^T M_d\Big\rangle \Big].
\end{align}
The corresponding shape parameters have fixed values as shown above.

The missing entries of $\bm{Y}$ follow, approximately, a Student's $t$-distribution
\begin{equation*}
    p(y_{i_1 \ldots i_d} | \bm{Y}_{\Omega}) \approx \text{St}\left(y_{i_1 \ldots i_d} | \langle M_1^T g_{1, i_1}, \ldots, M_d^T g_{d, i_d} \rangle, \xi, 2 c_0 \right),
\end{equation*}
where $\xi^{-1}$ equals
\begin{align*}
    \frac{d_0}{c_0} +\sum_{l = 1}^d \bigg( \bigodot_{s \neq l} M_s^T g_{s, i_s}\bigg)^T (I_k \otimes g_{l, i_l}^T) A_{l} (I_k \otimes g_{l, i_l} ) \bigg( \bigodot_{s \neq l} M_s^T g_{s, i_s}\bigg) ,
\end{align*}
giving mean $\langle M_1^T g_{1, i_1}, \ldots, M_d^T g_{d, i_d} \rangle$ and variance $\dfrac{c_0}{\xi(c_0 - 1)}$.
\section{Numerical experiments}\label{section:numerical}
\subsection{Computational complexity}
One step of our variational message passing algorithm consists in updating the parameters and hyperparameters of the posterior distributions, i.e. computing Eqs. \eqref{m_vmp:eq:update_Au}--\eqref{m_vmp:eq:update_d0} for matrix completion and Eqs. \eqref{t_vmp:eq:update_Al}--\eqref{t_vmp:eq:update_d0} for tensor completion.

Let $m = \max\{m_l\}$. Forming the matrix that needs to be inverted in \eqref{m_vmp:eq:update_Au} requires $\mathcal{O}(|\Omega| m^2 k^2)$ operations. We then compute its Cholesky factorization, which takes $\mathcal{O}(m^3 k^3)$ operations, and invert in $\mathcal{O}(m^2 k^2)$. In \eqref{m_vmp:eq:update_mu}, we can reuse the temporary vectors from \eqref{m_vmp:eq:update_Au} to compute the sum with $\mathcal{O}(|\Omega| mk)$ operations and multiply it by $A_{\overline{u}}$ using $\mathcal{O}(m^2 k^2)$ operations, thanks to the Cholesky factorization. This results in  $\mathcal{O}(|\Omega| m^2 k^2 + m^3 k^3)$ operations for updating the posterior distribution of one factor matrix. The cost of \eqref{m_vmp:eq:update_dj} is clearly dominated by that of factor matrix updates, and to compute the new $d_0$ via \eqref{m_vmp:eq:update_d0} requires $\mathcal{O}(|\Omega| m^2 k^2)$ operations. So, in total, one iteration of variational message passing for matrix completion with side information takes $\mathcal{O}(|\Omega| m^2 k^2 + m^3 k^3)$ operations (cf. \cite{KimChoiScalable2014}).

In the tensor case, the asymptotic computational complexity is also defined by the updates for canonical factors. Each of them now costs $\mathcal{O}(d |\Omega|m^2 k^2 + m^3 k^3)$ operations, and added up together they give the total complexity of $\mathcal{O}(d^2 |\Omega| m^2 k^2 + d m^3 k^3)$ operations.

This complexity can potentially be reduced, if we use a different strategy for matrix inversion, such as conjugate gradient or Newton iterations, or sparse Cholesky factorizations \cite{SchaferEtAlSparse2021}.

\subsection{Initialization}
In all our experiments we initialize the posterior rate and shape ($\{ c_j, d_j \}$ and $c_0, d_0$) of the precision parameters with $10^{-6}$. We draw the posterior means $\{ \mu_l \}$ of the factor matrices from the standard Gaussian distribution $\mathcal{N}(0, I)$ and set their covariances $\{ A_l \}$ to identities $I$; it is also possible to initialize using SVD \cite{ZhaoEtAlBayesian2015}.

\subsection{Synthetic data}
To extensively test the regularization properties of side information, we generate random data. For fixed values of $d$, $n$, $m$, and $r$, we draw $d$ latent factors $\{ U_l \}$ of sizes $m \times r$ with i.i.d elements from $\mathcal{N}(0, 1)$. The side information matrices $\{ G_l \}$ of sizes $n \times m$ are generated in the same way. If no side information is used, we generate $\{ U_l \}$ of sizes $n \times r$. We denote an instance of such tensors by $\bm{X}(d, n, r, m)$ and $\bm{X}(d, n, r)$, respectively. The sampling set $\Omega$ is chosen uniformly at random with replacement from $[n]^d$. The noise is Gaussian with zero mean and variance that gives the prescribed signal-to-noise ratio (SNR).

\subsubsection{Performance of completion}
\begin{figure*}[t]
\begin{subfigure}[b]{0.5\linewidth}
\centering
	\includegraphics[width=0.7\linewidth]{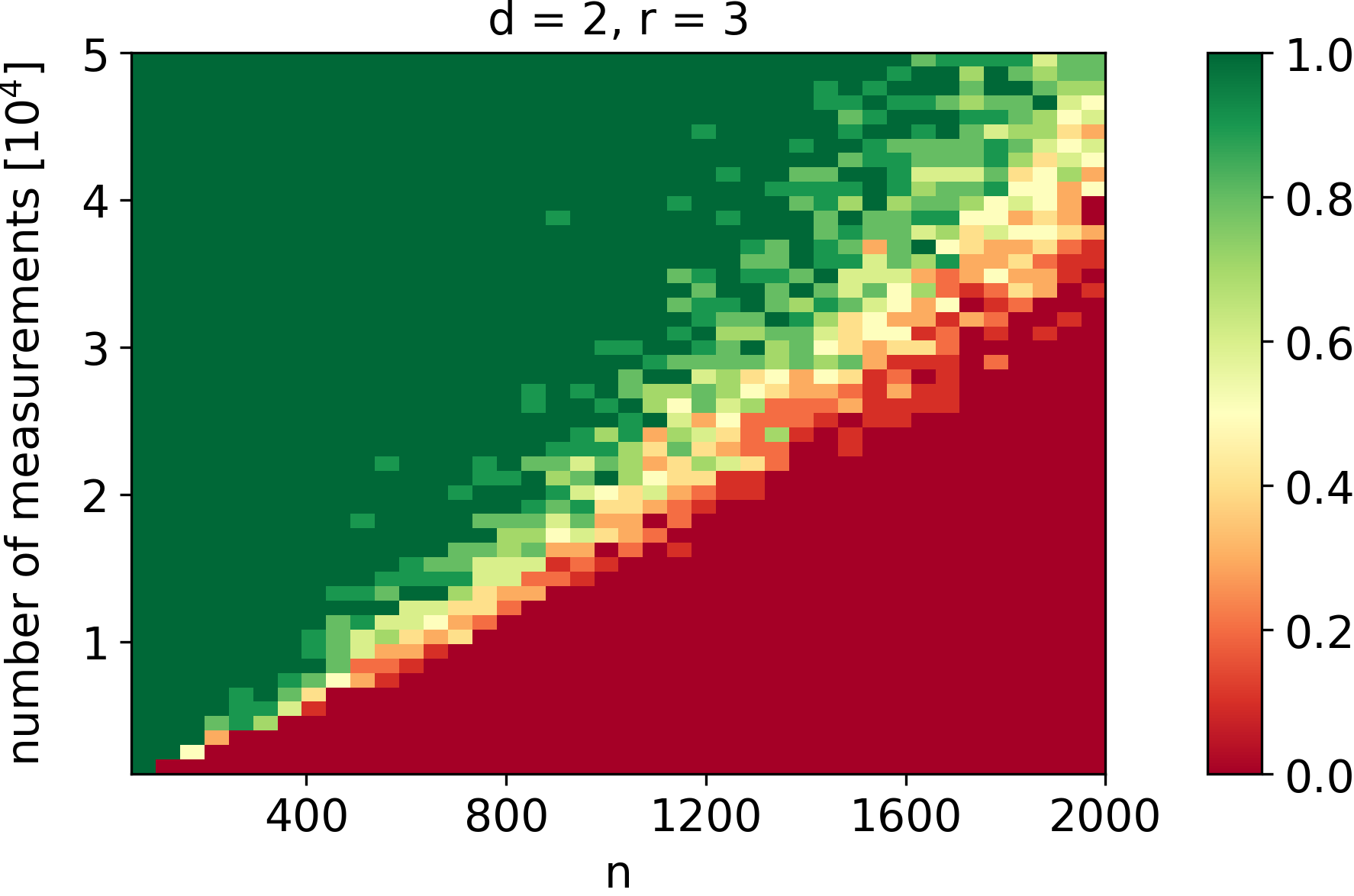}
	\caption{}
\end{subfigure}\hfill%
\begin{subfigure}[b]{0.5\linewidth}
\centering
	\includegraphics[width=0.7\linewidth]{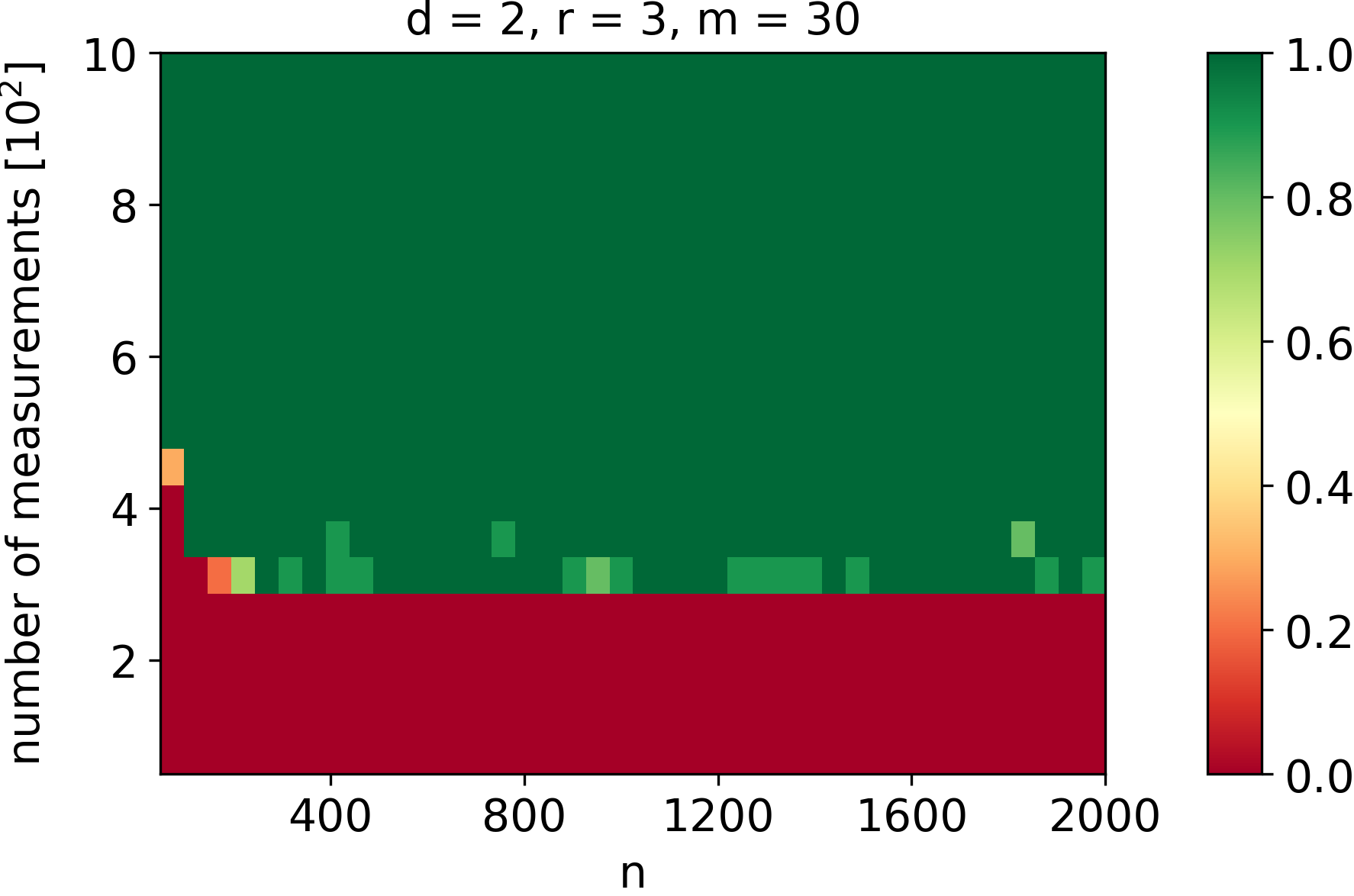}
	\caption{}
\end{subfigure}
\caption{Phase plots for noiseless matrix completion with rank $r = 3$ and perfect rank prediction $k = 3$: FBCP~(a) and FBCP-SI with side information size $m = 30$~(b).}
\label{num:fig:no_si_d2}
\end{figure*}
\begin{figure*}[t]
\begin{subfigure}[b]{0.5\linewidth}
\centering
	\includegraphics[width=0.7\linewidth]{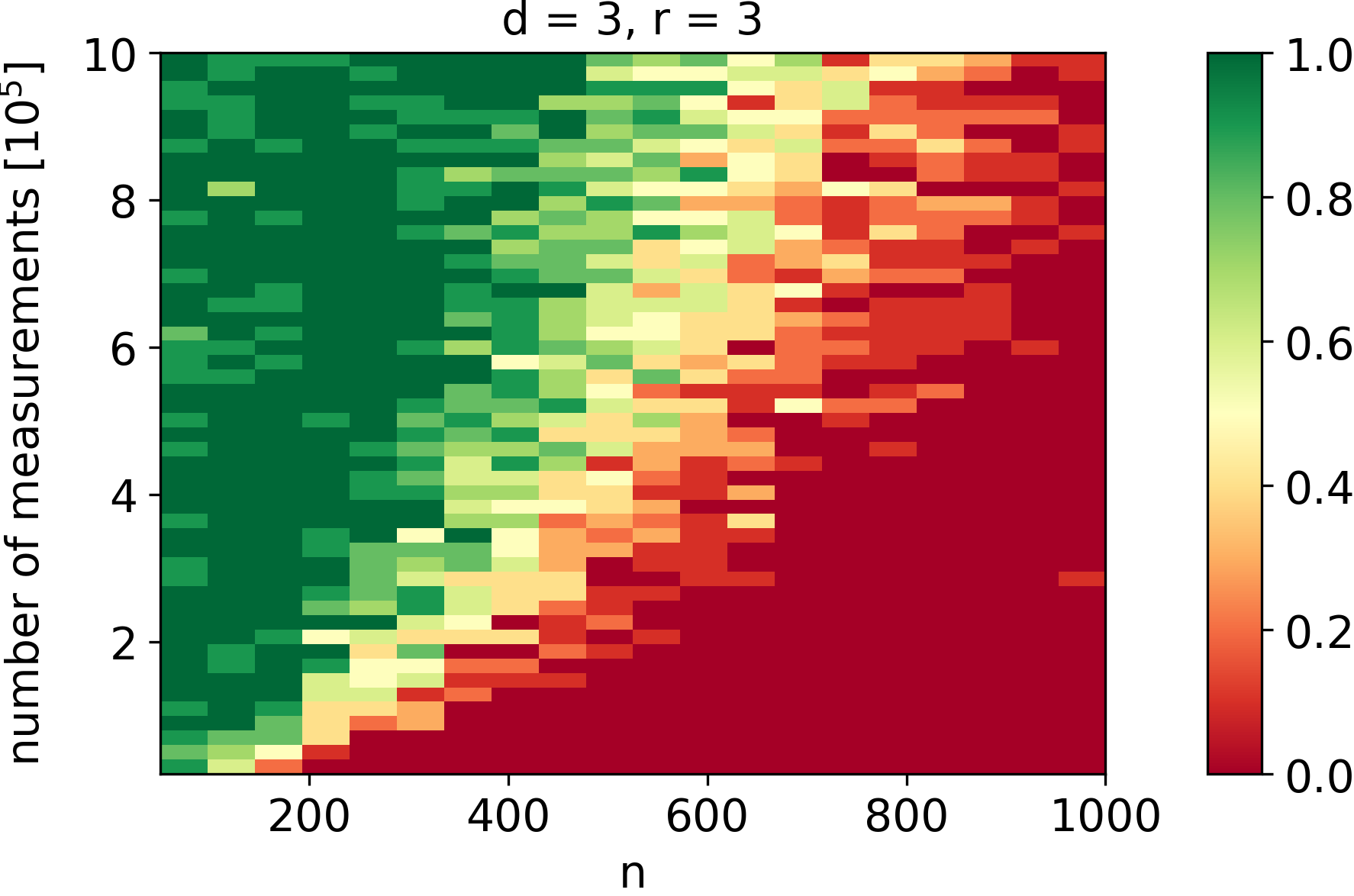}
	\caption{}
\end{subfigure}\hfill%
\begin{subfigure}[b]{0.5\linewidth}
\centering
	\includegraphics[width=0.7\linewidth]{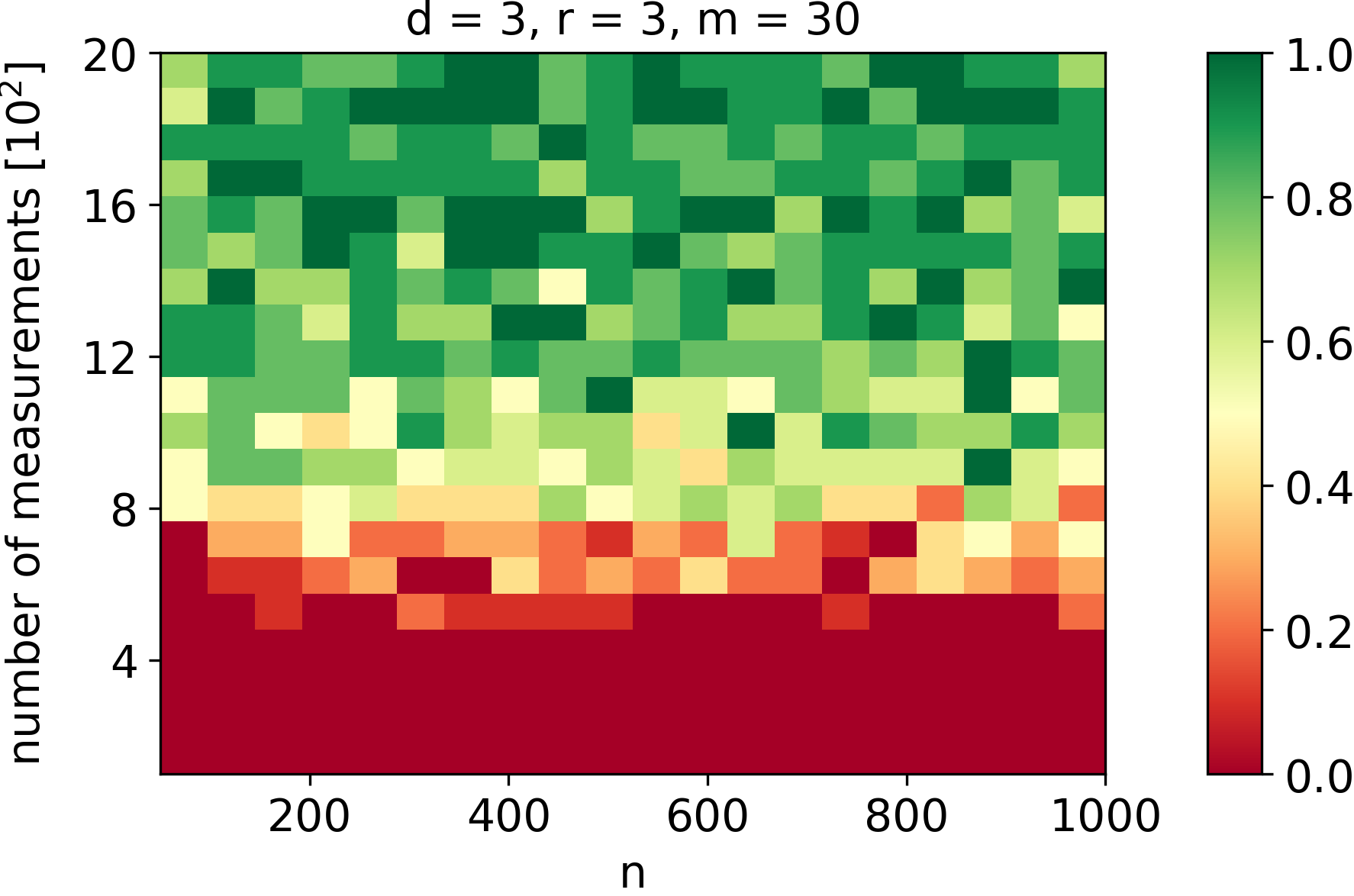}
	\caption{}
\end{subfigure}
\caption{Phase plots for noiseless $3$-dimensional CP completion with rank $r = 3$ and perfect rank prediction $k = 3$: FBCP~(a) and FBCP-SI with side information size $m = 30$~(b).}
\label{num:fig:no_si_d3}
\end{figure*}

In the first series of experiments, which are presented in Figs.~\ref{num:fig:no_si_d2} and \ref{num:fig:no_si_d3}, we study how side information can reduce the number of elements $|\Omega|$ needed to recover a low-rank CP tensor with FBCP \cite{ZhaoEtAlBayesian2015}. For every set of parameters, we generate $N_{trial}$ random problems
\begin{equation*}
    \Big\{ \bm{X}^{(t)}(d, n, r), \Omega^{(t)}, \Omega^{(t)}_{test} \Big\}_{t = 1}^{N_{trial}}
\end{equation*}
without noise and run $N_{iter}$ iterations of FBCP with $N_{ic}$ different random initial conditions with perfect rank prediction $k=r$. We say that a problem is successfully solved if RMSE on a test sampling set $\Omega^{(t)}_{test}$ of size $|\Omega|$ is smaller than $10^{-6}$:
\begin{equation*}
    \frac{\Big\| \mathcal{P}_{\Omega^{(t)}_{test}} \Big(\bm{X}^{(t)} - \bm{X}^{(t)}_{N_{iter}}\Big) \Big\|_F}{\Big\| \mathcal{P}_{\Omega^{(t)}_{test}} \bm{X}^{(t)} \Big\|_F} < 10^{-6}.
\end{equation*}
We then plot the frequency of successes among the $N_{trial} N_{ic}$ runs in the $(n, |\Omega|)$-plane. In the same way, we test our method, FBCP with side information (FBCP-SI), by generating
\begin{equation*}
    \Big\{ \bm{X}^{(t)}(d, n, r, m), \Omega^{(t)}, \Omega^{(t)}_{test} \Big\}_{t = 1}^{N_{trial}}.   
\end{equation*}

\begin{figure*}[t]
\begin{subfigure}[b]{0.5\linewidth}
\centering
	\includegraphics[width=0.7\linewidth]{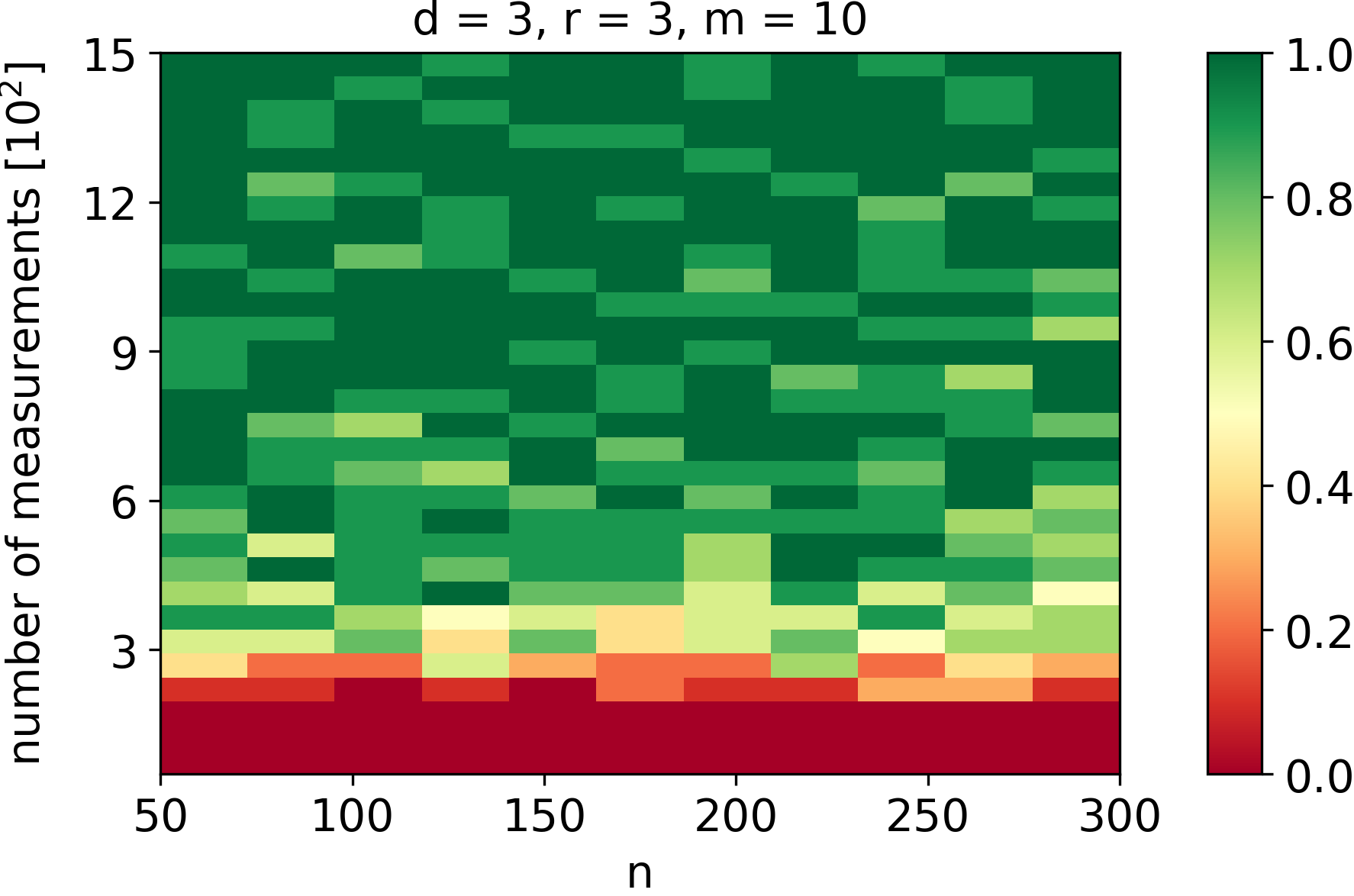}
	\caption{}
\end{subfigure}\hfill%
\begin{subfigure}[b]{0.5\linewidth}
\centering
	\includegraphics[width=0.7\linewidth]{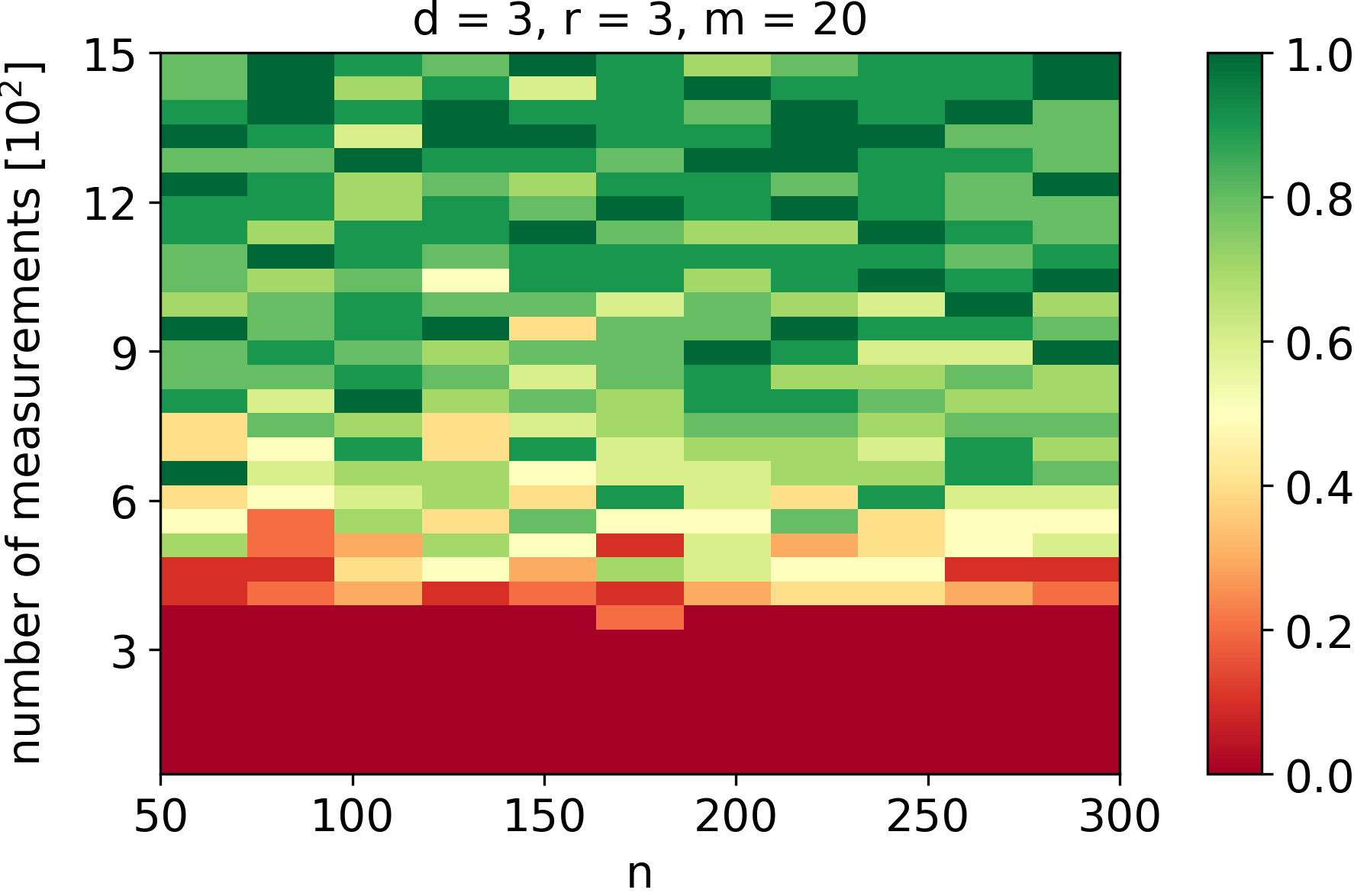}
	\caption{}
\end{subfigure}
\begin{subfigure}[b]{0.5\linewidth}
\centering
	\includegraphics[width=0.7\linewidth]{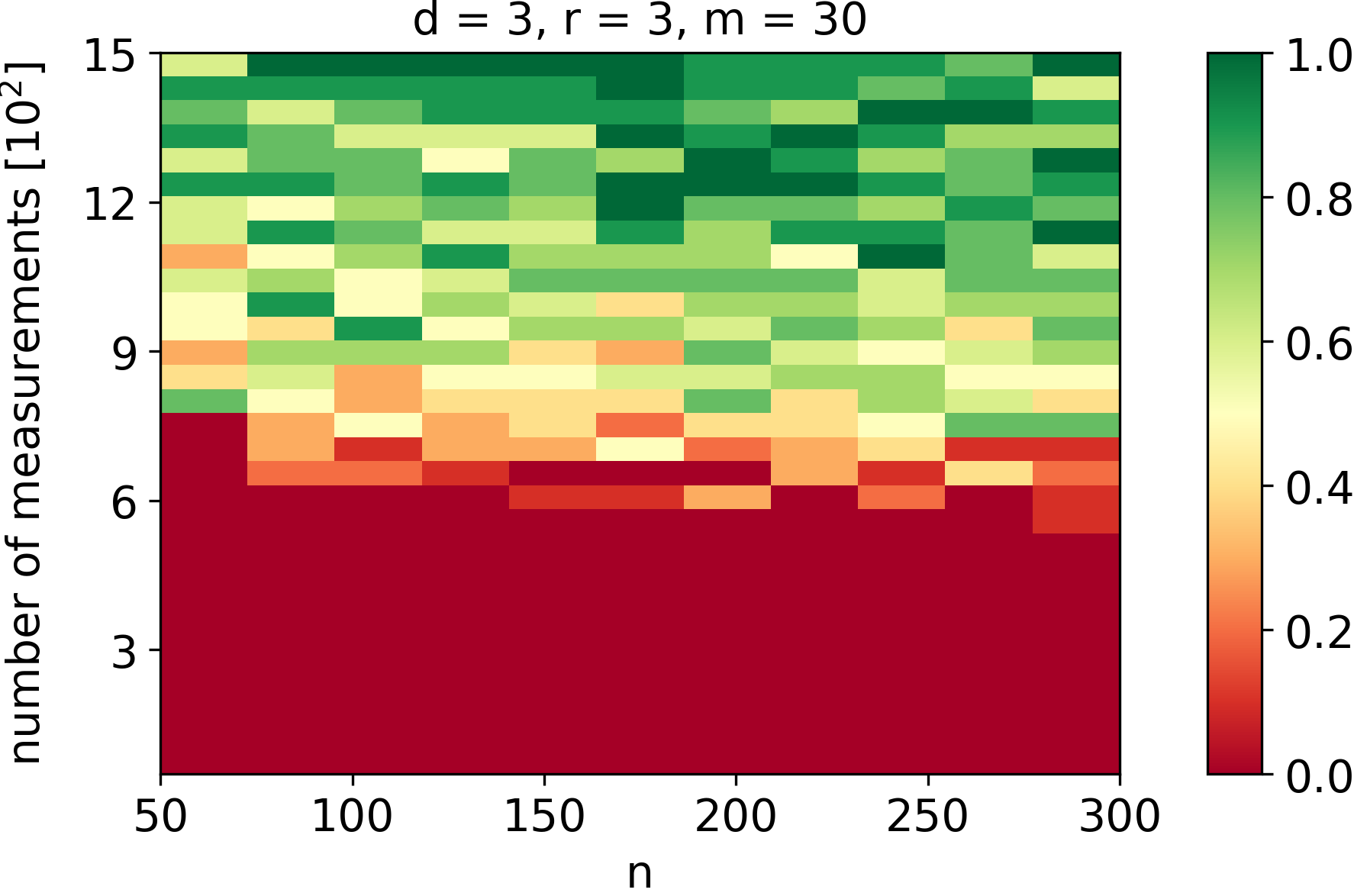}
	\caption{}
\end{subfigure}\hfill%
\begin{subfigure}[b]{0.5\linewidth}
\centering
	\includegraphics[width=0.7\linewidth]{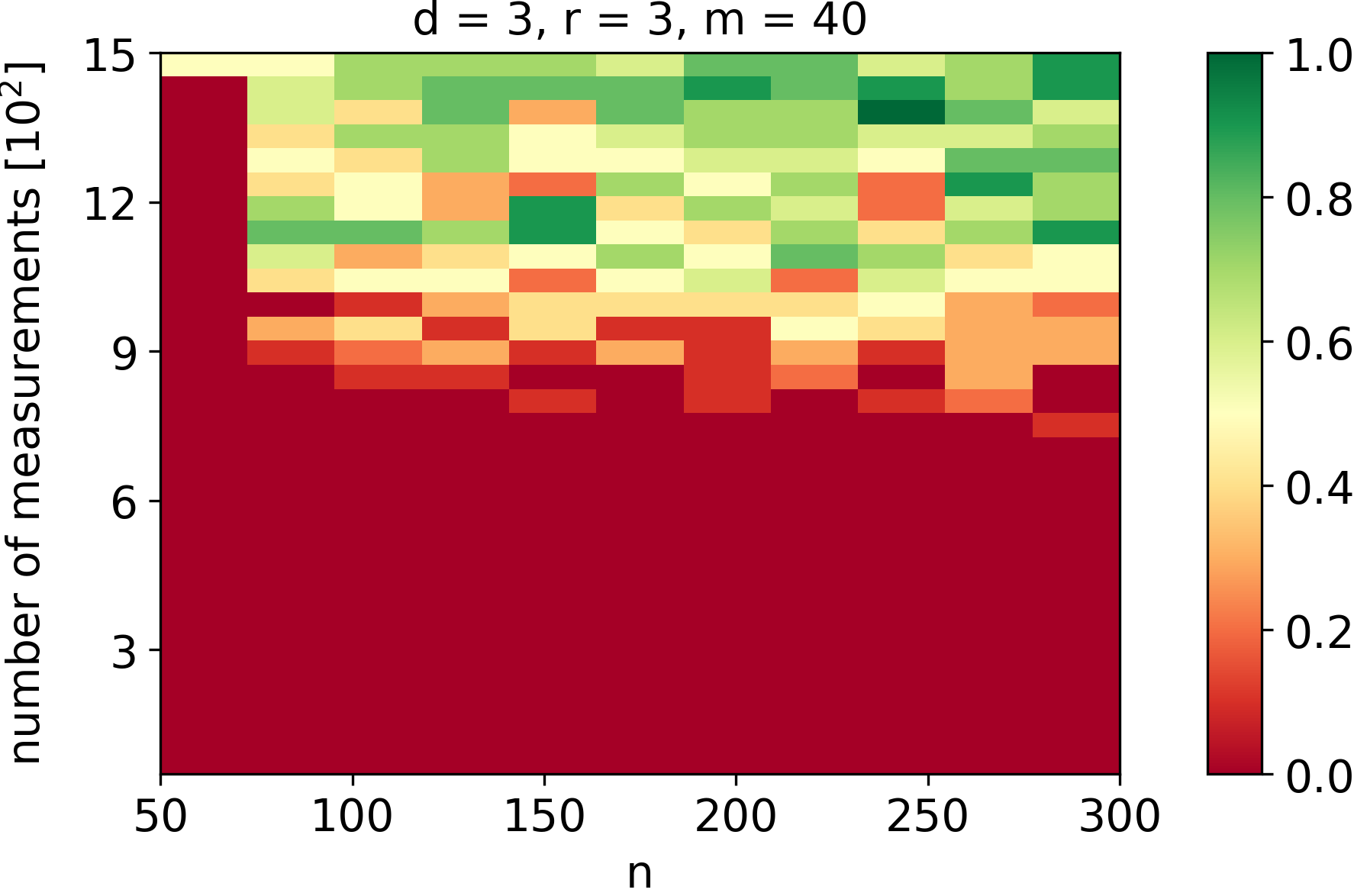}
	\caption{}
\end{subfigure}
\caption{Phase plots for noiseless $3$-dimensional CP completion with rank $r = 3$, perfect rank prediction $k = 3$, and different sizes of side information: $m = 10$~(a), $m = 20$~(b), $m = 30$~(c), and $m = 40$~(d).}
\label{num:fig:si_d3_bump}
\end{figure*}
\begin{figure*}[t]
\begin{subfigure}[b]{0.5\linewidth}
\centering
	\includegraphics[width=0.7\linewidth]{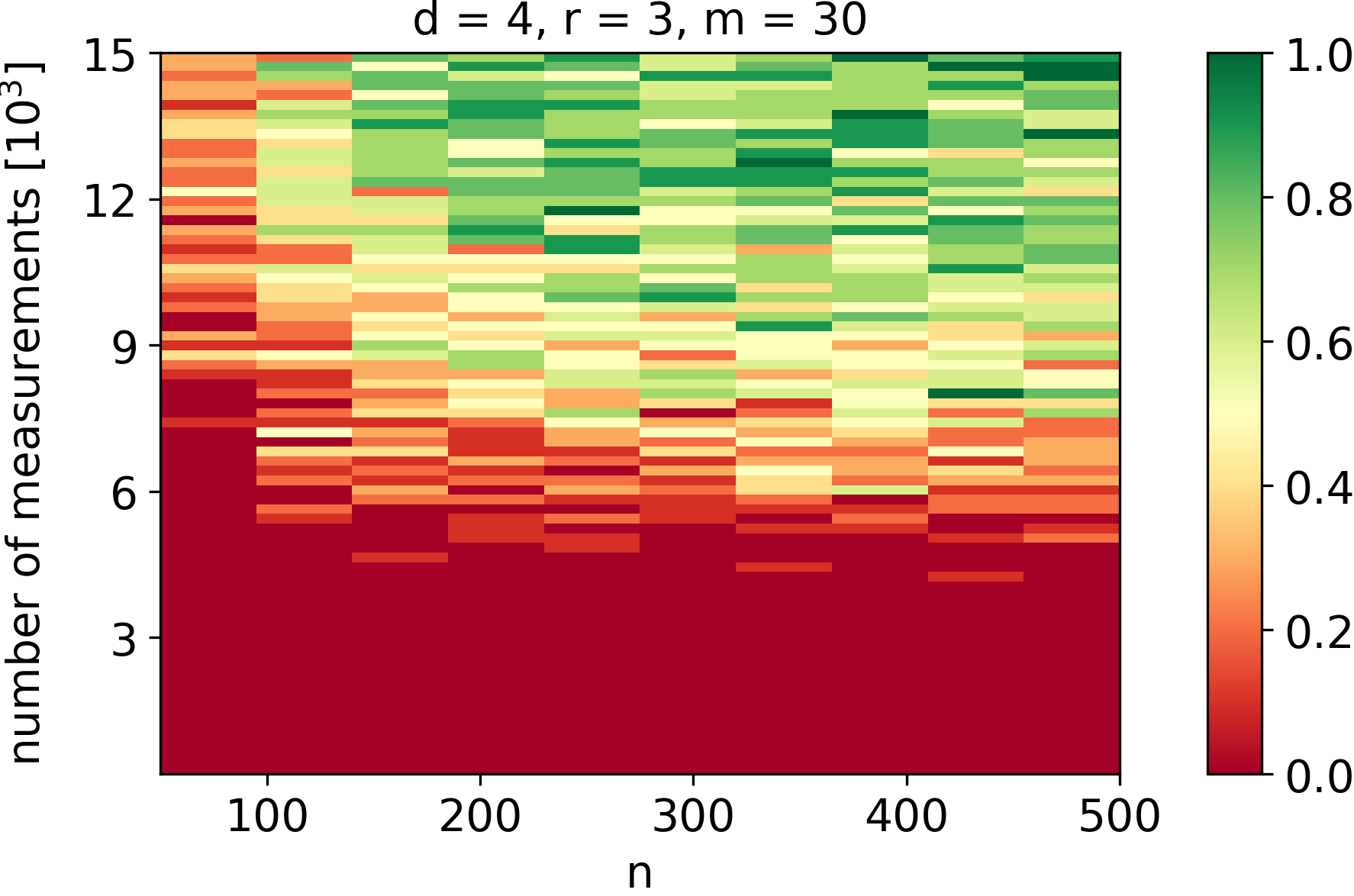}
	\caption{}
\end{subfigure}\hfill%
\begin{subfigure}[b]{0.5\linewidth}
\centering
	\includegraphics[width=0.7\linewidth]{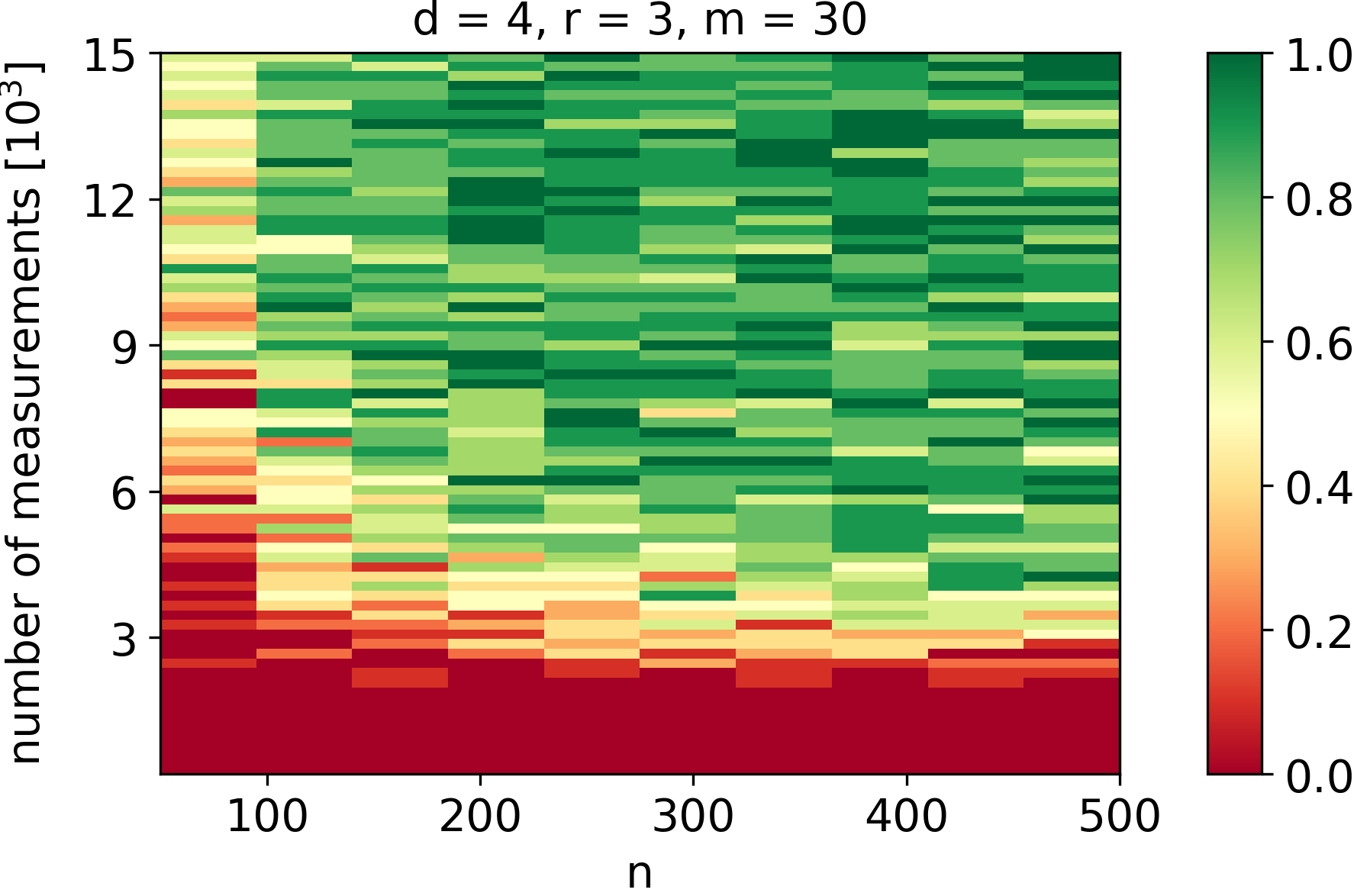}
	\caption{}
\end{subfigure}
\begin{subfigure}[b]{0.5\linewidth}
\centering
	\includegraphics[width=0.7\linewidth]{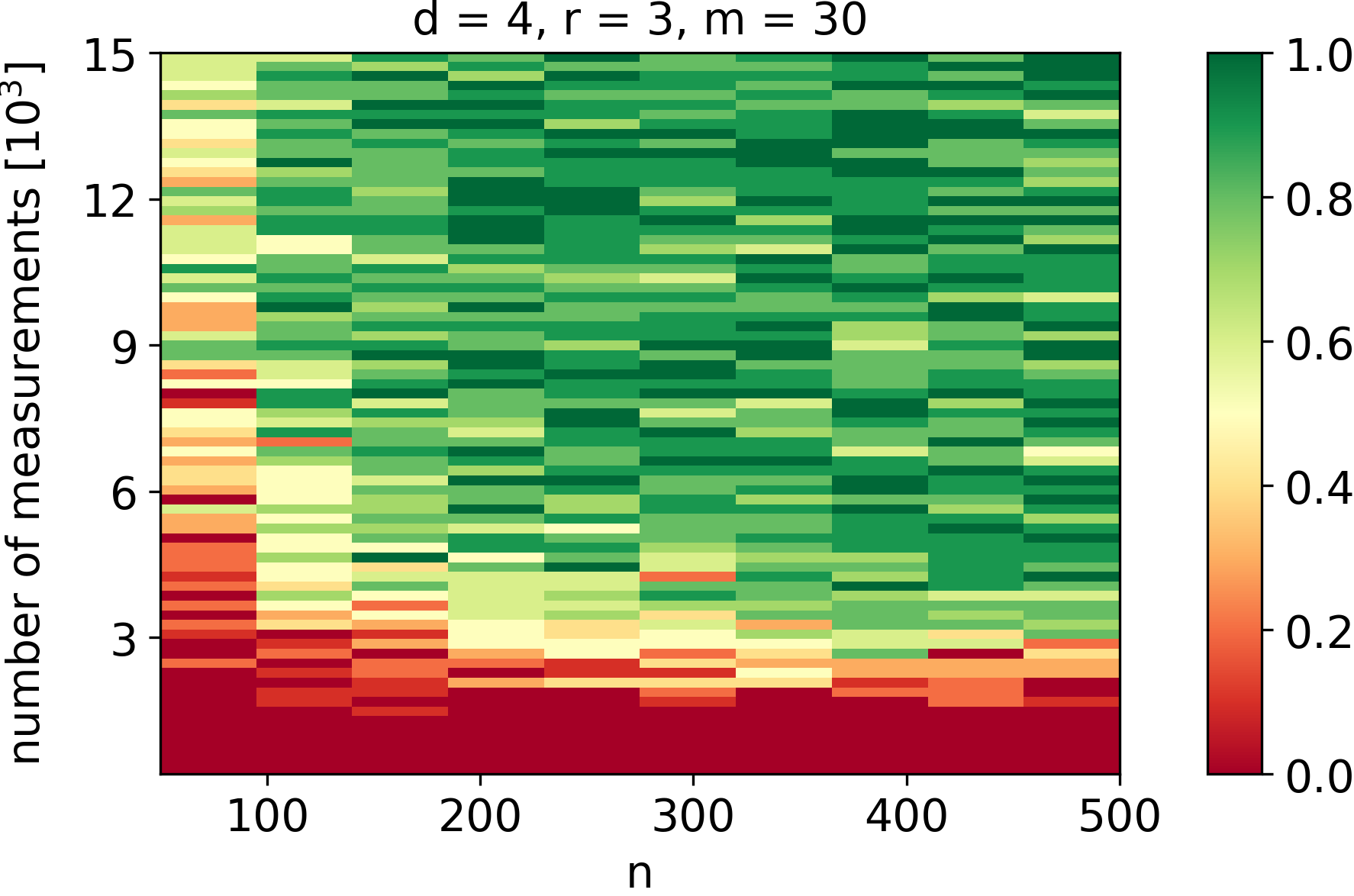}
	\caption{}
\end{subfigure}\hfill%
\begin{subfigure}[b]{0.5\linewidth}
\centering
	\includegraphics[width=0.7\linewidth]{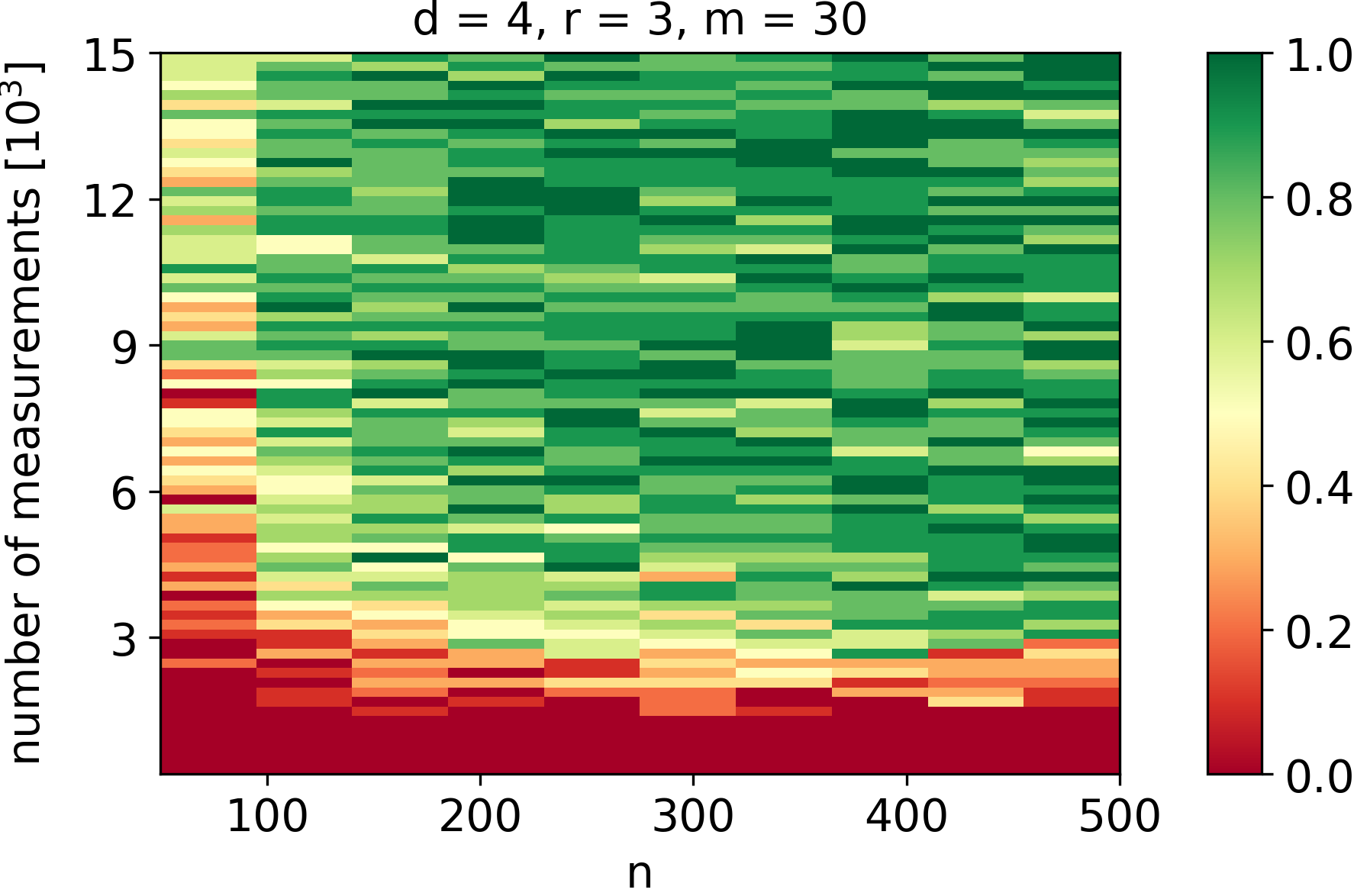}
	\caption{}
\end{subfigure}
\caption{Phase plots for noiseless $4$-dimensional CP completion with rank $r = 3$, perfect rank prediction $k = 3$, and side information size $m = 30$ after different numbers of iterations: $N_{iter} = 20$~(a), $N_{iter} = 50$~(b), $N_{iter} = 100$~(c), and $N_{iter} = 200$~(d).}
\label{num:fig:si_iter_d4}
\end{figure*}

We show the phase plots for $d = 2$, $r = 3$, $m = 30$ (Fig.~\ref{num:fig:no_si_d2}) and $d = 3$, $r = 3$, $m = 30$ (Fig.~\ref{num:fig:no_si_d3}). In both cases we made $N_{trial} = 5$ trials with $N_{ic} = 2$ different initial conditions, making $N_{iter} = 100$ iterations for $d = 2$ and $N_{iter} = 150$ iterations for $d = 3$. In the presence of side information, the phase transition curve for CP completion becomes horizontal, i.e. the critical size of $|\Omega|$ that makes completion possible is essentially independent of $n$ and, hence, greatly reduced. Indeed, a rank-3 CP tensor of size $300 \times 300 \times 300$ can be completed from $1\%$ of its elements, and only $0.004\%$ are needed when 30-dimensional side-information subspaces are available for all of its fibers (columns, rows, and tubes); for a larger $1000 \times 1000 \times 1000$ tensor with the same side information this reduces to $0.0001\%$. Similar behavior has been observed for Riemannian TT completion with side information \cite{BudzinskiyZamarashkinNote2020}.

Looking closely at Figs.~\ref{num:fig:no_si_d2} and \ref{num:fig:no_si_d3}, we can note that the phase transition curves have a bump for small values of $n$ (in fact, it is seen in the phase plots for TT completion as well \cite{BudzinskiyZamarashkinNote2020} but remained unnoticed). We explore this phenomenon by zooming in on the phase plots for $d = 3$, $r = 3$, and different values of $m$ (Fig.~\ref{num:fig:si_d3_bump}). The numerical results suggest that the bump occurs for $m$ comparable with $n$ and is absent when it is sufficiently small.

A possible explanation of the bump's existence is that FBCP-SI might require more iterations to converge when $m \lesssim n$. To check this, we carry out experiments for $d = 4$, $r = 3$, $m = 30$ with perfect rank prediction $k = 3$, $N_{trial} = 5$, $N_{ic} = 2$, and present the corresponding phase plots for different values of $N_{iter}$, ranging from $20$ to $200$ iterations; see Fig.~\ref{num:fig:si_iter_d4}. We see that even for $n = 100$ the threshold value of $|\Omega|$ descends rapidly with iterations and stabilizes after $N_{iter} = 50$. Meanwhile, the bump exists for $n = 50$ and persists nearly unchanged throughout 200 iterations. These results suggest that early stopping is likely not what keeps the phase transition curve from being completely horizontal. See Discussion for more thoughts about the bump.

\subsubsection{Performance of rank determination}
\begin{figure}[ht]
\begin{subfigure}[b]{0.5\linewidth}
\centering
	\includegraphics[width=0.7\linewidth]{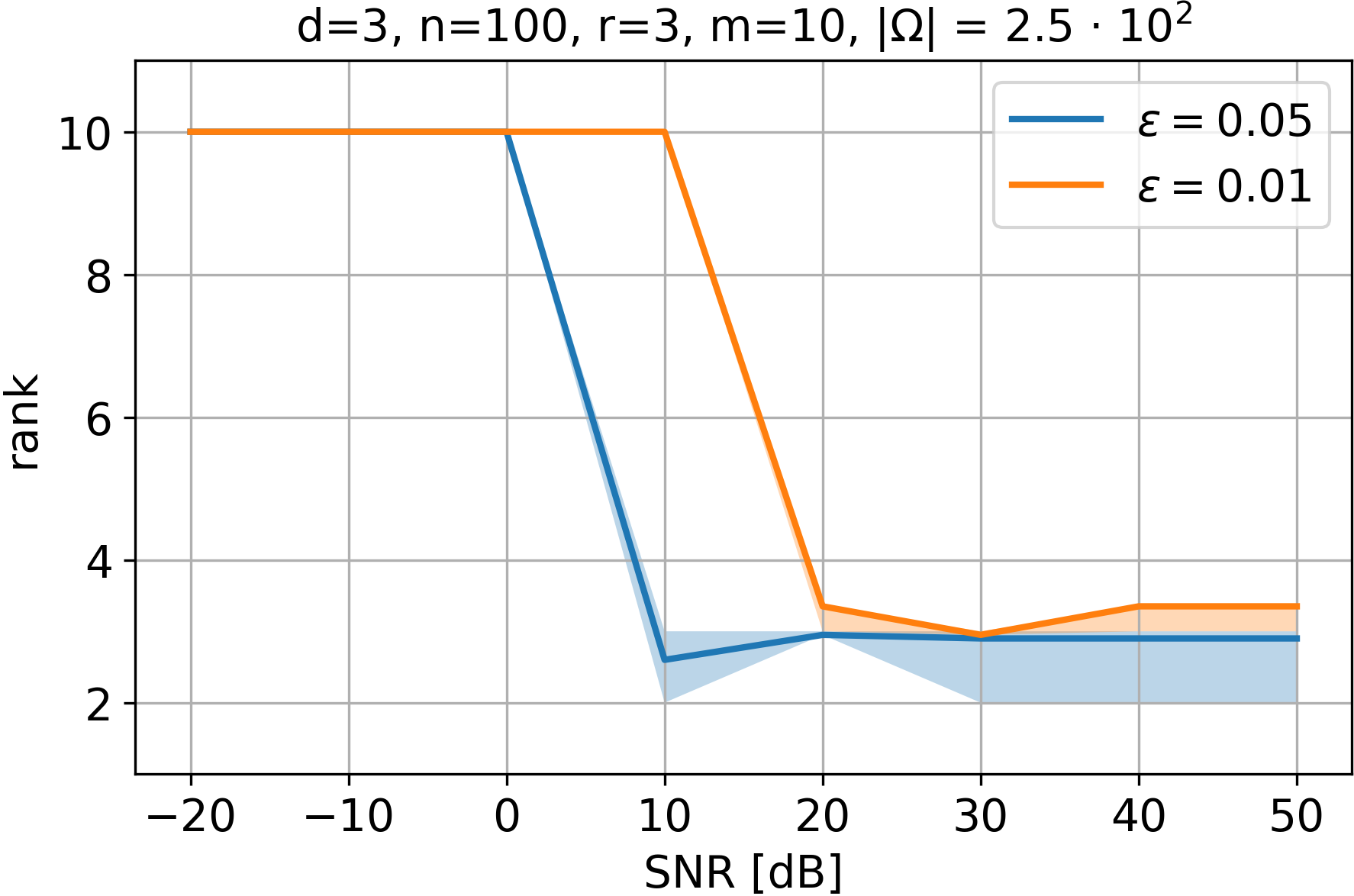}
	\caption{}
\end{subfigure}\hfill%
\begin{subfigure}[b]{0.5\linewidth}
\centering
	\includegraphics[width=0.7\linewidth]{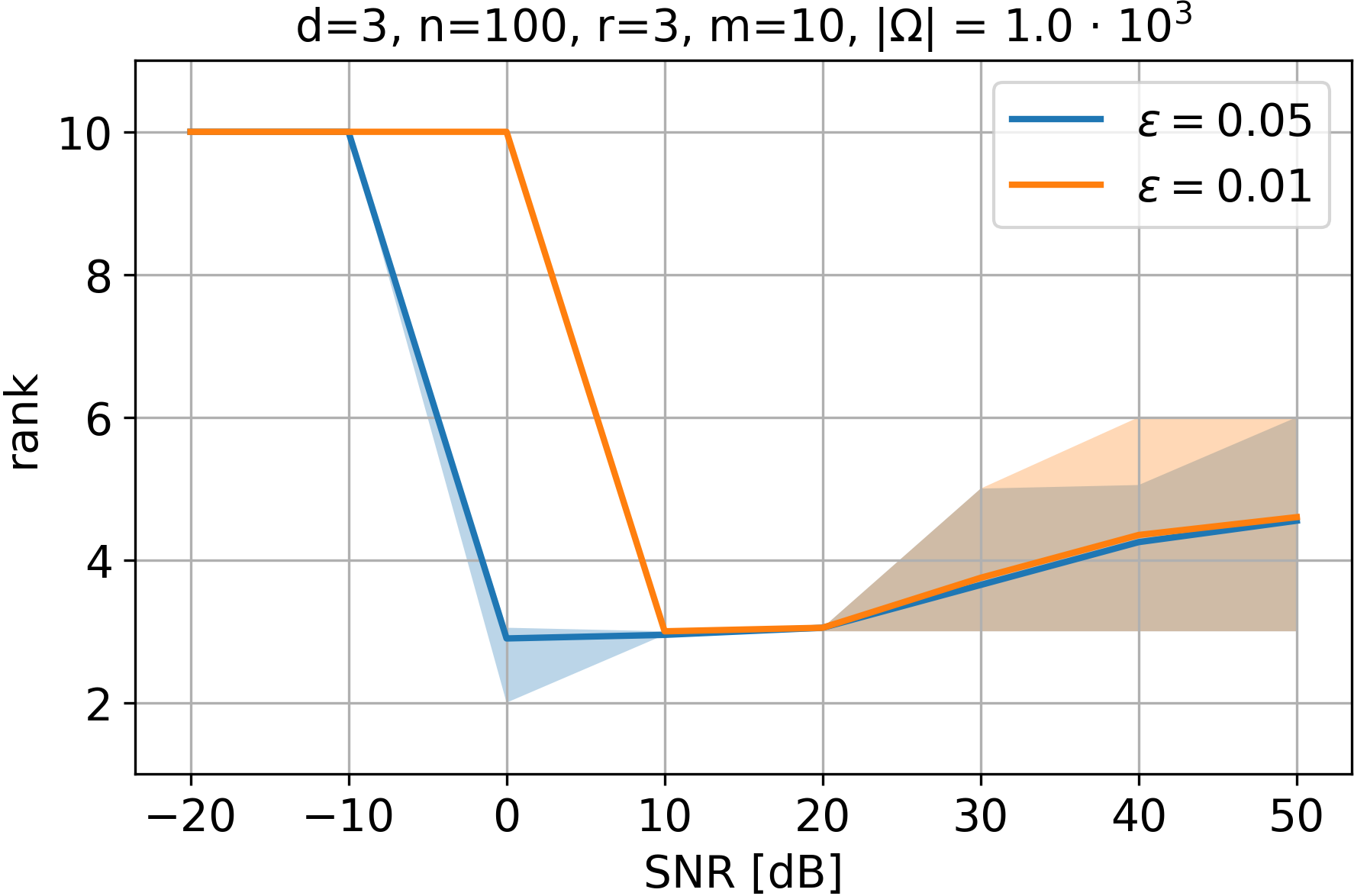}
	\caption{}
\end{subfigure}\hfill%
\begin{subfigure}[b]{0.5\linewidth}
\centering
	\includegraphics[width=0.7\linewidth]{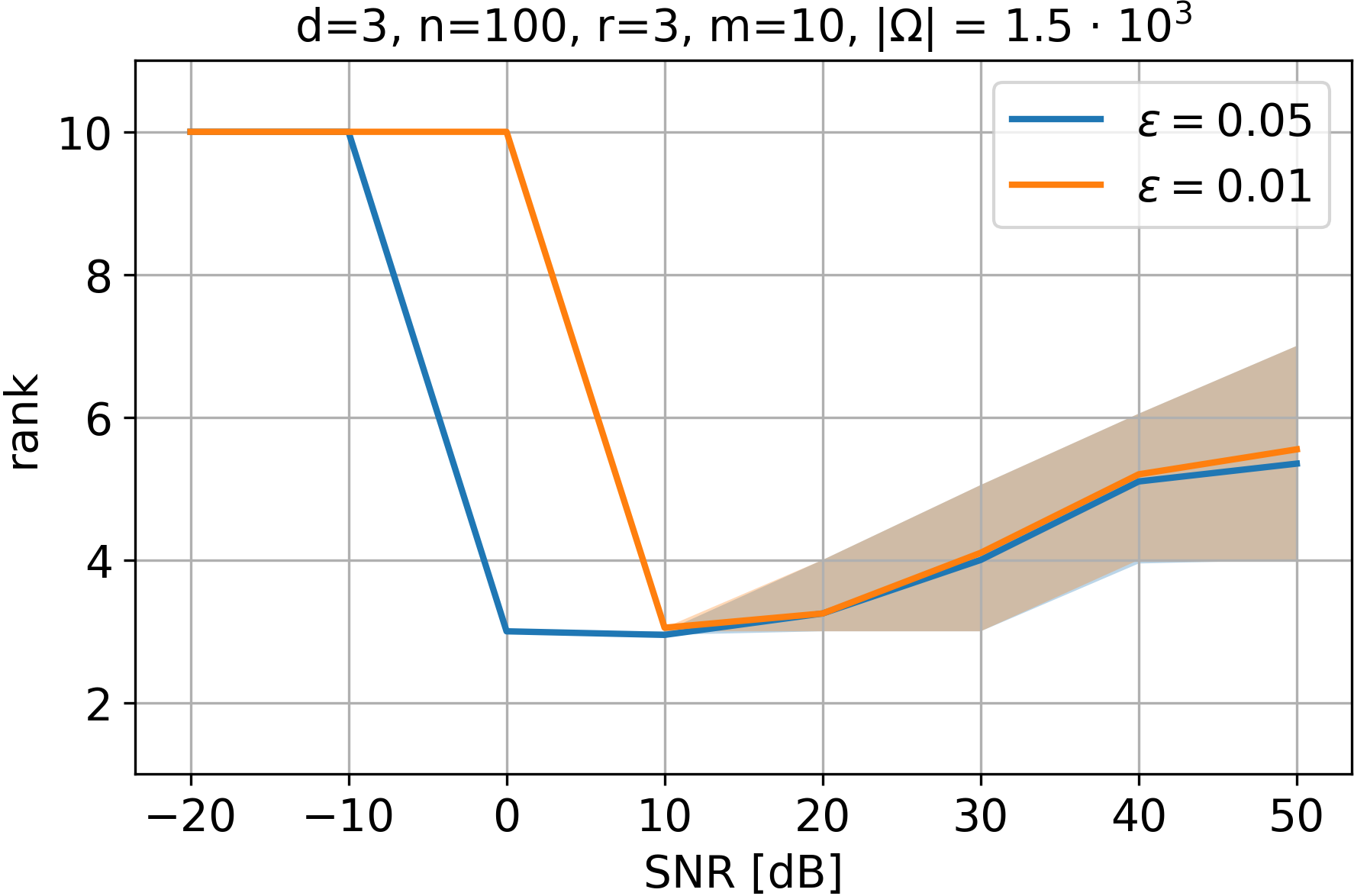}
	\caption{}
\end{subfigure}\hfill%
\begin{subfigure}[b]{0.5\linewidth}
\centering
	\includegraphics[width=0.7\linewidth]{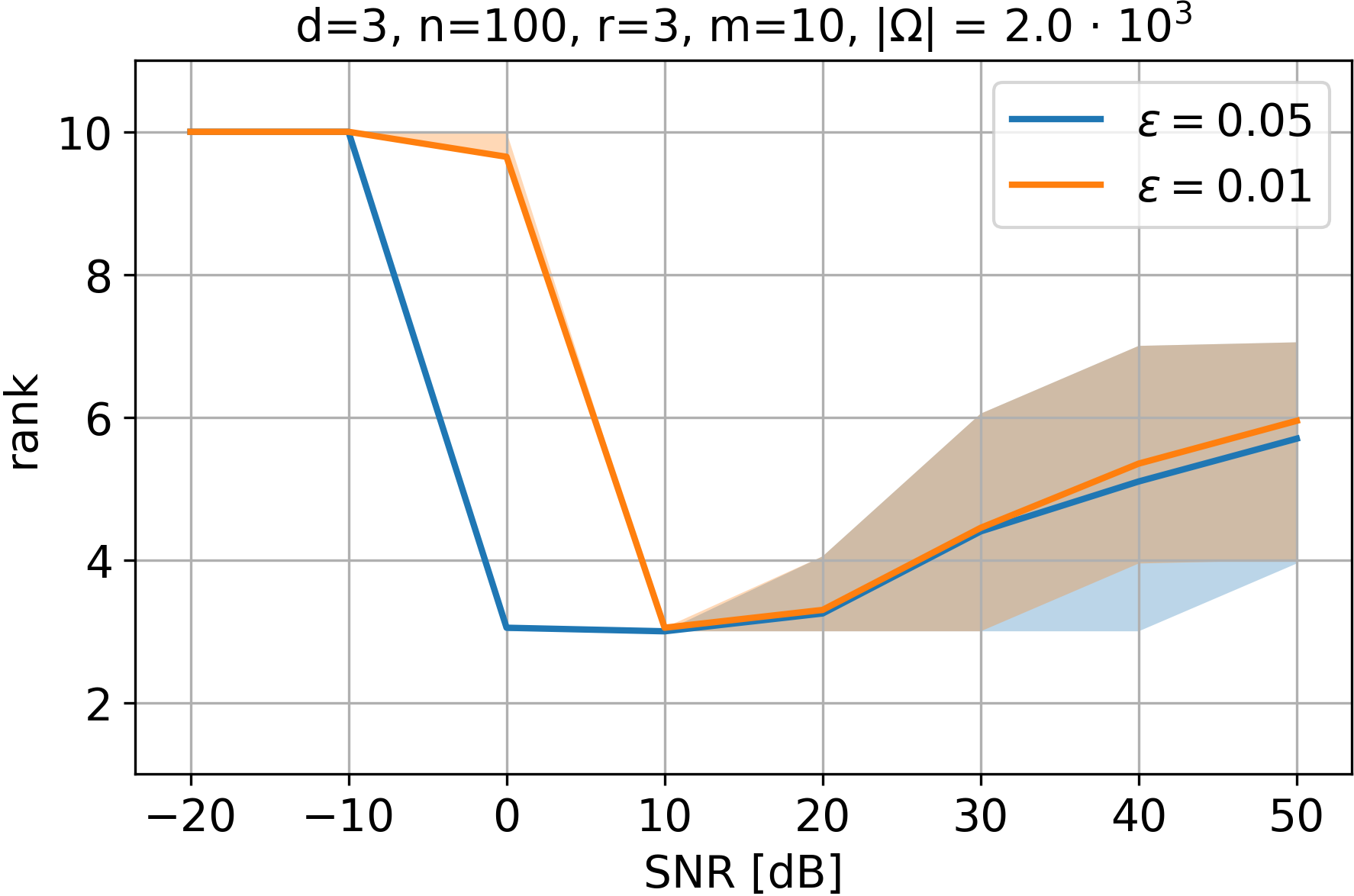}
	\caption{}
\end{subfigure}
\caption{Determined rank for 3-dimensional CP completion with size $n = 100$, rank $r = 3$, side information size $m = 10$, rank prediction $k = 10$, varying levels of noise and different numbers of samples: $|\Omega| = 2.5 \cdot 10^2$~(a), $|\Omega| = 1 \cdot 10^3$~(b), $|\Omega| = 1.5 \cdot 10^3$~(c), and $|\Omega| = 2 \cdot 10^3$~(d). The curves show the averaged rank together with the 5th and 95th percentiles for threshold values $\varepsilon = 0.05$ and $\varepsilon = 0.01$.}
\label{num:fig:si_d3_m10_snr_rank}
\end{figure}
\begin{figure}[ht]
\begin{subfigure}[b]{0.5\linewidth}
\centering
	\includegraphics[width=0.7\linewidth]{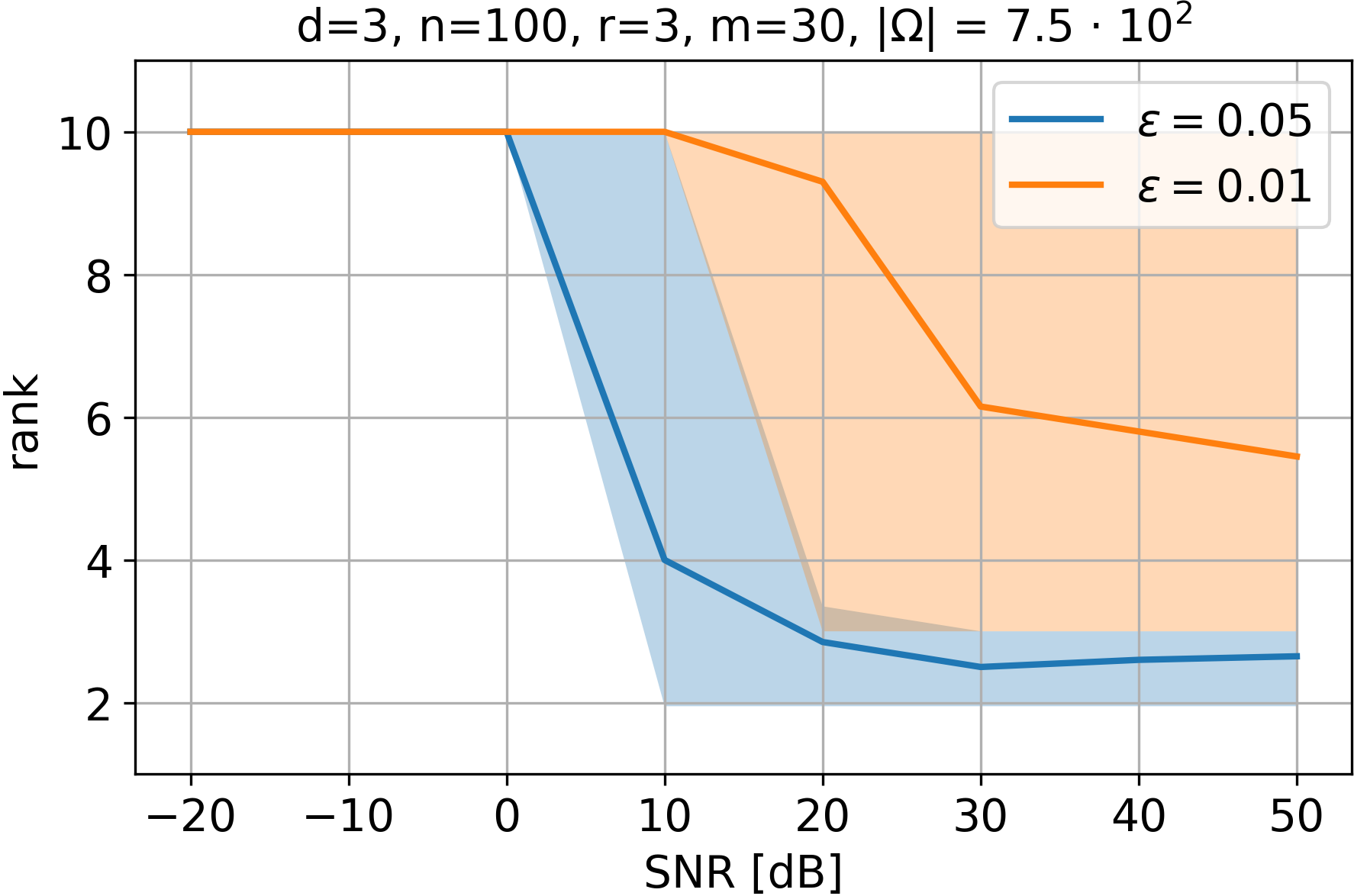}
	\caption{}
\end{subfigure}\hfill%
\begin{subfigure}[b]{0.5\linewidth}
\centering
	\includegraphics[width=0.7\linewidth]{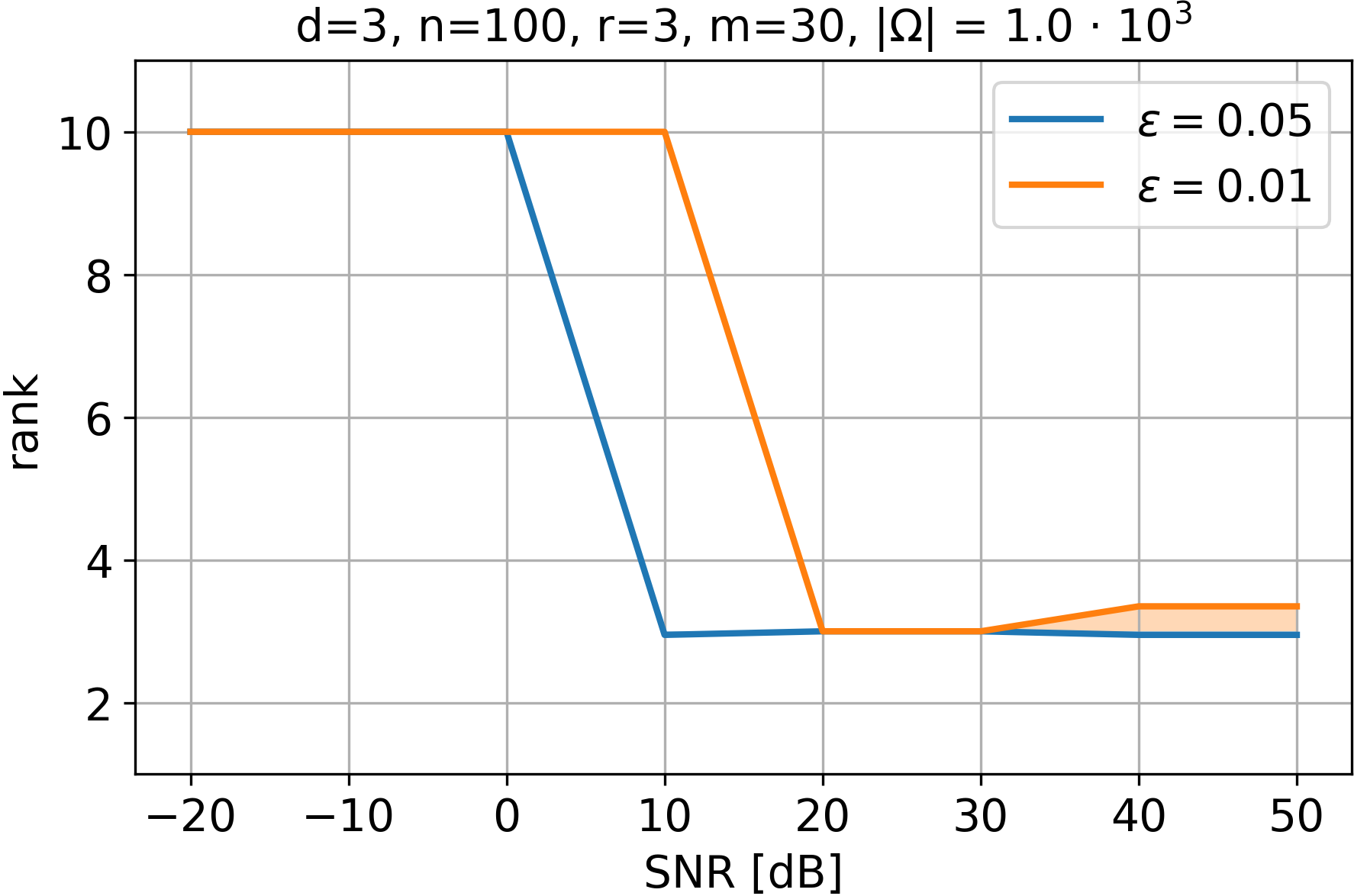}
	\caption{}
\end{subfigure}\hfill%
\begin{subfigure}[b]{0.5\linewidth}
\centering
	\includegraphics[width=0.7\linewidth]{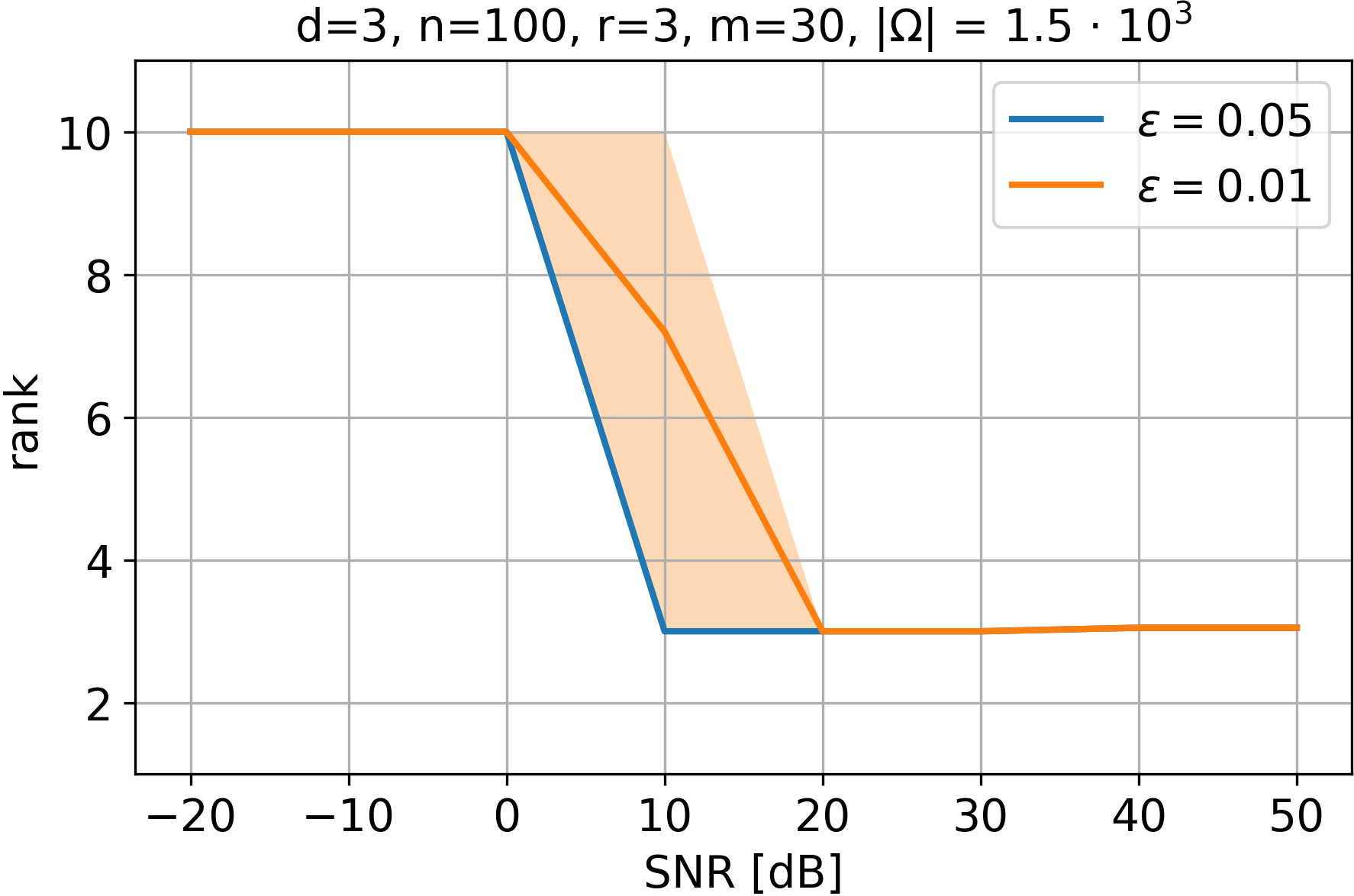}
	\caption{}
\end{subfigure}\hfill%
\begin{subfigure}[b]{0.5\linewidth}
\centering
	\includegraphics[width=0.7\linewidth]{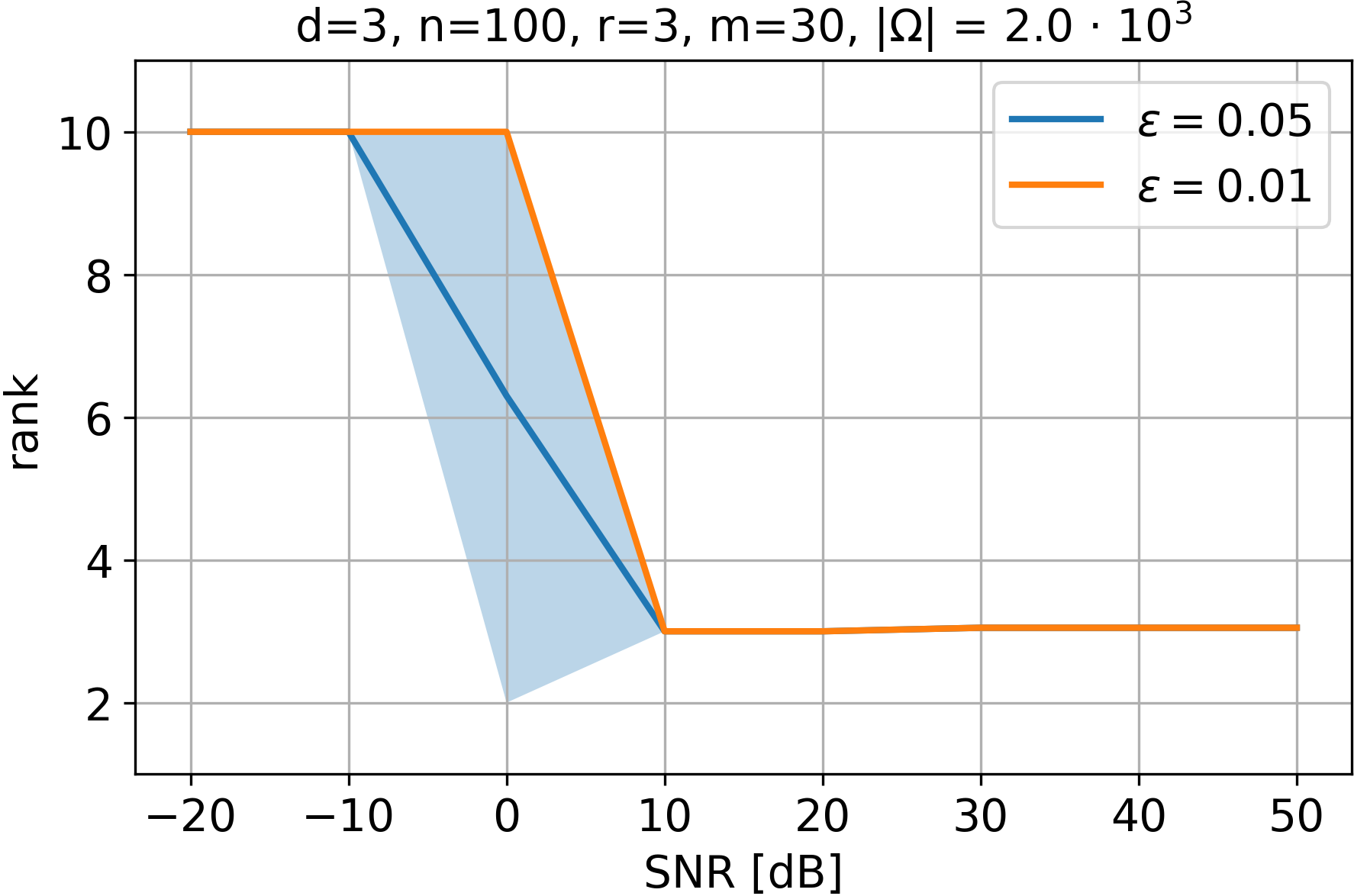}
	\caption{}
\end{subfigure}
\caption{Determined rank for 3-dimensional CP completion with size $n = 100$, rank $r = 3$, side information size $m = 30$, rank prediction $k = 10$, varying levels of noise and different numbers of samples: $|\Omega| = 2.5 \cdot 10^2$~(a), $|\Omega| = 1 \cdot 10^3$~(b), $|\Omega| = 1.5 \cdot 10^3$~(c), and $|\Omega| = 2 \cdot 10^3$~(d). The curves show the averaged rank together with the 5th and 95th percentiles for threshold values $\varepsilon = 0.05$ and $\varepsilon = 0.01$.}
\label{num:fig:si_d3_m30_snr_rank}
\end{figure}

In the previous examples, we assumed that the CP-rank was given in advance, i.e. the predicted rank $k$ was always equal to the true rank $r$. In the following experiments, we consider a more realistic scenario where only an upper bound of the rank is known. To test automatic rank determination of FBCP-SI, we generated random rank-3 CP tensors of size $100 \times 100 \times 100$ and ran $N_{iter} = 100$ iterations with $k = 10$ for different values of $|\Omega|$ and various levels of noise. We determine the rank based on the posterior means of the hyperparameters $\{ \lambda_j \}$ and a threshold parameter $\varepsilon > 0$:
\begin{equation*}
    r_{\varepsilon} = \Big| \Big\{ j : j \in [k], \frac{d_j}{c_j} \geq \varepsilon \max_{i \in [k]} \frac{d_i}{c_i} \Big\} \Big|.
\end{equation*}
We plot the rank $r_{\varepsilon}$, averaged over $N_{trial} = 20$ trials with $N_{ic} = 1$, against SNR for different values of $\varepsilon$; see Figs.~\ref{num:fig:si_d3_m10_snr_rank} and \ref{num:fig:si_d3_m30_snr_rank} with $m = 10$ and $m = 30$, respectively. Comparing the two Figs., we see that given the same number of samples $|\Omega|$, lower dimension of side information subspaces leads to better rank determination at SNR of 0dB; however, at the same time, the rank tends to be overestimated for high SNR.

Find more experiments with synthetic data in the Appendix.

\subsection{Facial images}
\begin{figure*}[ht]
\centering
\begin{subfigure}[b]{0.25\linewidth}
\centering
	\includegraphics[width=\linewidth]{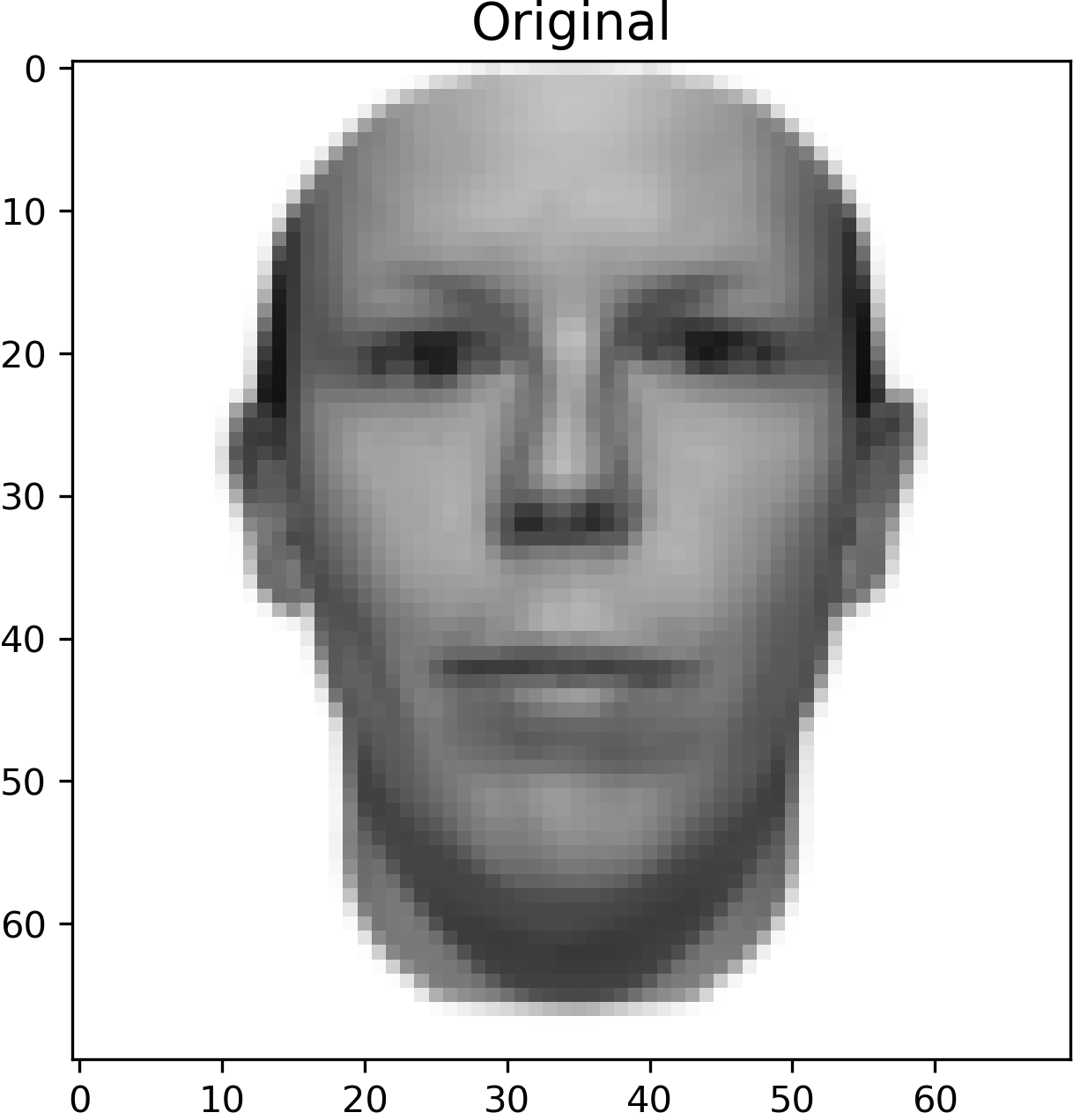}
	\caption{}
\end{subfigure}%
\begin{subfigure}[b]{0.25\linewidth}
\centering
	\includegraphics[width=\linewidth]{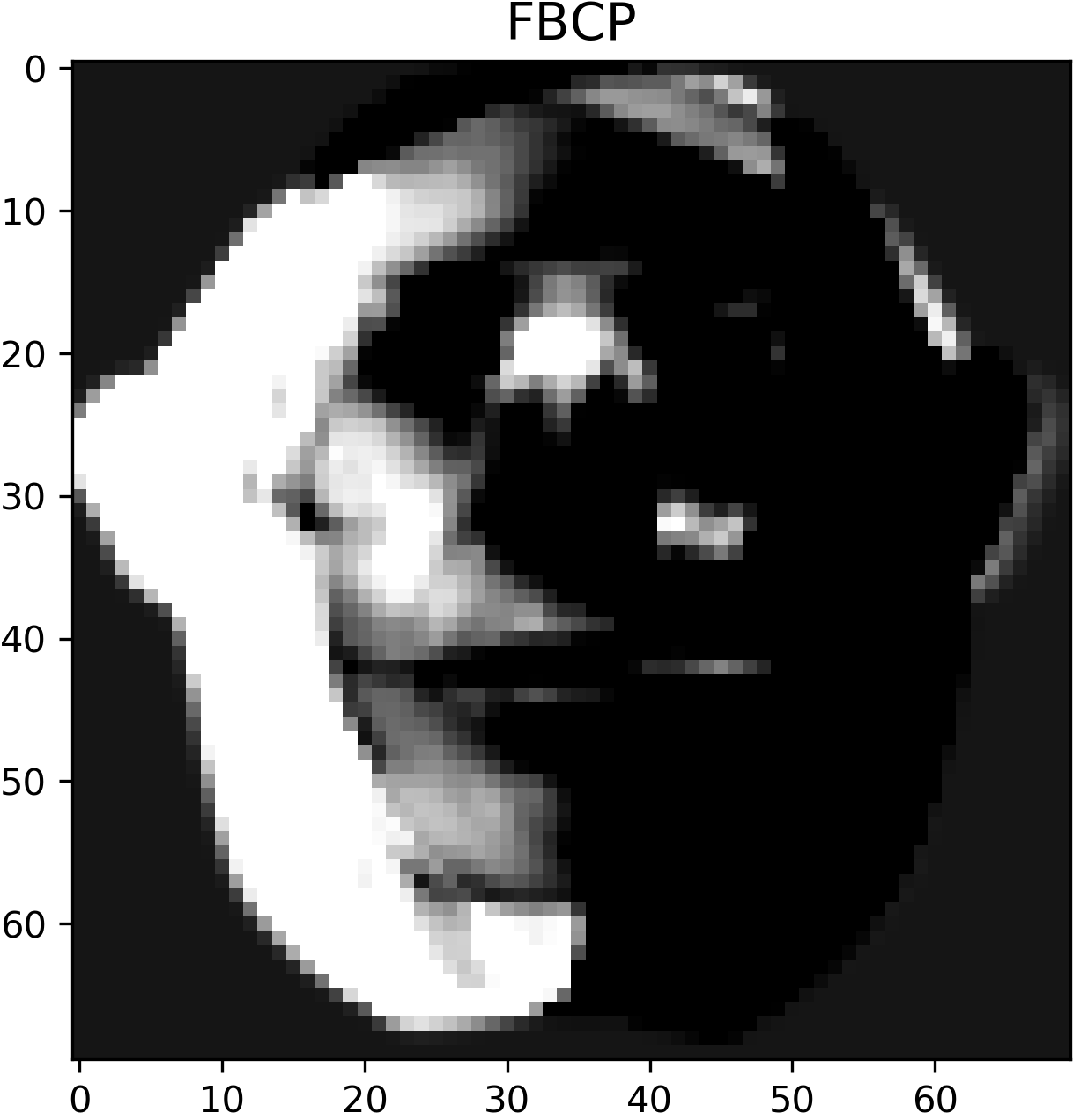}
	\caption{}
\end{subfigure}%
\begin{subfigure}[b]{0.25\linewidth}
\centering
	\includegraphics[width=\linewidth]{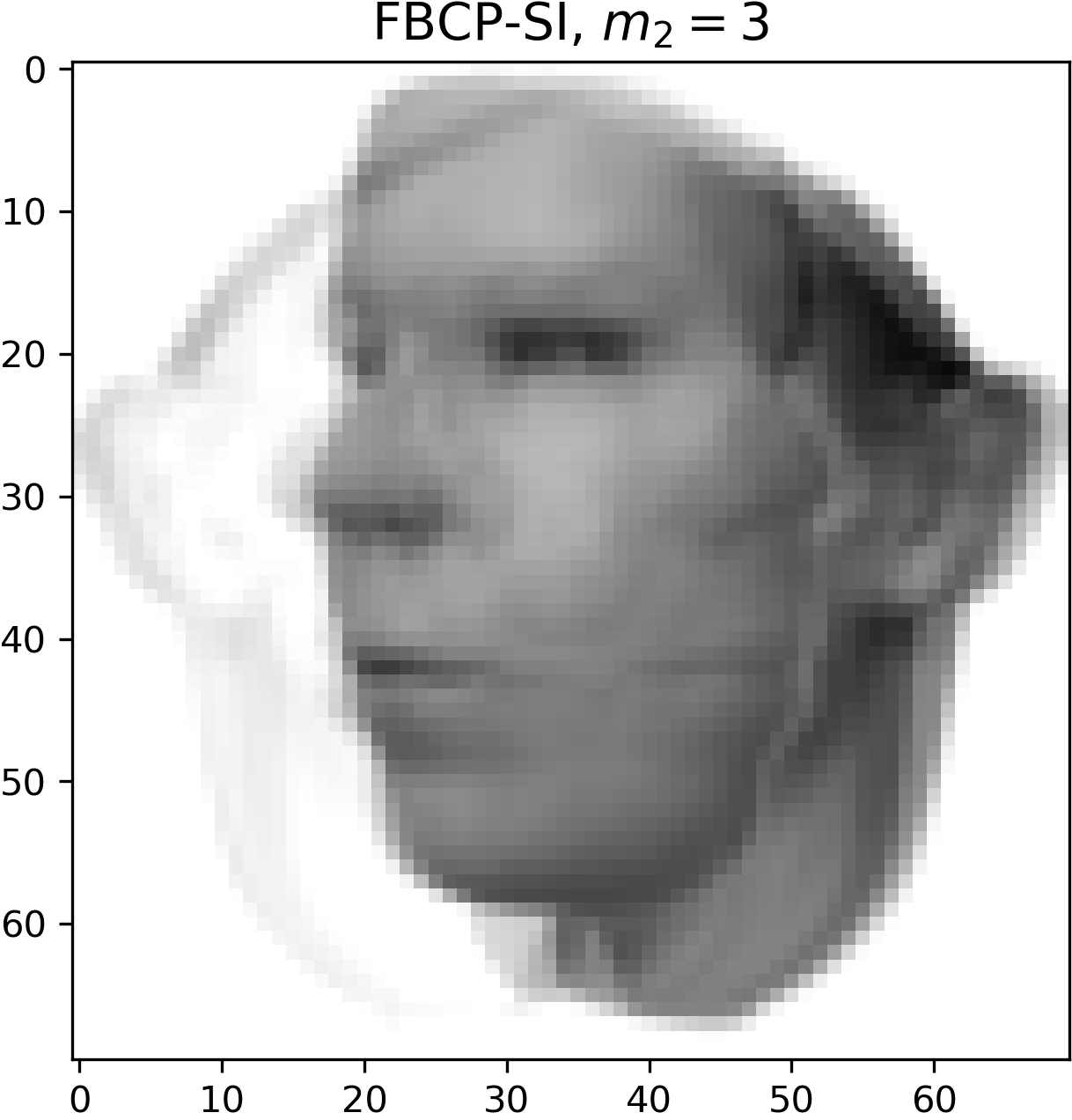}
	\caption{}
\end{subfigure}%
\begin{subfigure}[b]{0.25\linewidth}
\centering
	\includegraphics[width=\linewidth]{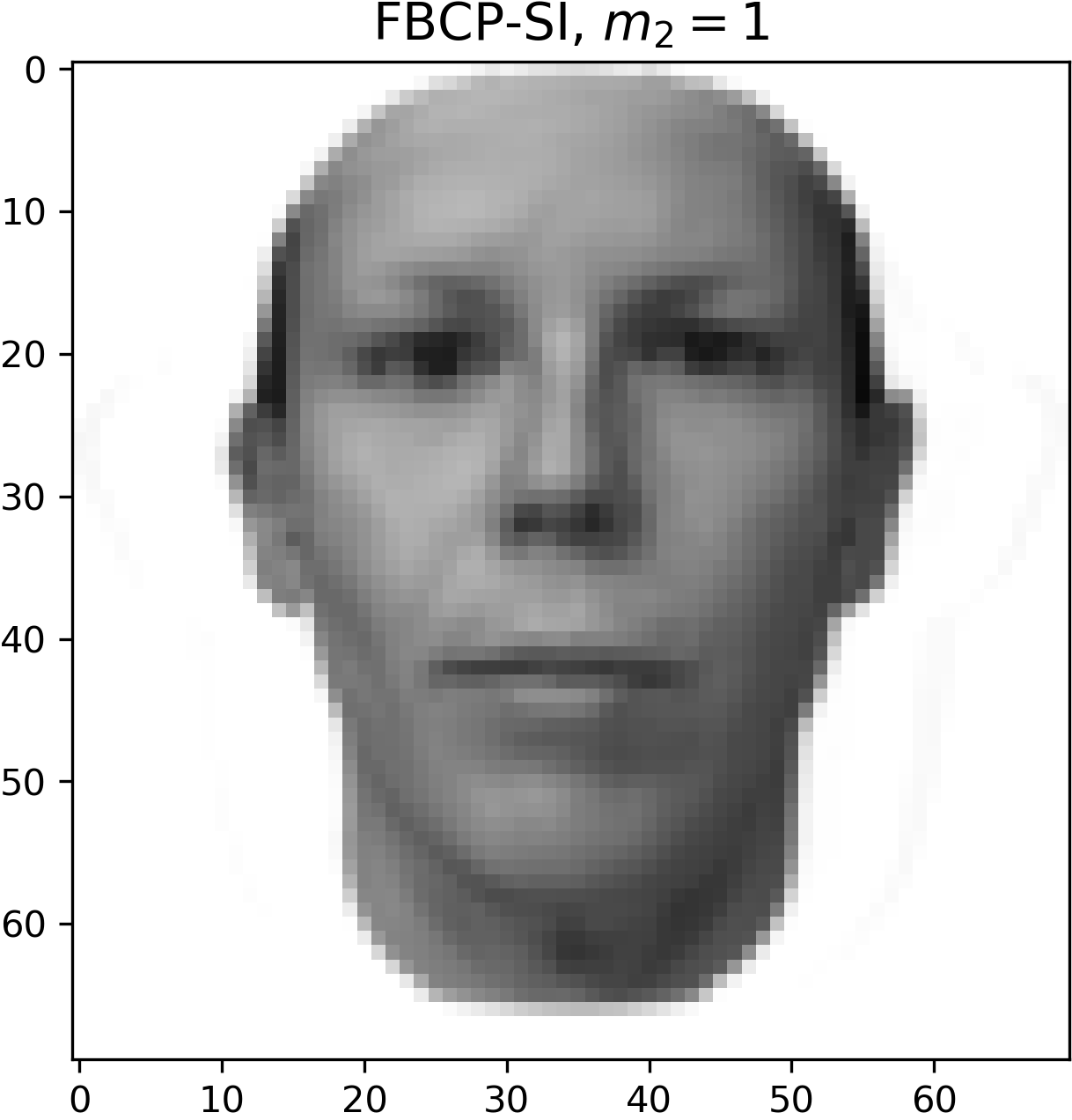}
	\caption{}
\end{subfigure}
\caption{An example of a recovered image from the 3D Basel Face Model by rank-10 $4900 \times 3 \times 9$ tensor completion, when 1 out of 3 images are known for each angle: original~(a); FBCP~(b); FBCP-SI with $m_1 = 9$, $m_2 = 3$, $m_3 = 9$~(c); FBCP-SI with $m_1 = 9$, $m_2 = 1$, $m_3 = 9$~(d).}
\label{num:fig:faces}
\end{figure*}
To verify the performance of our algorithm on real-world data, we used the 3D Basel Face Model \cite{paysan20093d}: a collection of facial images of 10 people, taken from 9 angles under 3 light settings. We cropped and rescaled each image to $70 \times 70$ pixels.

For the first experiment, we pick out one person and consider her portraits as a $4900 \times 3 \times 9$ tensor. We choose $m_1 = 9$, $m_2 = 3$, and $m_3 = 9$, with the side information subspace for 'faces' spanned by the 9 first singular vectors of the $4900 \times 27$ flattening. In Table~\ref{numerical:tab:many}, we compare how FBCP and FBCP-SI find a rank-10 approximation of the data when all 27 portraits are known and when 7 of them, chosen at random, are missing. In the approximation case (all portraits are known), FBCP-SI shows better accuracy both in RMSE and SSIM metrics. In the completion case, FBCP-SI recovers the missing images worse than FBCP: it struggles in complete accordance with the bump phenomenon that we saw in synthetic experiments, since $m_2 = n_2$.
\begin{table}[h]\centering
\caption{The results of rank-10 $4900 \times 3 \times 9$ CP completion for the 3D Basel Face Model by FBCP and FBCP-SI with $m_1 = 9$, $m_2 = 3$, $m_3 = 9$.}
\begin{tabular}{@{}llcccc@{}}
    \toprule
    & & \multicolumn{2}{c}{RMSE} & \multicolumn{2}{c}{SSIM} \\
    \cmidrule(lr){3-4}\cmidrule(lr){5-6}
    Observed & Method & Obs. & Miss. & Obs. & Miss. \\
    \midrule
    27/27 & FBCP & 0.04 & N/A & 0.91 & N/A \\
     & FBCP-SI & 0.03 & N/A & 0.97 & N/A \\
    \midrule
    20/27 & FBCP & 0.05 & 0.09 & 0.90 & 0.81 \\
     & FBCP-SI & 0.04 & 0.22 & 0.91 & 0.61 \\
    \bottomrule\\
\end{tabular}
\label{numerical:tab:many}
\end{table}

We check this in the second experiment, where we keep only 9 images out of 27: one per angle. We compare three methods: FBCP, FBCP-SI with the same parameters are above, and FBCP-SI with $m_2$ reduced to 1. For the latter, the side information subspace is learnt from the 9 other people in the dataset as the dominant left singular vector of their $3 \times (4900 \cdot 9 \cdot 9)$ flattening. The results are presented in Table~\ref{numerical:tab:few}. We see that FBCP-SI with $m_2 = 1$ succeeds in recovering 18 missing images, while FBCP and FBCP-SI with $m_2 = 3$ fail to do so (see Fig.~\ref{num:fig:faces}).
\begin{table}[h]\centering
\caption{The results of rank-10 $4900 \times 3 \times 9$ CP completion for the 3D Basel Face Model, when 1 out of 3 images are known for each angle: FBCP; FBCP-SI with $m_1 = 9$, $m_2 = 3$, $m_3 = 9$; FBCP-SI with $m_1 = 9$, $m_2 = 1$, $m_3 = 9$.}
\begin{tabular}{@{}lcccc@{}}
    \toprule
    & \multicolumn{2}{c}{RMSE} & \multicolumn{2}{c}{SSIM} \\
    \cmidrule(lr){2-3}\cmidrule(lr){4-5}
    Method & Obs. & Miss. & Obs. & Miss. \\
    \midrule
    FBCP & 0.08 & 0.96 & 0.86 & 0.26 \\
    FBCP-SI, $m_2 = 3$ & 1.00 & 0.82 & 0.001 & 0.09 \\
    FBCP-SI, $m_2 = 1$ & 0.04 & 0.07 & 0.91 & 0.88 \\
    \bottomrule\\
\end{tabular}
\label{numerical:tab:few}
\end{table}
\clearpage
\section{Discussion}
We considered the problem of low-rank CP tensor completion with side information in the Bayesian framework. Having fixed a probabilistic model, we derived formulas for variational approximate Bayesian inference and the corresponding message passing algorithm. The results of numerical experiments allow us to analyze the regularization properties induced by side information: how it affects the phase plots, the rate of convergence, the attainable errors in the presence of noise, and automatic rank determination. The strongest point of our algorithm is that it significantly reduces the number of elements needed for successful completion of a tensor. For instance, a rank-3 CP tensor of size $300 \times 300 \times 300$ can be recovered from $1\%$ of its entries without side information and from only $0.004\%$ if there is 30-dimensional side information. This suggests that our method can be useful for applications where data are exceptionally scarce.

We would also like to add a few words about the bump in the phase transition curves that we observed for $m \lesssim n$. In \cite{BudzinskiyZamarashkinNote2020}, such bump can be recognized on the phase plot corresponding to Riemannian tensor train completion with side information of a 10-dimensional tensor. At the same time, the results of \cite{BudzinskiyZamarashkinTensor2021} tell us that Riemannian gradient descent converges \textit{locally} if the number of samples $|\Omega|$ exceeds a certain threshold that depends on $m$ and is independent of $n$, and this behavior is indeed seen on the phase plots for larger values of $n$. However, random initialization that is used in \cite{BudzinskiyZamarashkinNote2020} (and in this paper too) certainly does not put the initial condition into the basin of local attraction. Recent results on non-convex optimization \cite{ChenEtAlGradient2019, ChiEtAlNonconvex2019, MaEtAlImplicit2020} show that gradient descent converges \textit{globally} in certain problems (including matrix completion) when initialized randomly, provided, of course, that $|\Omega|$ is large enough. So the existence of the bump could find its explanation in the delicate analysis of global convergence from a random initial point for tensor completion with side information: it is possible that the threshold value of $|\Omega|$ that guarantees global convergence depends on $n$ when $m \lesssim n$.

\section*{Acknowledgements}
This work was supported by Russian Science Foundation (project 21-71-10072).

\bibliography{common/bib}
\bibliographystyle{unsrt}

\clearpage
\appendix
\section{Used distributions}
The main distributions that we use throughout the text are
\begin{itemize}
\item the Gaussian distribution with mean $\mu \in \Real$ and precision $\beta > 0$ (the inverse of variance)
\begin{equation*}
    \mathcal{N}(x | \mu, \beta^{-1}) = \sqrt{\frac{\beta}{2\pi}} e^{-\frac{\beta}{2} (x - \mu)^2};
\end{equation*}
\item the multivariate Gaussian distribution with mean $\mu \in \Real^n$ and positive definite precision matrix $B$
\begin{equation*}
    \mathcal{N}(x | \mu, B^{-1} ) = \sqrt{\frac{\det B}{(2\pi)^n}} e^{-\frac{1}{2} (x - \mu)^T B (x - \mu)};
\end{equation*}
\item the Gamma distribution with shape $a > 0$ and rate $b > 0$ parameters
\begin{equation*}
    \mathcal{G}(x | a, b) = \frac{b^a x^{a-1} e^{-bx}}{\Gamma(a)}.
\end{equation*}
\end{itemize}
All of them belong to the exponential family of distributions because their densities can be expressed as
\begin{equation*}
    \mathcal{N}(x | \mu, \beta^{-1}) = \exp\left\{ \begin{bmatrix} \beta\mu \\ -\frac{\beta}{2} \end{bmatrix}^T \begin{bmatrix} x \\ x^2 \end{bmatrix} + \frac{1}{2} (\log \beta - \beta\mu^2 - \log 2\pi) \right\},
\end{equation*}
\begin{equation*}
    \mathcal{G}(x | a, b) = \exp\left\{ \begin{bmatrix} -b \\ a - 1 \end{bmatrix}^T \begin{bmatrix} x \\ \log x \end{bmatrix} + (a \log b - \log\Gamma(a)) \right\},
\end{equation*}
\begin{equation*}
    \mathcal{N}(x | \mu, B^{-1} ) = \exp\left\{ \begin{bmatrix} B \mu \\ \text{col}(B) \end{bmatrix}^T  \begin{bmatrix} x \\ \text{col}(x x^T) \end{bmatrix} + \frac{1}{2} (\log\det B - \mu^T B \mu - n \log 2\pi) \right\}.
\end{equation*}

Consider a probabilistic model, where a Gamma prior $\mathcal{G}(\beta | a, b)$ is put on the precision of a Gaussian distribution $\mathcal{N}(x | \mu, \beta^{-1})$. This a simple example of what is known as a conjugate-exponential model: the parameter $\beta$ that establishes the link between the two distributions enters both of them as $\begin{bmatrix} \beta & \log \beta \end{bmatrix}^T$. If we marginalize it out, we get the Student's $t$-distribution
\begin{equation*}
    \text{St}\left(x \Big| \mu, \frac{a}{b}, 2a \right) = \int_0^{\infty} \mathcal{N}(x | \mu, \beta^{-1}) \mathcal{G}(\beta | a, b) d\beta = \frac{\Gamma(a + 1/2)}{\Gamma(a)} \sqrt{\frac{1}{2 \pi b}} \left[ 1 + \frac{(x - \mu)^2}{2b} \right]^{\frac{1}{2}-a}
\end{equation*}
with mean $\mu$ and variance $b / (a - 1)$.
\section{Optimal factorized variational distributions}
\subsection{General form}
Recall that in variational Bayesian inference our goal is to minimize the Kullback--Leibler divergence between the variational posterior $q(\Theta)$ and the real posterior $p(\Theta | Y_{\Omega})$, or, equivalently, to minimize
\begin{equation*}
    F(q) = \int q(\Theta) \log \frac{q(\Theta)}{p(Y_{\Omega}, \Theta)} d\Theta \to \min_{q(\Theta)}.
\end{equation*}
Let us look for $q(\Theta)$ in a factorized form
\begin{equation*}
    q(\Theta) = \prod_i q(\theta_i)
\end{equation*}
and substitute it into the minimization problem:
\begin{align*}
    F(q) &= \int q(\Theta) \log q(\Theta) d\Theta - \int q(\Theta) \log p(Y_{\Omega}, \Theta) d\Theta \\
    &= \sum_{j} \int q(\theta_j) \log q(\theta_j) d\theta_j - \int q(\Theta) \log p(Y_{\Omega}, \Theta) d\Theta \\
    &= \sum_{j} \int q(\theta_j) \log q(\theta_j) d\theta_j - \int q(\theta_i) \underbrace{\left\{ \int \prod_{j \neq i} q(\theta_j) \log p(Y_{\Omega}, \Theta) \prod_{j \neq i} d\theta_j \right\}}_{\log q^*(\theta_i) + \text{const}}  d\theta_i  \\
    &= \sum_{j \neq i} \int q(\theta_j) \log q(\theta_j) d\theta_j - \int q(\theta_i) \log \frac{q^*(\theta_i)}{q(\theta_i)} d\theta_i + \text{const} \\
    &= \sum_{j \neq i} \int q(\theta_j) \log q(\theta_j) d\theta_j + \mathcal{KL}\big[q(\theta_i) ~||~ q^*(\theta_i)\big] + \text{const}.
\end{align*}
Since the Kullback--Leibler divergence is non-negative and equals to zero if and only if the two distributions coincide almost everywhere, we must choose $q(\theta_i) = q^*(\theta_i)$ to minimize $F(q)$ if all the remaining $q(\theta_j)$ are fixed.

\subsection{Matrix case}
To compute $q^*(\theta_i)$ according to
\begin{equation*}
    \log q^*(\theta_i) = \E_{{\sim} \theta_i} \{ \log p(Y_{\Omega}, \Theta) \} + \mathrm{const},
\end{equation*}
we basically need to expand $\log p(Y_{\Omega}, \Theta)$ and collect the relevant terms. This is where the exponential conjugacy of our model comes in handy.

\subsubsection{Factor matrices $U$ and $V$}
Let us begin with the factor matrix $U$. Recall that we write $\overline{u} \in \Real^{m_1 k}$ and $\overline{v} \in \Real^{m_2 k}$ for the vectorizations of $U$ and $V$ obtained by stacking their columns as
\begin{equation*}
    \overline{u} = 
    \begin{bmatrix}
    u_{1 1} & \ldots & u_{m_1 1} & u_{1 2} & \ldots & u_{m_1 k}
    \end{bmatrix}^T, \quad 
    \overline{v} = 
    \begin{bmatrix}
    v_{1 1} & \ldots & v_{m_2 1} & v_{1 2} & \ldots & v_{m_2 k}
    \end{bmatrix}^T.
\end{equation*}
For brevity, we will also use the following notation:
\begin{equation*}
    g_{i_1}^T U V^T h_{i_2} = \overline{u}^T (I_k \otimes g_{i_1} h_{i_2}^T) \overline{v} = \overline{u}^T \overline{\varphi}_{i_1 i_2}.
\end{equation*}
Then we have
\begin{align*}
    \log p(Y_{\Omega}, \Theta) &= \sum_{\Omega} \log p(y_{i_1 i_2} | U,  V, \tau) + \sum_{j_1 = 1}^{m_1} \log p(u_{j_1} | \Lambda) + \ldots \\
    &= \sum_{\Omega} \left\{ \tau y_{i_1 i_2} \overline{u}^T \overline{\varphi}_{i_1 i_2} - \frac{\tau}{2} \overline{u}^T \overline{\varphi}_{i_1 i_2} \overline{\varphi}_{i_1 i_2}^T \overline{u} \right\} + \sum_{j_1 = 1}^{m_1} \left\{ -\frac{1}{2} u_{j_1}^T \Lambda u_{j_1} \right\} + \ldots \\
    &= \sum_{\Omega} \left\{ \tau y_{i_1 i_2} \overline{u}^T \overline{\varphi}_{i_1 i_2} - \frac{\tau}{2} \overline{u}^T \overline{\varphi}_{i_1 i_2} \overline{\varphi}_{i_1 i_2}^T \overline{u} \right\} - \frac{1}{2} \overline{u}^T \left( \Lambda \otimes I_{m_1} \right) \overline{u} + \ldots \\
    &= -\frac{1}{2} \overline{u}^T \left( \Lambda \otimes I_{m_1} + \tau \sum_{\Omega} \overline{\varphi}_{i_1 i_2} \overline{\varphi}_{i_1 i_2}^T \right) \overline{u} + \overline{u}^T \left( \tau \sum_{\Omega} y_{i_1 i_2} \overline{\varphi}_{i_1 i_2} \right) + \ldots
\end{align*}
By comparing this formula with the multivariate Gaussian distribution, we readily see that $q^*(U)$ is a Gaussian distribution
\begin{equation*}
    q^*(U) = \mathcal{N}(\overline{u} | \mu_{\overline{u}}, A_{\overline{u}})
\end{equation*}
with covariance
\begin{align*}
    A_{\overline{u}} &= \left[ \E\Big\{\Lambda \otimes I_{m_1} + \tau \sum_{\Omega} \overline{\varphi}_{i_1 i_2} \overline{\varphi}_{i_1 i_2}^T \Big\} \right]^{-1} \\
    &= \left[ \E\{\Lambda\} \otimes I_{m_1} + \E\{\tau\} \sum_{\Omega} \E\{ \overline{\varphi}_{i_1 i_2} \overline{\varphi}_{i_1 i_2}^T \} \right]^{-1} \\
    &= \left[ \E\{\Lambda\} \otimes I_{m_1} + \E\{\tau\} \sum_{\Omega} (I_k \otimes g_{i_1} h_{i_2}^T) \E\{ \overline{v} \overline{v}^T \} (I_k \otimes h_{i_2} g_{i_1}^T) \right]^{-1} \\
    &= \left[ \E\{\Lambda\} \otimes I_{m_1} + \E\{\tau\} \sum_{\Omega} (I_k \otimes h_{i_2}^T) \E\{ \overline{v} \overline{v}^T \} (I_k \otimes h_{i_2}) \otimes g_{i_1} g_{i_1}^T \right]^{-1}
\end{align*}
and mean
\begin{align*}
    \mu_{\overline{u}} &= A_{\overline{u}} \E\Big\{ \tau \sum_{\Omega} y_{i_1 i_2} \overline{\varphi}_{i_1 i_2} \Big\} \\
    &= \E\{\tau\} A_{\overline{u}} \sum_{\Omega} y_{i_1 i_2} (I_k \otimes g_{i_1} h_{i_2}^T) \E\{ \overline{v} \} \\
    &= \E\{\tau\} A_{\overline{u}} \sum_{\Omega} y_{i_1 i_2} (I_k \otimes h_{i_2}^T) \E\{ \overline{v} \} \otimes g_{i_1}.
\end{align*}
We used the factorized form of $q(\Theta)$ in decoupling the expectations. The second factor matrix $V$ is dealt with in complete analogy.

\subsubsection{Precision matrix $\Lambda$}
For the precision matrix, we keep track of different terms in the expansion:
\begin{align*}
    \log p(Y_{\Omega}, \Theta) &= \sum_{i_1 = 1}^{m_1} \log p(u_{i_1} | \Lambda) + \sum_{i_2 = 1}^{m_2} \log p(v_{i_2} | \Lambda) + \sum_{j = 1}^{k} \log p(\lambda_j) + \ldots \\
    &= \sum_{i_1 = 1}^{m_1} \left\{ -\frac{1}{2} u_{i_1}^T \Lambda u_{i_1} + \frac{1}{2} \sum_{j = 1}^{k} \log \lambda_j \right\} + \sum_{i_2 = 1}^{m_2} \left\{ -\frac{1}{2} v_{i_2}^T \Lambda v_{i_2} + \frac{1}{2} \sum_{j = 1}^{k} \log \lambda_j \right\} \\ 
    &+ \sum_{j = 1}^{k} \left\{ -b_j \lambda_j + (a_j - 1) \log\lambda_j \right\} + \ldots \\
    &= \sum_{j = 1}^{k} \left\{ \lambda_j \left( -b_j - \frac{1}{2} \sum_{i_1 = 1}^{m_1} u_{i_1 j}^2 - \frac{1}{2} \sum_{i_2 = 1}^{m_2} v_{i_2 j}^2 \right) + \log\lambda_j \left( a_j - 1 + \frac{m_1}{2} + \frac{m_2}{2} \right) \right\} + \ldots \\
    &= \sum_{j = 1}^{k} \left\{ \lambda_j \left( -b_j - \frac{1}{2} (U^T U)_{jj} - \frac{1}{2} (V^T V)_{jj} \right) + \log\lambda_j \left( a_j - 1 + \frac{m_1}{2} + \frac{m_2}{2} \right) \right\} + \ldots
\end{align*}
It follows, given the expression of the Gamma distribution as an element of the exponential family, that
\begin{equation*}
    q^*(\Lambda) = \prod_{j = 1}^{k} \mathcal{G}(\lambda_j | c_j, d_j)
\end{equation*}
with
\begin{equation*}
    c_j = a_j + \frac{m_1 + m_2}{2}, \quad d_j = b_j + \frac{1}{2} \E\{ U^T U \}_{jj} + \frac{1}{2} \E\{ V^T V \}_{jj}, \quad j = 1, \ldots, k.
\end{equation*}

\subsubsection{Noise precision $\tau$}
In the same vein, we get
\begin{align*}
    \log p(Y_{\Omega}, \Theta) &= \sum_{\Omega} \log p(y_{i_1 i_2} | U, V, \tau) + \log p(\tau) + \ldots \\
    &= \sum_{\Omega} \left\{ -\frac{\tau}{2} (y_{i_1 i_2} - g_{i_1}^T U V^T h_{i_2} )^2 + \frac{1}{2} \log\tau \right\} + \left\{ -b_0 \tau + (a_0 - 1) \log\tau \right\} + \ldots \\
    &= \tau \left( -b_0 - \frac{1}{2} \sum_{\Omega} \left( y_{i_1 i_2} - g_{i_1}^T U V^T h_{i_2} \right)^2 \right) + \log\tau \left( a_0 - 1 + \frac{|\Omega|}{2} \right) + \ldots
\end{align*}
Hence
\begin{equation*}
    q^*(\tau) = \mathcal{G}(\tau | c_0, d_0)
\end{equation*}
with
\begin{equation*}
    c_0 = a_0 + \frac{|\Omega|}{2}, \quad d_0 = b_0 + \frac{1}{2} \E\Big\{ \| Y_\Omega - \mathcal{P}_{\Omega}(G U V^T H^T) \|_F^2 \Big\}.
\end{equation*}

\subsection{Tensor case}
\subsubsection{Canonical factors $U_l$}
For tensors, it is sufficient to note that
\begin{equation*}
    \langle U_1^T g_{1, i_1}, \ldots, U_d^T g_{d, i_d} \rangle = \overline{u}_l^T (I_k \otimes g_{l, i_l}) \bigodot_{s \neq l} (I_k \otimes g_{s, i_s}^T) \overline{u}_s = \overline{u}_l^T \overline{\varphi}_{i_1 \ldots i_d}.
\end{equation*}
We then immediately arrive at
\begin{align*}
    &\log p(\bm{Y}_{\Omega}, \Theta) = \sum_{\Omega} \log p(y_{i_1 \ldots i_d} | U_1, \ldots,  U_d, \tau) + \sum_{j_l = 1}^{m_l} \log p(u_{l, j_l} | \Lambda) + \ldots \\
    &= -\frac{1}{2} \overline{u}_l^T \left( \Lambda \otimes I_{m_l} + \tau \sum_{\Omega} \overline{\varphi}_{i_1 \ldots i_d} \overline{\varphi}_{i_1 \ldots i_d}^T \right) \overline{u}_l + \overline{u}_l^T \left( \tau \sum_{\Omega} y_{i_1 \ldots i_d} \overline{\varphi}_{i_1 \ldots i_d} \right) + \ldots
\end{align*}
and recognize a Gaussian distribution. On taking the expectation $\E_{\sim U_l}$, we get the covariance matrix $A_l$ of the optimal posterior distribution
\begin{align*}
    A_{l} &= \left[ \E\Big\{\Lambda \otimes I_{m_l} + \tau \sum_{\Omega} \overline{\varphi}_{i_1 \ldots i_d} \overline{\varphi}_{i_1 \ldots i_d}^T \Big\} \right]^{-1} \\
    &= \left[ \E\{\Lambda\} \otimes I_{m_l} + \E\{\tau\} \sum_{\Omega} \E\{ \overline{\varphi}_{i_1 \ldots i_d} \overline{\varphi}_{i_1 \ldots i_d}^T \} \right]^{-1} \\
    &= \left[ \E\{\Lambda\} \otimes I_{m_l} + \E\{\tau\} \sum_{\Omega} (I_k \otimes g_{l, i_l}) \E\left\{ \bigodot_{s \neq l} (I_k \otimes g_{s, i_s}^T) \overline{u}_s \overline{u}_s^T (I_k \otimes g_{s, i_s}) \right\} (I_k \otimes g_{l, i_1}^T) \right]^{-1} \\
    &= \left[ \E\{\Lambda\} \otimes I_{m_l} + \E\{\tau\} \sum_{\Omega} (I_k \otimes g_{l, i_l}) \left( \bigodot_{s \neq l} (I_k \otimes g_{s, i_s}^T) \E\{\overline{u}_s \overline{u}_s^T\} (I_k \otimes g_{s, i_s}) \right) (I_k \otimes g_{l, i_1}^T) \right]^{-1} \\
    &= \left[ \E\{\Lambda\} \otimes I_{m_l} + \E\{\tau\} \sum_{\Omega} \left( \bigodot_{s \neq l} (I_k \otimes g_{s, i_s}^T) \E\{\overline{u}_s \overline{u}_s^T\} (I_k \otimes g_{s, i_s}) \right) \otimes g_{l, i_l} g_{l, i_l}^T \right]^{-1}
\end{align*}
and the corresponding mean
\begin{align*}
    \mu_l &= A_{l} \E\Big\{ \tau \sum_{\Omega} y_{i_1 \ldots i_d} \overline{\varphi}_{i_1 \ldots i_d} \Big\} \\
    &= \E\{\tau\} A_l \sum_{\Omega} y_{i_1 \ldots i_d} (I_k \otimes g_{l, i_l}^T) \E\left\{ \bigodot_{s \neq l} U_s^T g_{s, i_s} \right\} \\
    &= \E\{\tau\} A_l \sum_{\Omega} y_{i_1 \ldots i_d} (I_k \otimes g_{l, i_l}^T) \left( \bigodot_{s \neq l} \E\{U_s^T\} g_{s, i_s} \right) \\
    &= \E\{\tau\} A_l \sum_{\Omega} y_{i_1 \ldots i_d} \left( \bigodot_{s \neq l} \E\{U_s^T\} g_{s, i_s} \right) \otimes g_{l, i_l}.
\end{align*}
In the derivation, we used the mixed-product property of the Hadamard product
\begin{equation*}
    \left[ \bigodot_{s \neq l} (I_k \otimes g_{s, i_s}^T) \overline{u}_s \right] \left[ \bigodot_{s \neq l} (I_k \otimes g_{s, i_s}^T) \overline{u}_s \right]^T = \bigodot_{s \neq l} (I_k \otimes g_{s, i_s}^T) \overline{u}_s \overline{u}_s^T (I_k \otimes g_{s, i_s}),
\end{equation*}
and the factorized form of $q(\Theta)$ allowed us to compute its expectation.

\subsubsection{Precision matrix $\Lambda$}
In analogy with the matrix case, we have
\begin{align*}
    \log p(\bm{Y}_{\Omega}, \Theta) &= \sum_{l = 1}^{d}\sum_{i_l = 1}^{m_l} \log p(u_{l, i_l} | \Lambda) + \sum_{j = 1}^{k} \log p(\lambda_j) + \ldots \\
    &= \sum_{l = 1}^{d} \sum_{i_l = 1}^{m_l} \left\{ -\frac{1}{2} u_{l, i_l}^T \Lambda u_{l, i_l} + \frac{1}{2} \sum_{j = 1}^{k} \log \lambda_j \right\} + \sum_{j = 1}^{k} \left\{ -b_j \lambda_j + (a_j - 1) \log\lambda_j \right\} + \ldots \\
    &= \sum_{j = 1}^{k} \left\{ \lambda_j \left( -b_j - \frac{1}{2} \sum_{l = 1}^{d} \sum_{i_l = 1}^{m_l} u_{l, i_l j}^2 \right) + \log\lambda_j \left( a_j - 1 + \frac{1}{2} \sum_{l = 1}^{d} m_l \right) \right\} + \ldots
\end{align*}
and, as a result,
\begin{equation*}
    q^*(\Lambda) = \prod_{j = 1}^{k} \mathcal{G}(\lambda_j | c_j, d_j)
\end{equation*}
with
\begin{equation*}
    c_j = a_j + \frac{1}{2} \sum_{l = 1}^{d} m_l, \quad d_j = b_j + \frac{1}{2} \sum_{l = 1}^{d} \E\{ U_l^T U_l \}_{jj}, \quad j = 1, \ldots, k.
\end{equation*}

\subsubsection{Noise precision $\tau$}
We repeat the computation yet again to get
\begin{align*}
    &\log p(\bm{Y}_{\Omega}, \Theta) = \sum_{\Omega} \log p(y_{i_1 \ldots i_d} | U_1, \ldots, U_d, \tau) + \log p(\tau) + \ldots \\
    &= \sum_{\Omega} \left\{ -\frac{\tau}{2} \Big(y_{i_1 \ldots i_d} - \langle U_1^T g_{1, i_1}, \ldots, U_d^T g_{d, i_d} \rangle \Big)^2 + \frac{1}{2} \log\tau \right\} + \left\{ -b_0 \tau + (a_0 - 1) \log\tau \right\} + \ldots \\
    &= \tau \left( -b_0 - \frac{1}{2} \sum_{\Omega} \Big( y_{i_1 \ldots i_d} - \langle U_1^T g_{1, i_1}, \ldots, U_d^T g_{d, i_d} \rangle \Big)^2 \right) + \log\tau \left( a_0 - 1 + \frac{|\Omega|}{2} \right) + \ldots
\end{align*}
Consequently,
\begin{equation*}
    q^*(\tau) = \mathcal{G}(\tau | c_0, d_0)
\end{equation*}
with
\begin{equation*}
    c_0 = a_0 + \frac{|\Omega|}{2}, \quad d_0 = b_0 + \frac{1}{2} \E\Big\{ \| \bm{Y}_{\Omega} - \mathcal{P}_{\Omega} [G_1 U_1, \ldots, G_d U_d] \|_F^2 \Big\}.
\end{equation*}

\subsection{Computing the expectations}
To get explicit formulas for the optimal distributions, we need to take the expectations. The only non-trivial ones are related to the rate parameters $d_j$ and $d_0$ of precision matrix $\Lambda$ and noise precision $\tau$, respectively.

First, note that $(U_l^T U_l)_{jj}$ is the squared Euclidean norm of the $j$-th column of $U_l$ or, in other words, of the $j$-th subvector of $\overline{u}_l$ of length $m_l$. The outer product of this subvector with itself is exactly the $j$-th diagonal block of $\overline{u}_l \overline{u}_l^T$ of size $m_l \times m_l$, and its components squared lie on the diagonal of the block; hence the trace.

Second, we use the following property of the multi-linear product
\begin{equation*}
    \langle U_1^T g_{1, i_1}, \ldots, U_d^T g_{d, i_d} \rangle^2 = \langle U_1^T g_{1, i_1} g_{1, i_1}^T U_1, \ldots, U_d^T g_{d, i_d} g_{d, i_d}^T U_d \rangle
\end{equation*}
together with $U_l^T g_{l, i_l} = (I_{m_l} \otimes g_{l, i_l}^T) \overline{u}_l$. The factorized form of $q(\Theta)$ allows us to compute the expectations individually for each matrix in the multi-linear product.
\section{Distribution of predicted values}
Let us show how the Student's $t$-distribution arises when we try to predict the unknown elements of a matrix/tensor based on the given ones $\bm{Y}_{\Omega}$. We have
\begin{align*}
    p(y_{i_1 \ldots i_d} | \bm{Y}_{\Omega}) &= \int p(y_{i_1 \ldots i_d} | U_1, \ldots, U_d, \tau) p(U_1, \ldots, U_d, \Lambda, \tau | \bm{Y}_{\Omega})  \prod_{l = 1}^d dU_l d\Lambda d\tau \\
    &\approx \int p(y_{i_1 \ldots i_d} | U_1, \ldots, U_d, \tau) \prod_{l = 1}^d q^*(U_l) q^*(\tau) \prod_{l = 1}^d dU_l d\tau.
\end{align*}

If one puts a Gaussian prior over the mean of another Gaussian random variable and marginalizes it out, the resulting distribution will be Gaussian as well, whose variance is the sum of two variances:
\begin{equation*}
    \int \mathcal{N}(x | \mu, \tau^{-1}) \mathcal{N}(\mu | m, s^{-1}) d\mu = \mathcal{N}(x | m, \tau^{-1} + s^{-1}). 
\end{equation*}
A similar result can be obtained with the Sherman--Morrison formula for
\begin{equation*}
    \int \mathcal{N}(x | u^T v, \tau^{-1}) \mathcal{N}(u | \mu, A) du = \mathcal{N}(y | \mu^T v, \tau^{-1} + v^T A v). 
\end{equation*}
It then follows that we can marginalize out the contribution of $U_1$:
\begin{align*}
    &p(y_{i_1 \ldots i_d} | \bm{Y}_{\Omega}) \approx \int \mathcal{N}(y_{i_1 \ldots i_d} | \langle U_1^T g_{1, i_1}, \ldots, U_d^T g_{d, i_d} \rangle, \tau^{-1}) \prod_{l = 1}^d \mathcal{N}(\overline{u}_l | \mu_{l}, A_{l}) \mathcal{G}(\tau | c_0, d_0) \prod_{l = 1}^d dU_l d\tau \\
    &= \int \mathcal{N}(y_{i_1 \ldots i_d} | \langle M_1^T g_{1, i_1}, \ldots, U_d^T g_{d, i_d} \rangle, \tau^{-1} + \Tilde{\eta}_1) \prod_{l = 2}^d \mathcal{N}(\overline{u}_l | \mu_{l}, A_{l}) \mathcal{G}(\tau | c_0, d_0) \prod_{l = 2}^d dU_l d\tau,
\end{align*}
where
\begin{equation*}
    \Tilde{\eta}_1 = \left( \bigodot_{s \neq 1} U_s^T g_{s, i_s}\right)^T (I_k \otimes g_{1, i_1}^T) A_{1} (I_k \otimes g_{1, i_1} ) \left( \bigodot_{s \neq 1} U_s^T g_{s, i_s}\right).
\end{equation*}
Now, we cannot apply directly the same idea since both mean and variance depend on $U_2$. So, in order to proceed, we have to make an approximation by replacing all $U_l$ with their mean values:
\begin{equation*}
    \Tilde{\eta}_1 \approx \eta_1 = \left( \bigodot_{s \neq 1} M_s^T g_{s, i_s}\right)^T (I_k \otimes g_{1, i_1}^T) A_{1} (I_k \otimes g_{1, i_1} ) \left( \bigodot_{s \neq 1} M_s^T g_{s, i_s}\right).
\end{equation*}
Repeating this for $l$ from $2$ through $d$, we arrive at
\begin{equation*}
    p(y_{i_1 \ldots i_d} | \bm{Y}_{\Omega}) \approx \int \mathcal{N}(y_{i_1 \ldots i_d} | \langle M_1^T g_{1, i_1}, \ldots, M_d^T g_{d, i_d} \rangle, \tau^{-1} + \eta) \mathcal{G}(\tau | c_0, d_0) d\tau
\end{equation*}
with
\begin{equation*}
    \eta = \sum_{l = 1}^d \eta_l = \sum_{l = 1}^d \left( \bigodot_{s \neq l} M_s^T g_{s, i_s}\right)^T (I_k \otimes g_{l, i_l}^T) A_{l} (I_k \otimes g_{l, i_l} ) \left( \bigodot_{s \neq l} M_s^T g_{s, i_s}\right).
\end{equation*}
The integral is exactly the Student's $t$-distribution
\begin{equation*}
    \text{St}\left(y_{i_1 \ldots i_d} | \langle M_1^T g_{1, i_1}, \ldots, M_d^T g_{d, i_d} \rangle, (d_0 / c_0 + \eta)^{-1}, 2 c_0 \right).
\end{equation*}
\section{Additional numerical experiments with synthetic data}
\begin{figure}[h]
\begin{subfigure}[b]{0.5\linewidth}
\centering
	\includegraphics[width=0.7\linewidth]{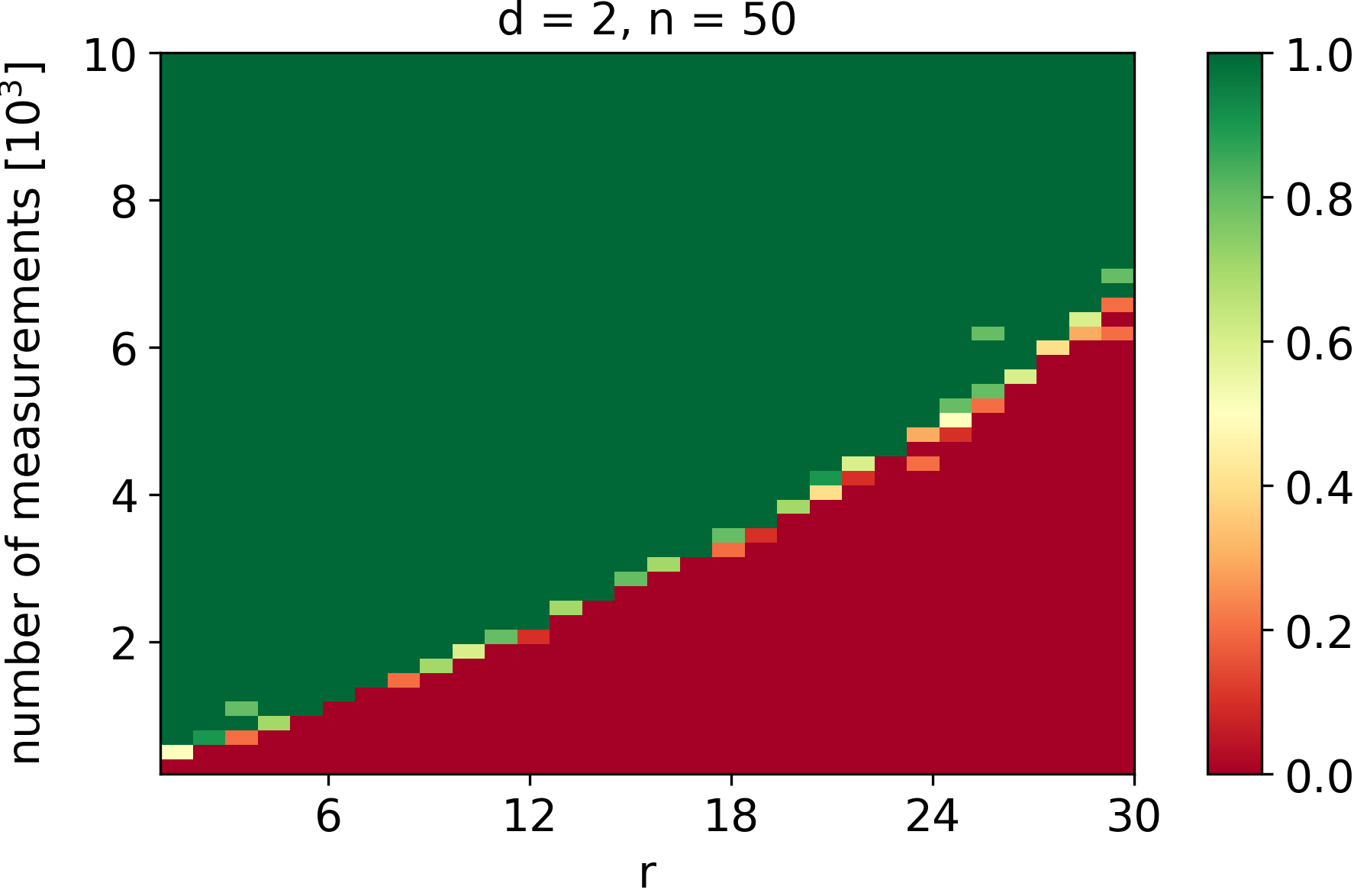}
	\caption{}
\end{subfigure}\hfill%
\begin{subfigure}[b]{0.5\linewidth}
\centering
	\includegraphics[width=0.7\linewidth]{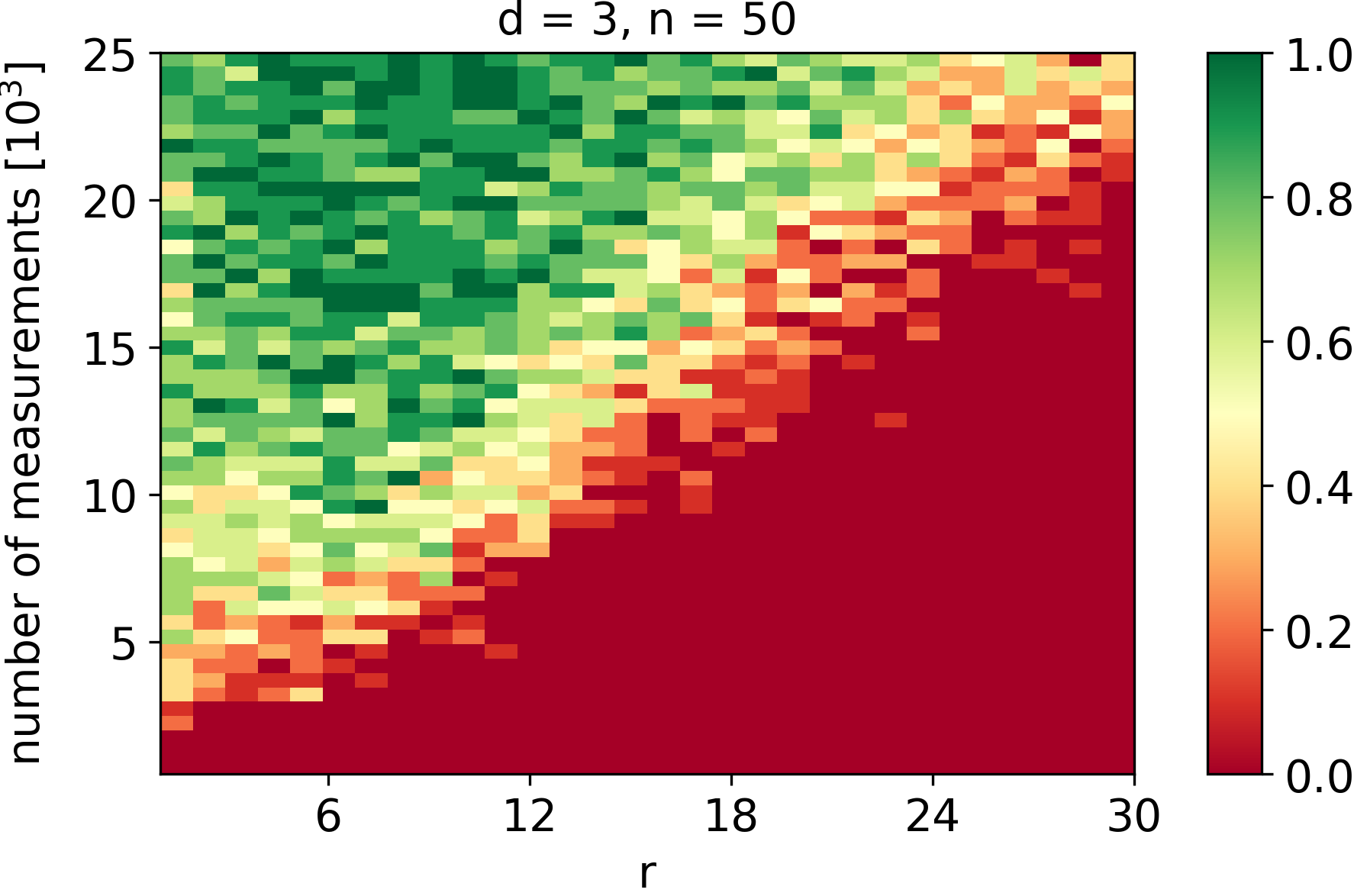}
	\caption{}
\end{subfigure}
\caption{Phase plots for noiseless CP completion with size $n = 50$, varying rank $r$, perfect rank prediction $k = r$, and no side information for different orders of the tensor: $d = 2$~(a) and $d = 3$~(b).}
\label{num:fig:no_varr}
\end{figure}

\begin{figure}[h]
\begin{subfigure}[b]{0.5\linewidth}
\centering
	\includegraphics[width=0.7\linewidth]{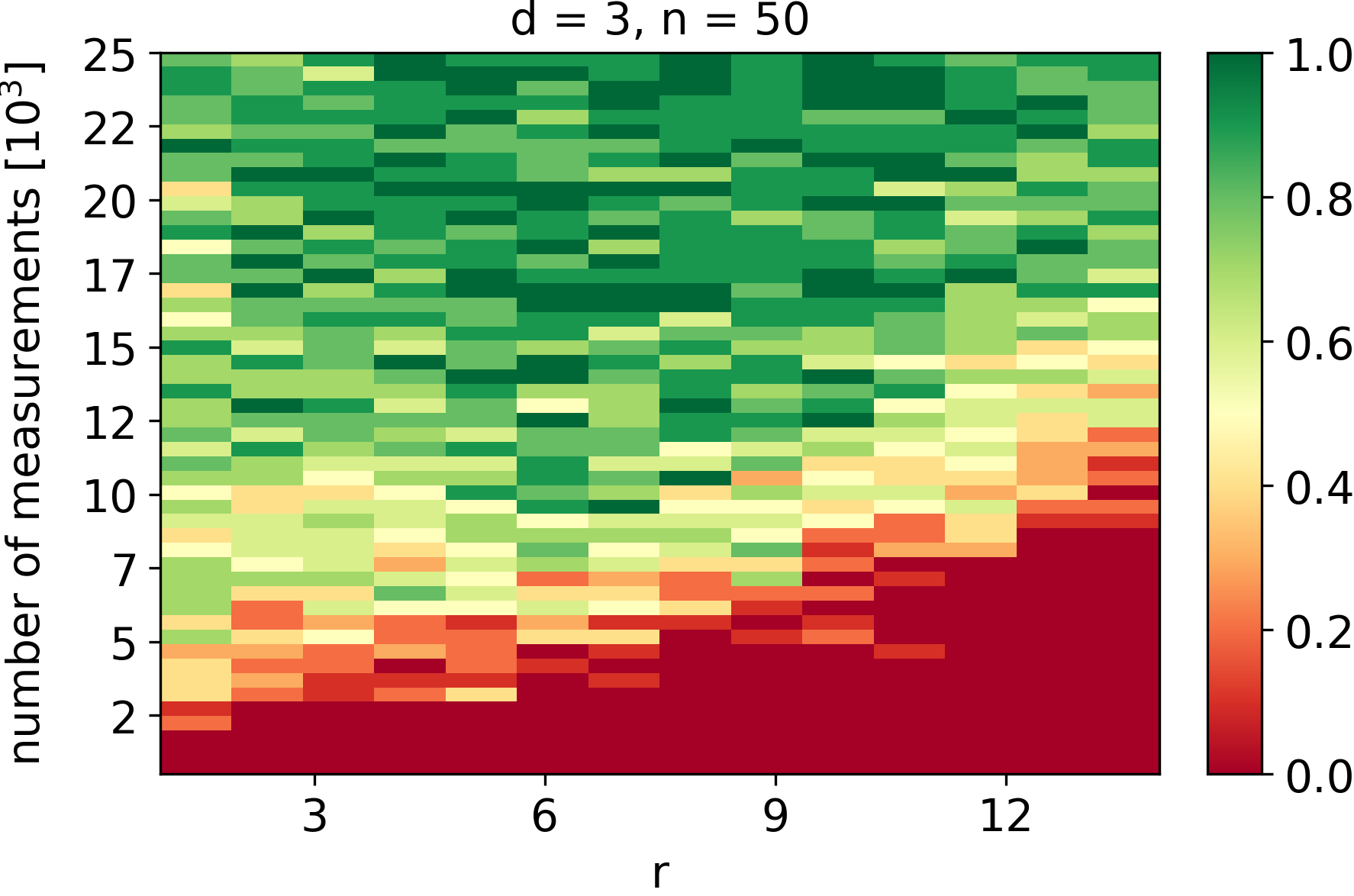}
	\caption{}
\end{subfigure}\hfill%
\begin{subfigure}[b]{0.5\linewidth}
\centering
	\includegraphics[width=0.7\linewidth]{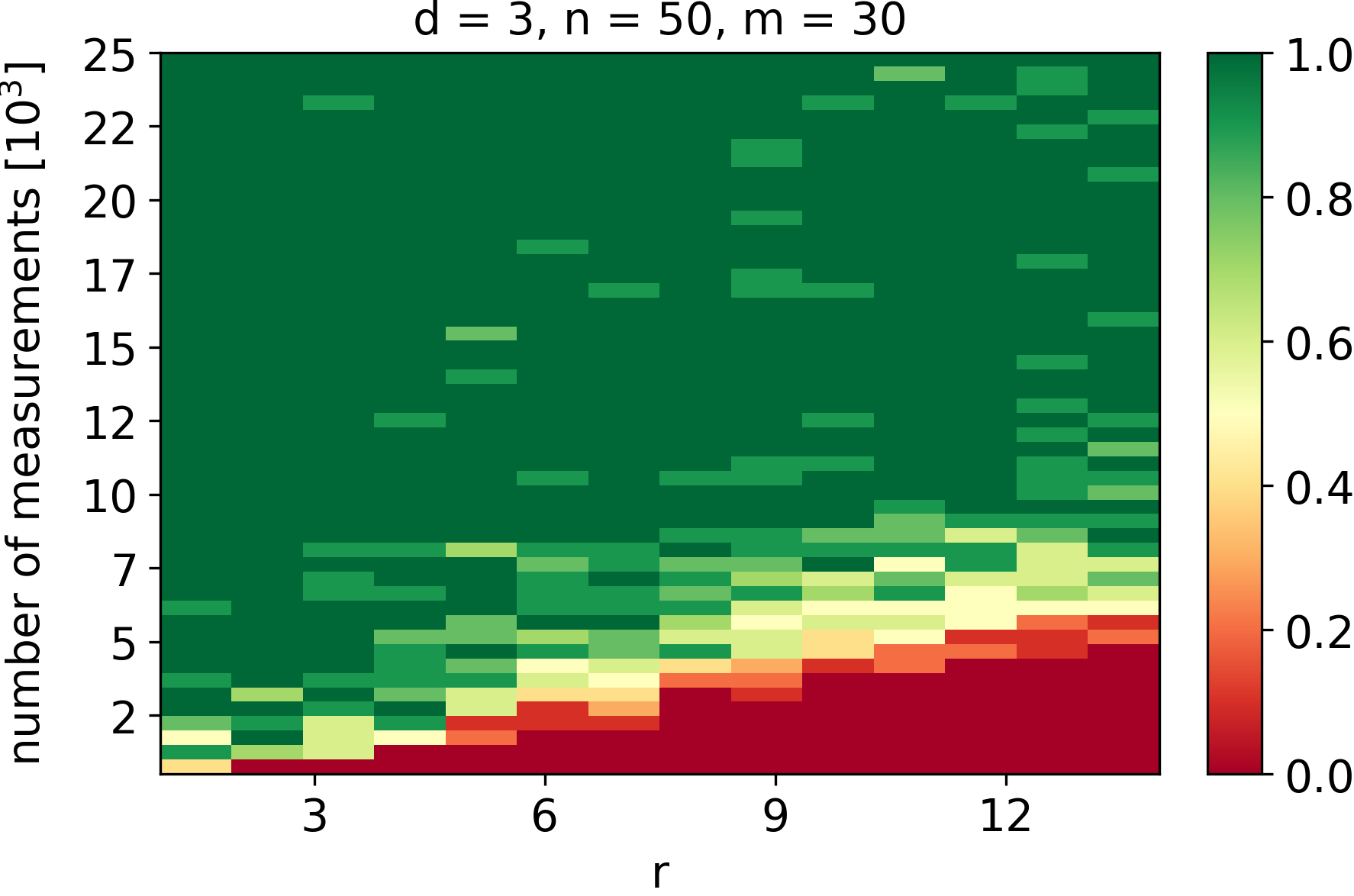}
	\caption{}
\end{subfigure}
\caption{Phase plots for noiseless 3-dimensional CP completion with size $n = 50$, varying rank $r$, and perfect rank prediction $k = r$: FBCP~(a) and FBCP-SI with $m=30$~(b).}
\label{num:fig:no_si_varr}
\end{figure}

Phase plots can be built for fixed $n$ and varying $r$ too. In Fig~\ref{num:fig:no_varr} we compare the phase transitions of matrices and 3-dimensional tensors without side information obtained with FBCP (we made $N_{trial} = 5$ trials with $N_{ic} = 2$ different initial conditions, making $N_{iter} = 100$ iterations for $d = 2$ and $N_{iter} = 150$ iterations for $d = 3$.). Fig.~\ref{num:fig:no_si_varr} shows the regularization effects of side information for $d = 3$, $n = 50$, $m = 30$: the phase transition curve becomes lower, and the successes become more consistent above the curve.

Next, we compare the convergence rates of FBCP and FBCP-SI on the test samples $\Omega^{(t)}_{test}$ for rank-3 CP tensors of size $400 \times 400 \times 400$. In both cases, there are two phases of convergence: a plateau of nearly constant error followed by its linear decay. The more elements of the tensor are known, the shorter the plateau, and side information decreases its length even further. However, the length also depends on the initialization; for instance, we observe that the iterations can converge with one random initialization but not with the other. The rate of linear convergence differs for FBCP and FBCP-SI: the former takes 5-10 iterations to drop the relative error below $10^{-6}$ while the latter requires 30-40 iterations (for $m = 30$).
\begin{figure}[!h]
\begin{subfigure}[b]{0.5\linewidth}
\centering
	\includegraphics[width=0.7\linewidth]{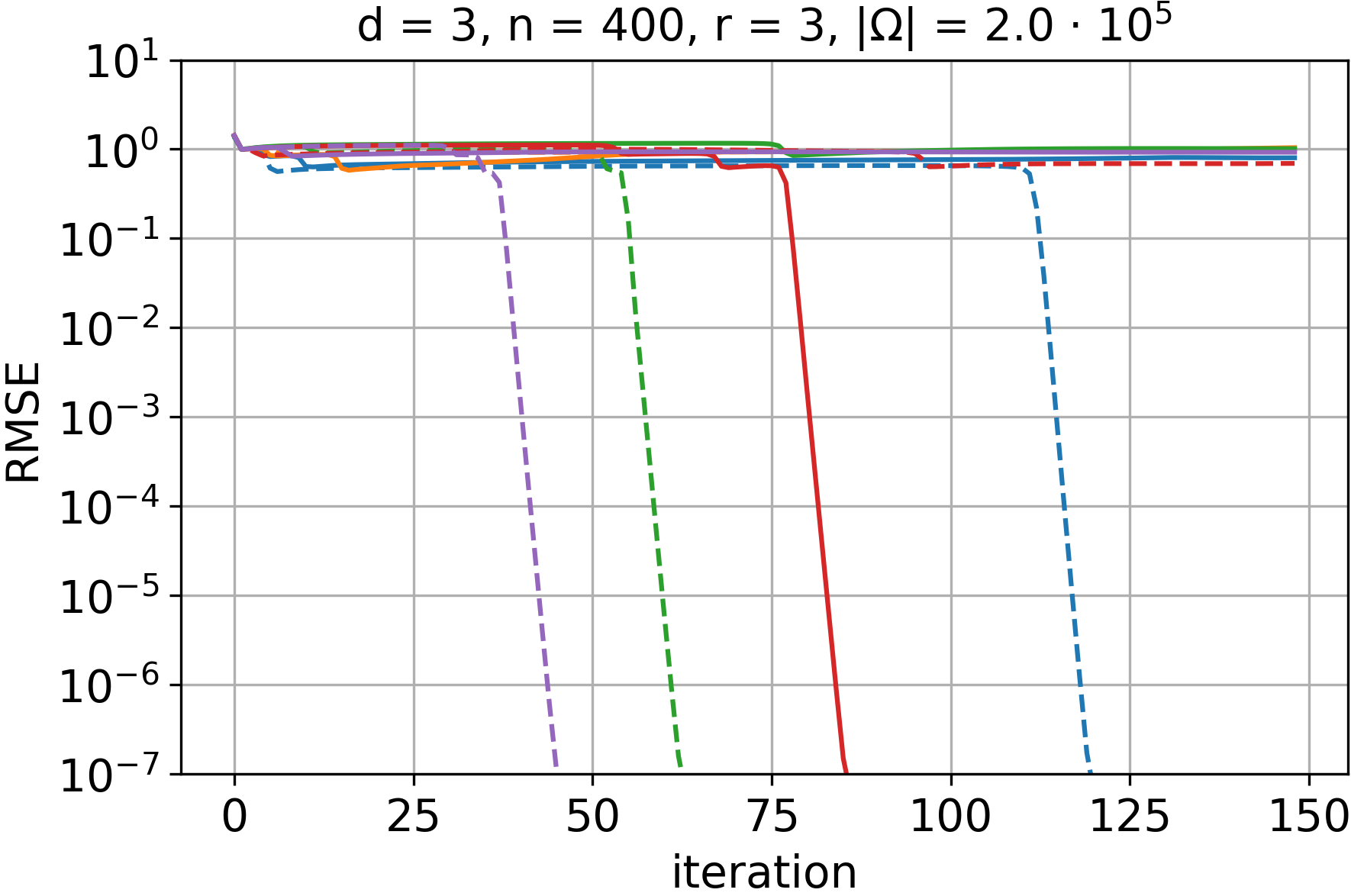}
	\caption{}
\end{subfigure}\hfill%
\begin{subfigure}[b]{0.5\linewidth}
\centering
	\includegraphics[width=0.7\linewidth]{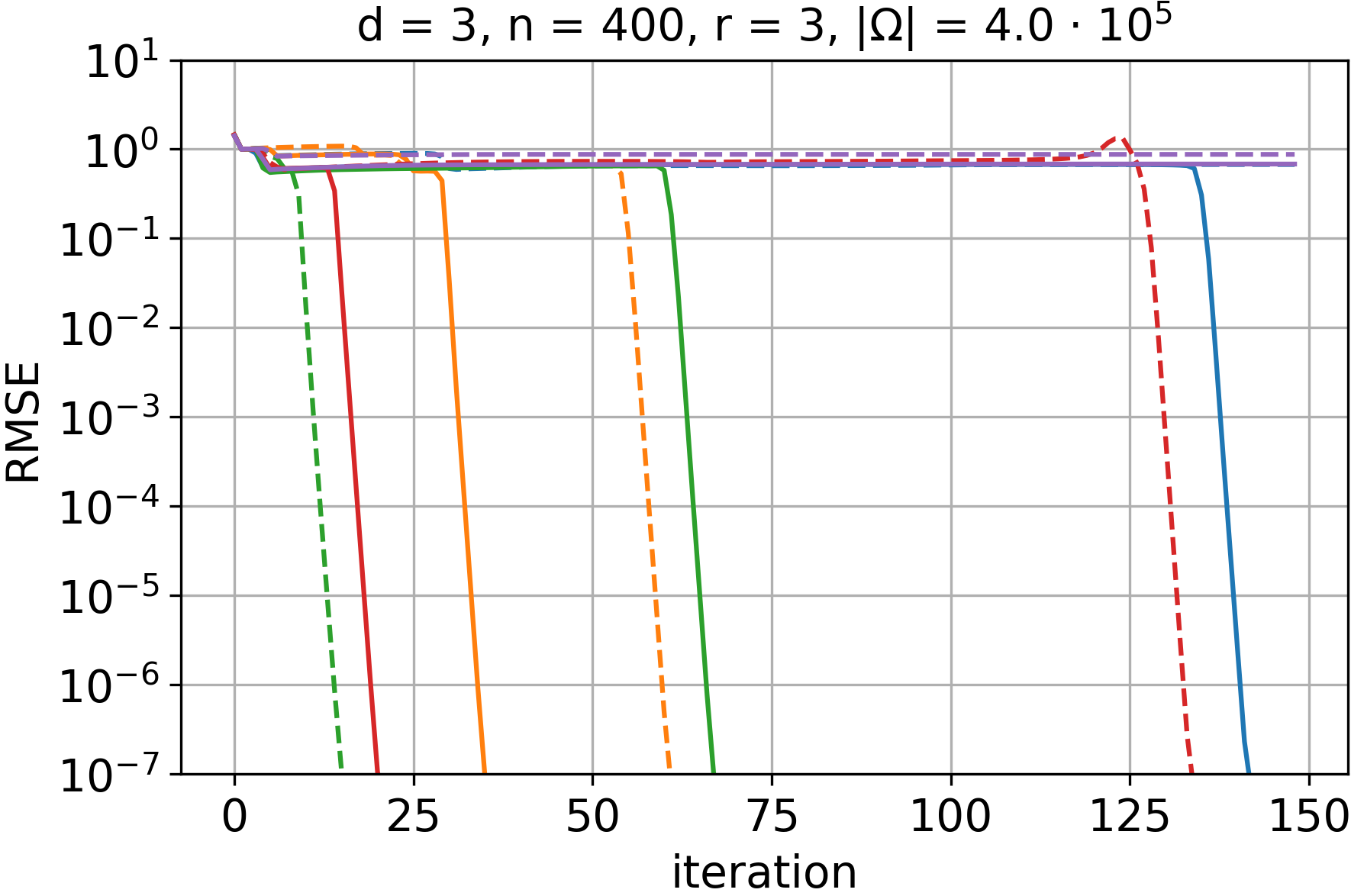}
	\caption{}
\end{subfigure}
\begin{subfigure}[b]{0.5\linewidth}
\centering
	\includegraphics[width=0.7\linewidth]{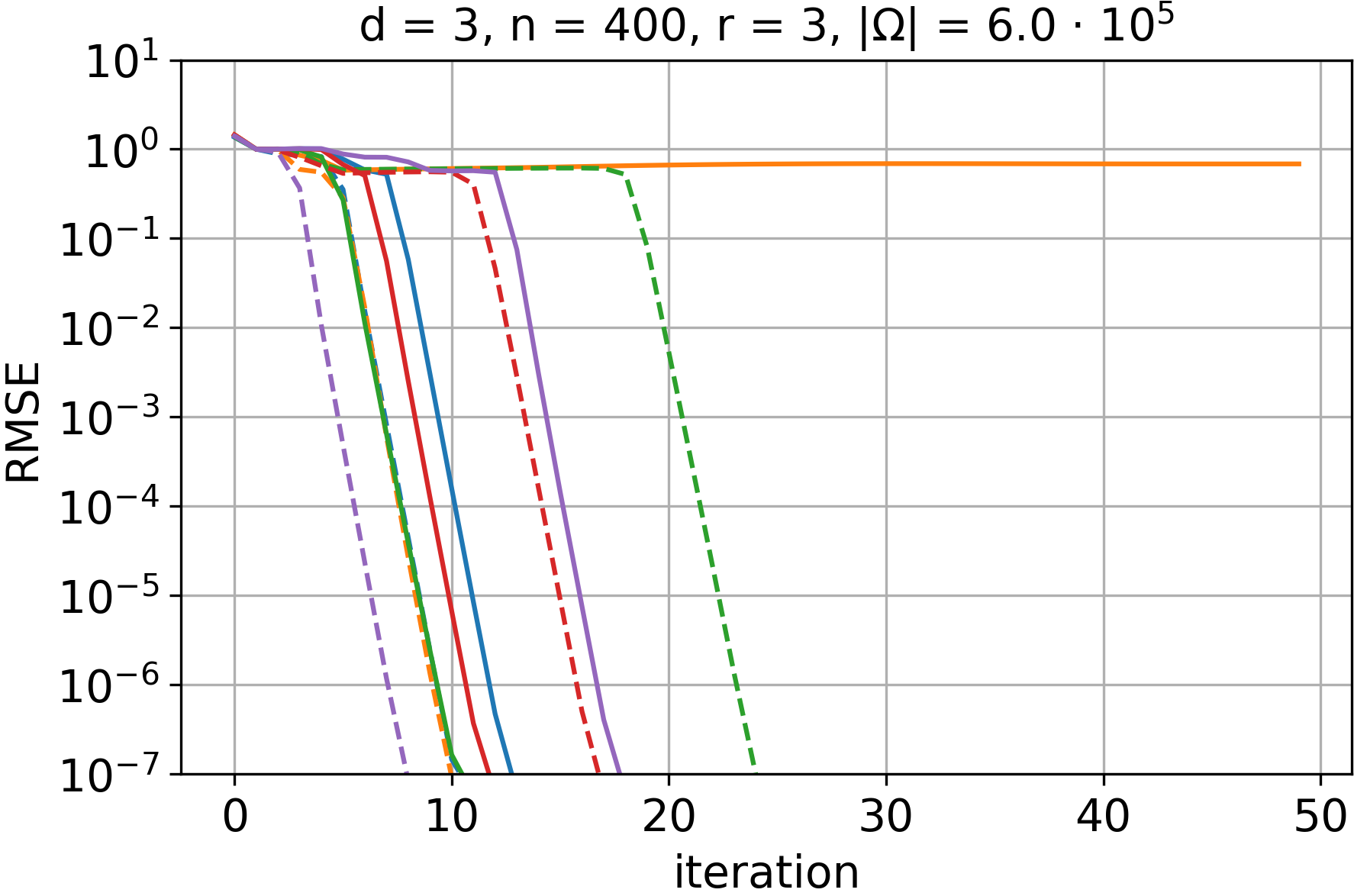}
	\caption{}
\end{subfigure}\hfill%
\begin{subfigure}[b]{0.5\linewidth}
\centering
	\includegraphics[width=0.7\linewidth]{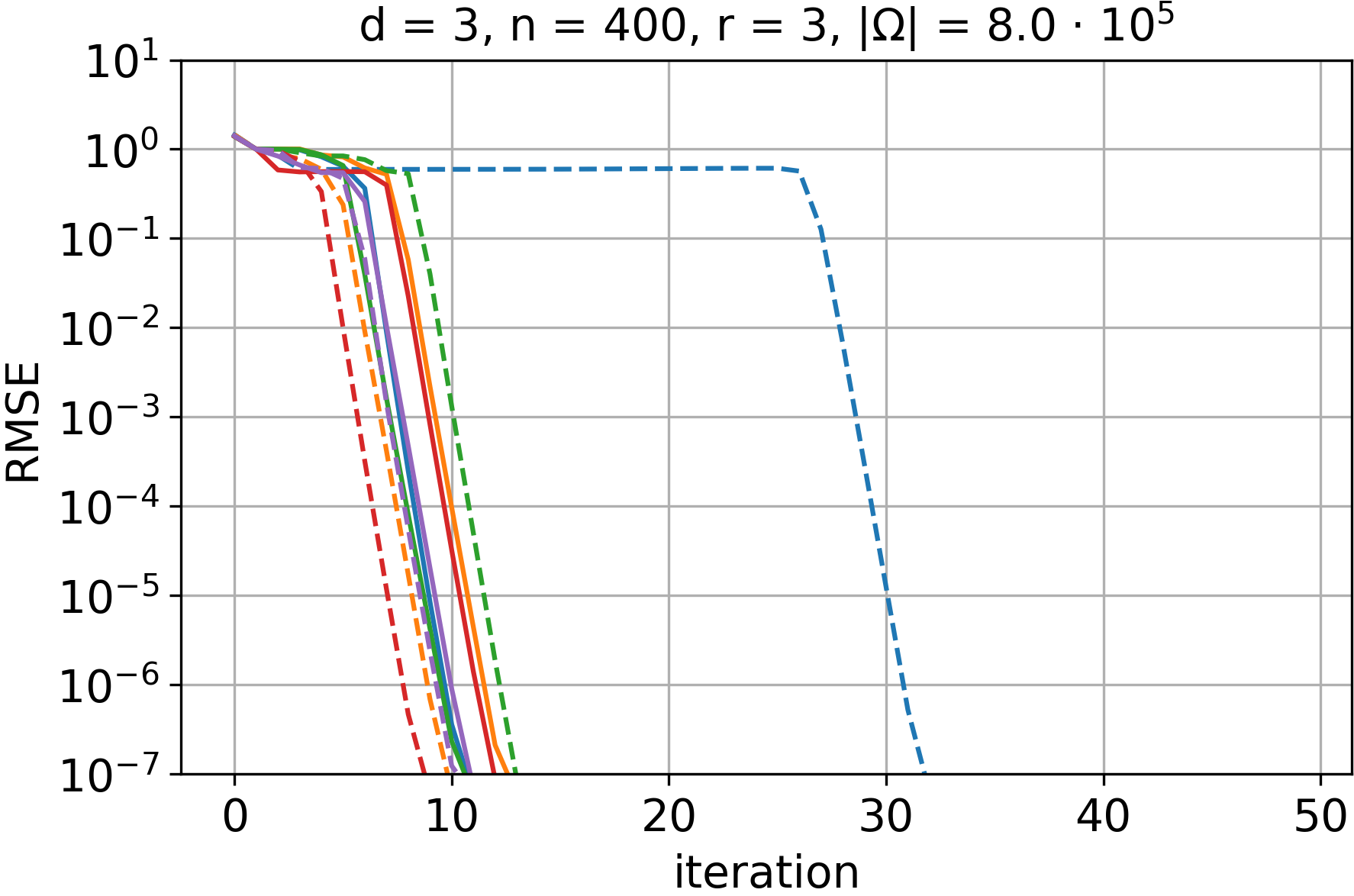}
	\caption{}
\end{subfigure}
\caption{Convergence curves for noiseless 3-dimensional CP completion with size $n = 400$, rank $r = 3$, perfect rank prediction $k = 3$, no side information, and different numbers of samples: $|\Omega| = 2 \cdot 10^5$~(a), $|\Omega| = 4 \cdot 10^5$~(b), $|\Omega| = 6 \cdot 10^5$~(c), and $|\Omega| = 8 \cdot 10^5$~(d). Solid and dashed lines of the same color correspond to 2 different initial conditions.}
\label{num:fig:no_conv_d3}
\end{figure}
\begin{figure}[h]
\begin{subfigure}[b]{0.5\linewidth}
\centering
	\includegraphics[width=0.7\linewidth]{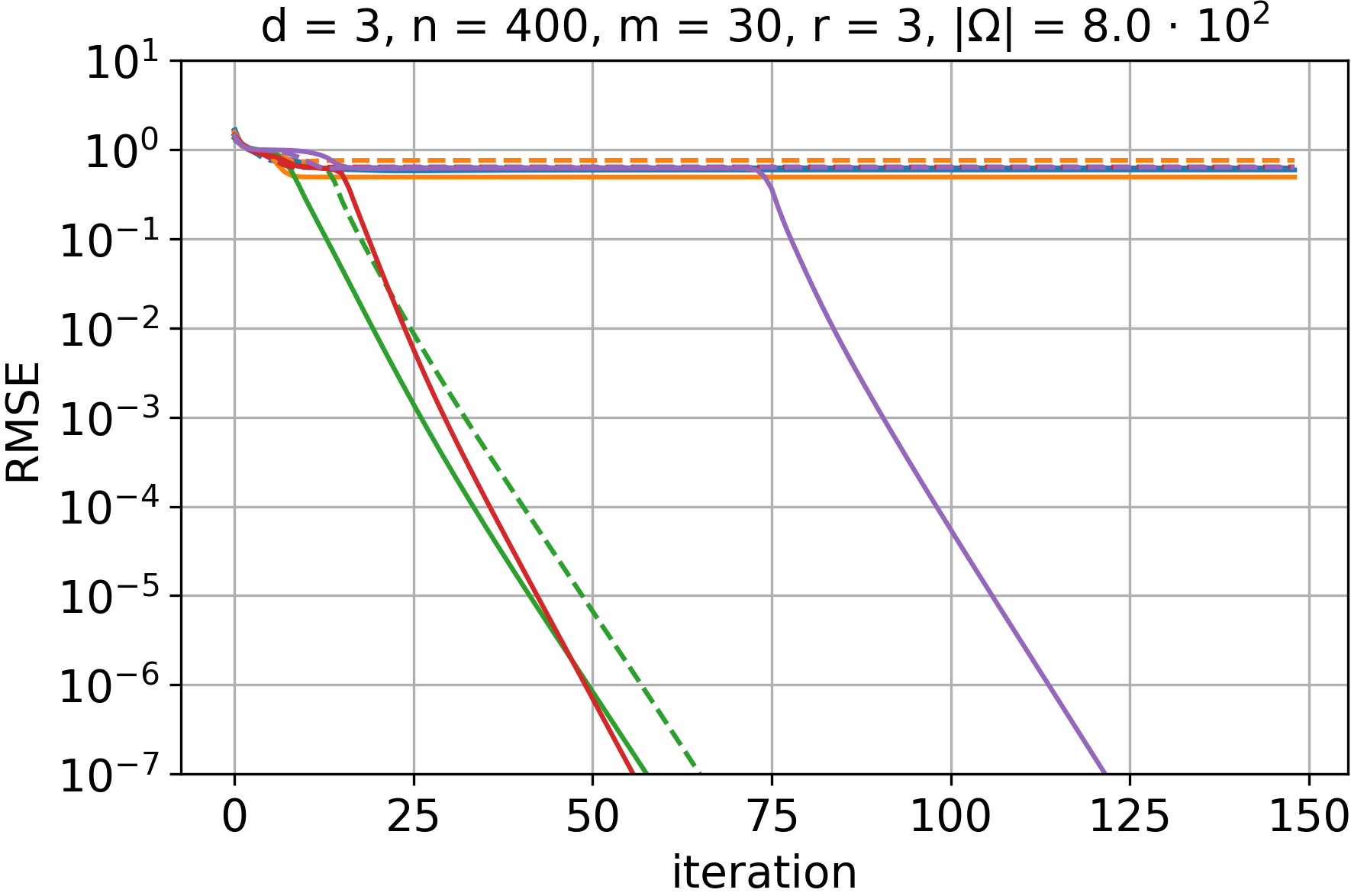}
	\caption{}
\end{subfigure}\hfill%
\begin{subfigure}[b]{0.5\linewidth}
\centering
	\includegraphics[width=0.7\linewidth]{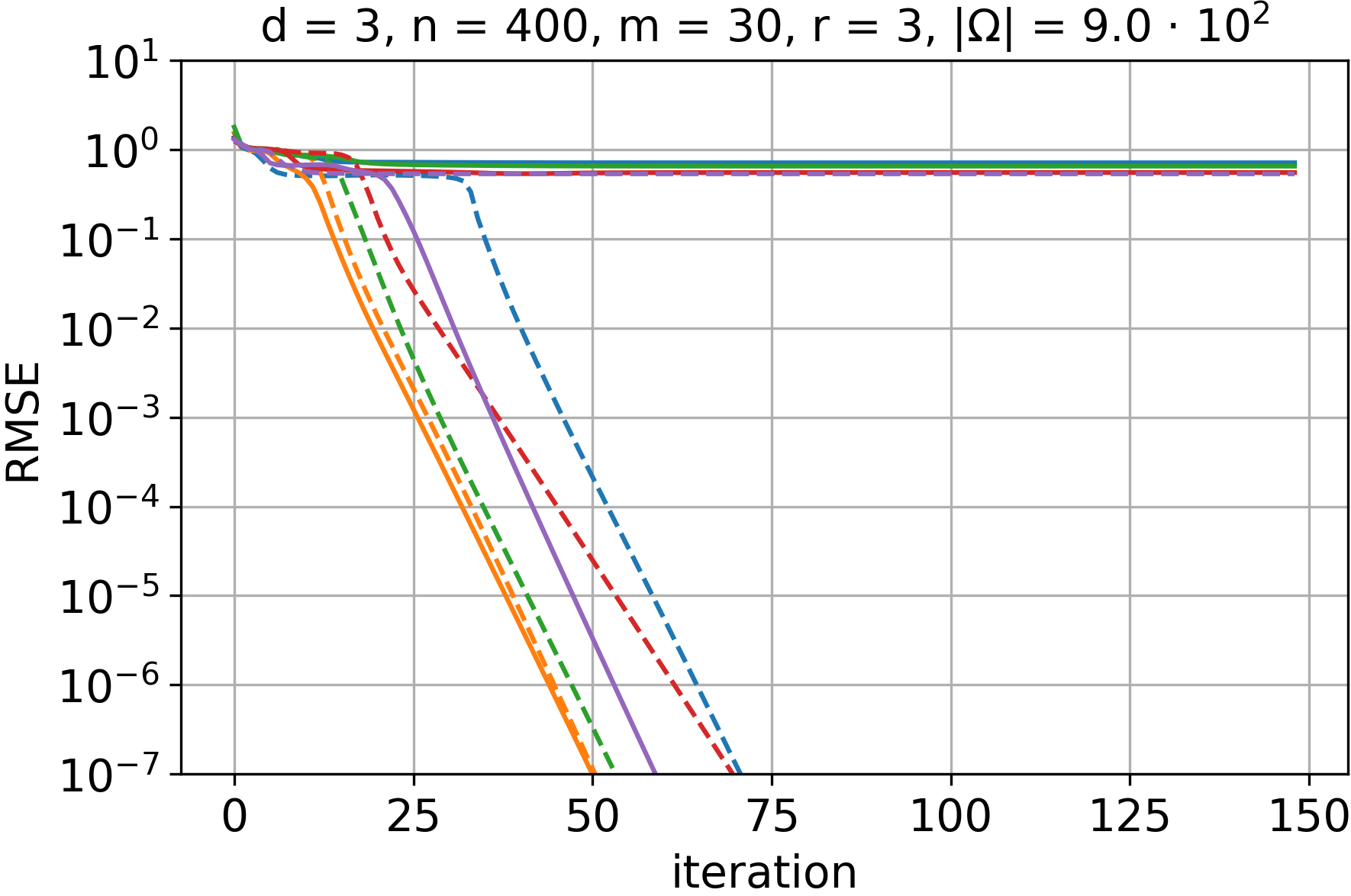}
	\caption{}
\end{subfigure}
\begin{subfigure}[b]{0.5\linewidth}
\centering
	\includegraphics[width=0.7\linewidth]{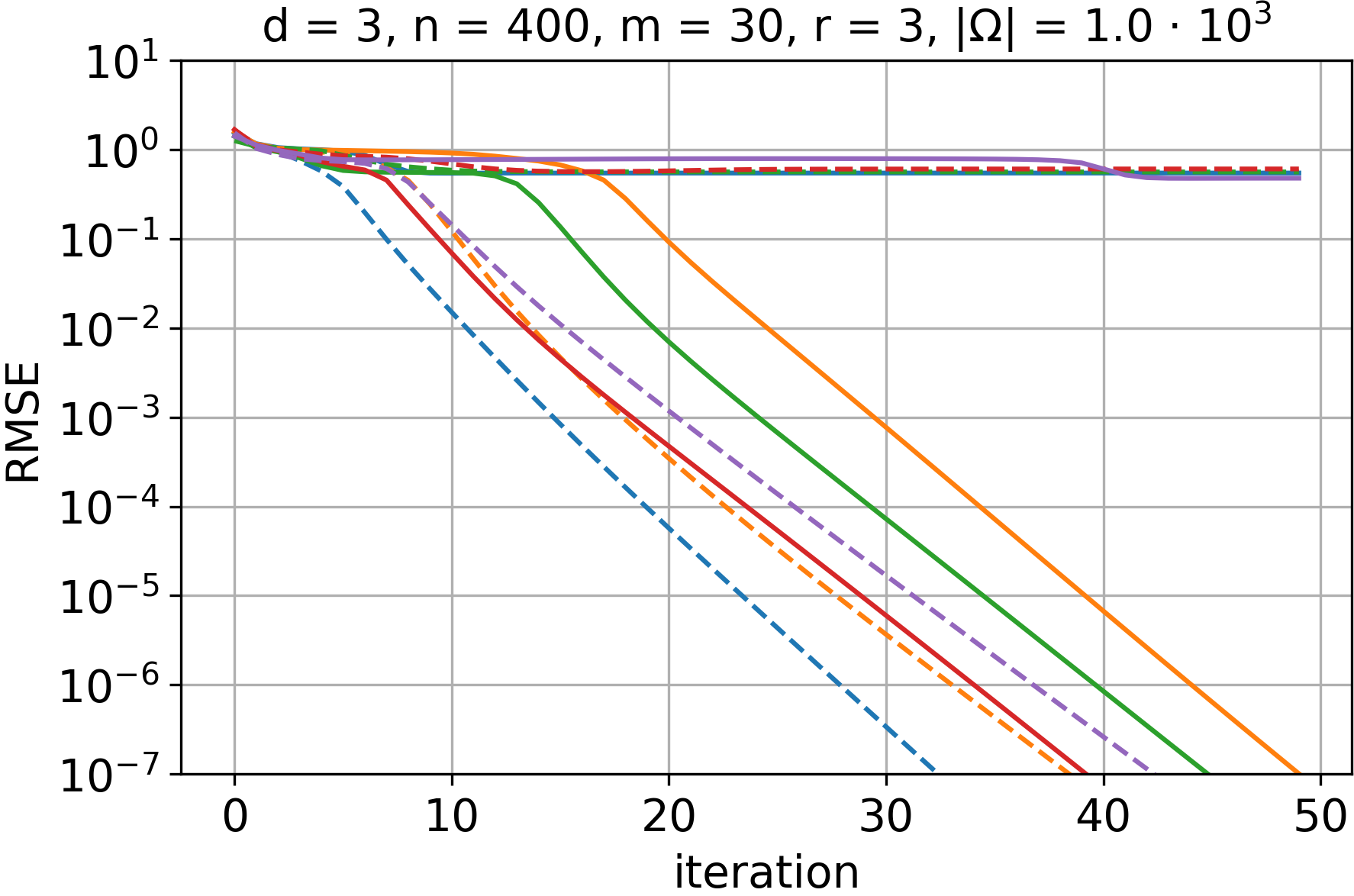}
	\caption{}
\end{subfigure}\hfill%
\begin{subfigure}[b]{0.5\linewidth}
\centering
	\includegraphics[width=0.7\linewidth]{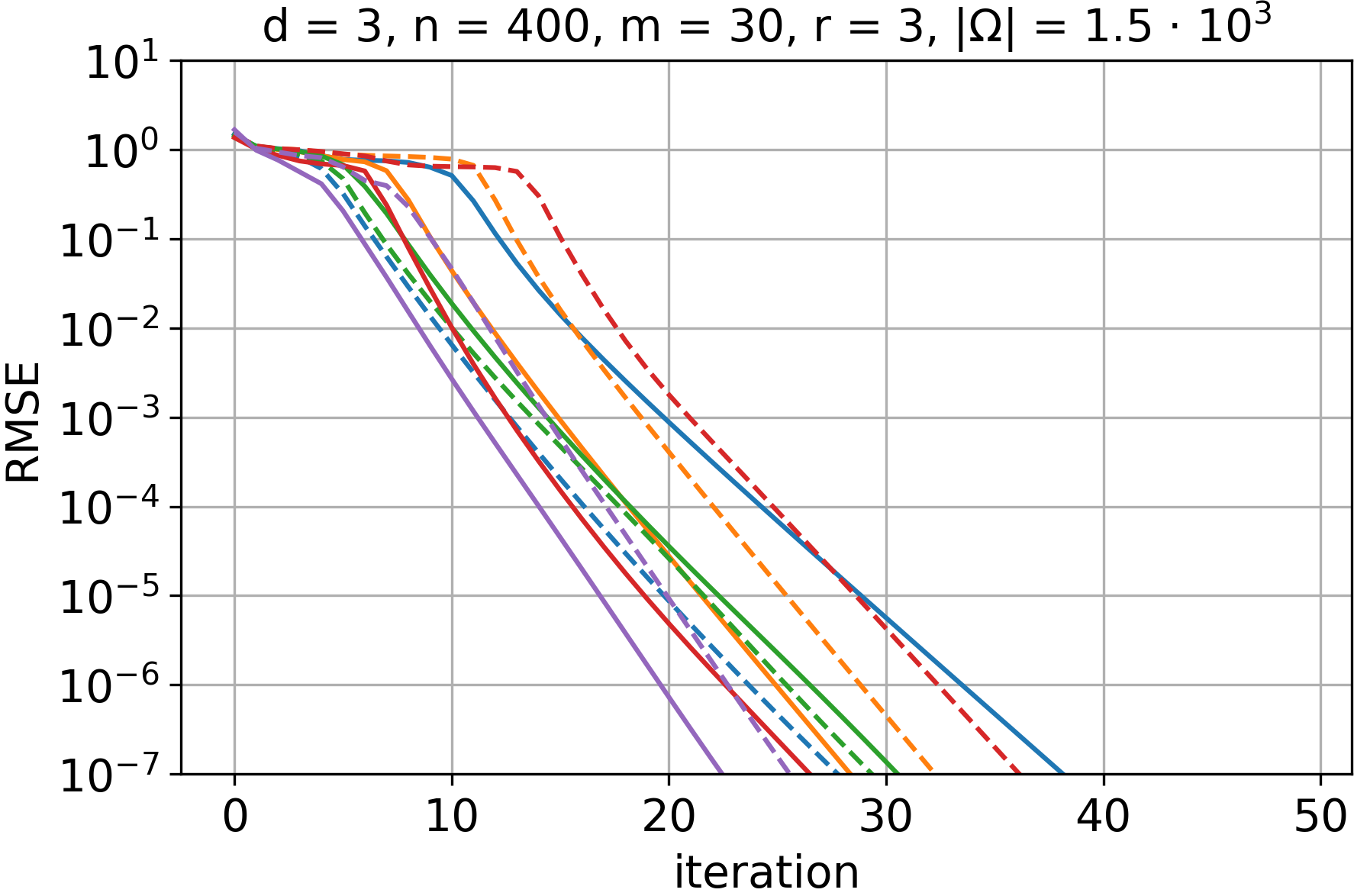}
	\caption{}
\end{subfigure}
\caption{Convergence curves for noiseless 3-dimensional CP completion with size $n = 400$, rank $r = 3$, perfect rank prediction $k = 3$, side information size $m = 30$, and different numbers of samples: $|\Omega| = 8 \cdot 10^2$~(a), $|\Omega| = 9 \cdot 10^2$~(b), $|\Omega| = 1 \cdot 10^3$~(c), and $|\Omega| = 1.5 \cdot 10^3$~(d). Solid and dashed lines of the same color correspond to 2 different initial conditions.}
\label{num:fig:si_conv_d3}
\end{figure}

\begin{figure}[h!]
\begin{subfigure}[b]{0.5\linewidth}
\centering
	\includegraphics[width=0.7\linewidth]{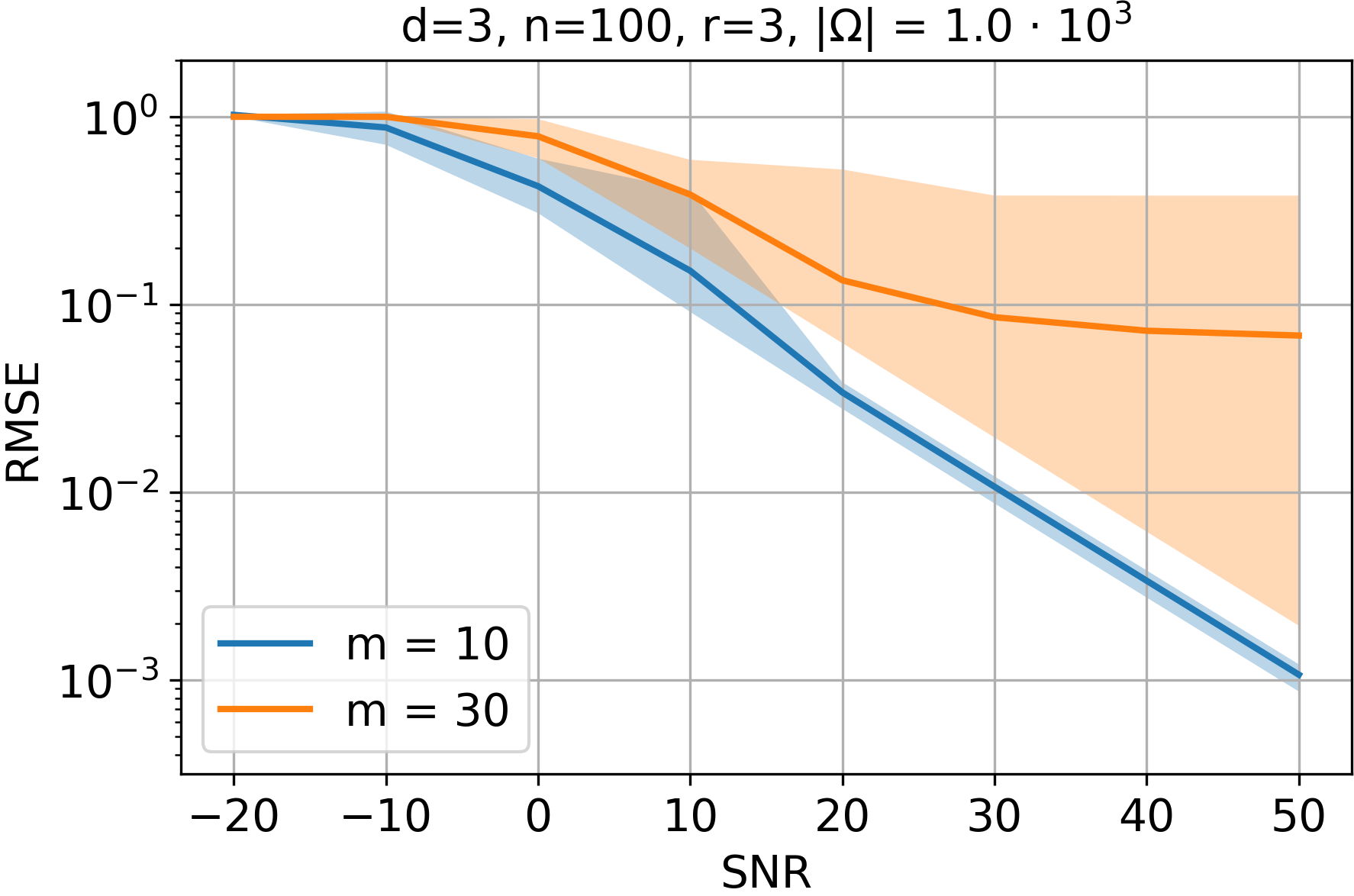}
	\caption{}
\end{subfigure}\hfill%
\begin{subfigure}[b]{0.5\linewidth}
\centering
	\includegraphics[width=0.7\linewidth]{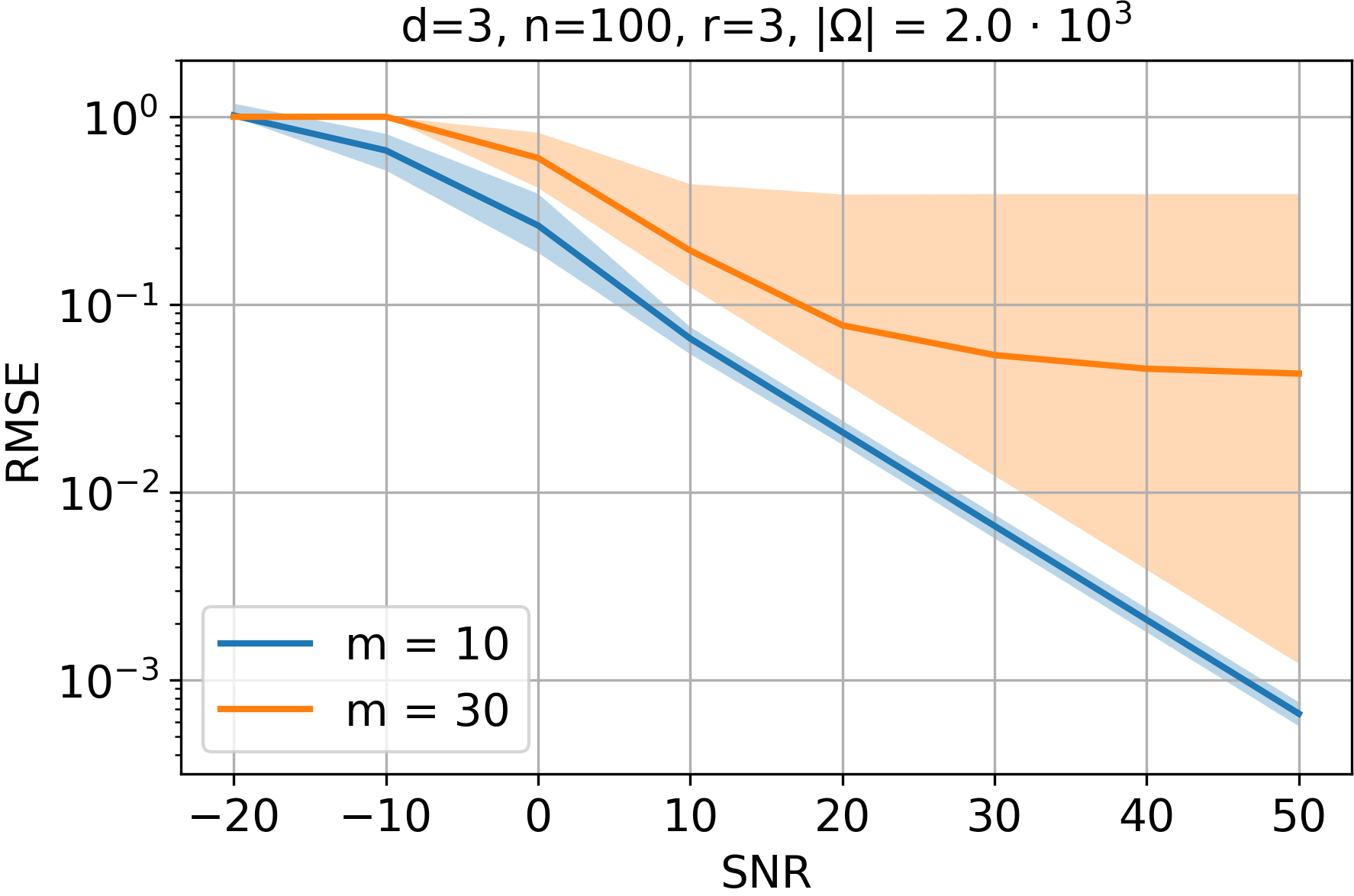}
	\caption{}
\end{subfigure}
\begin{subfigure}[b]{0.5\linewidth}
\centering
	\includegraphics[width=0.7\linewidth]{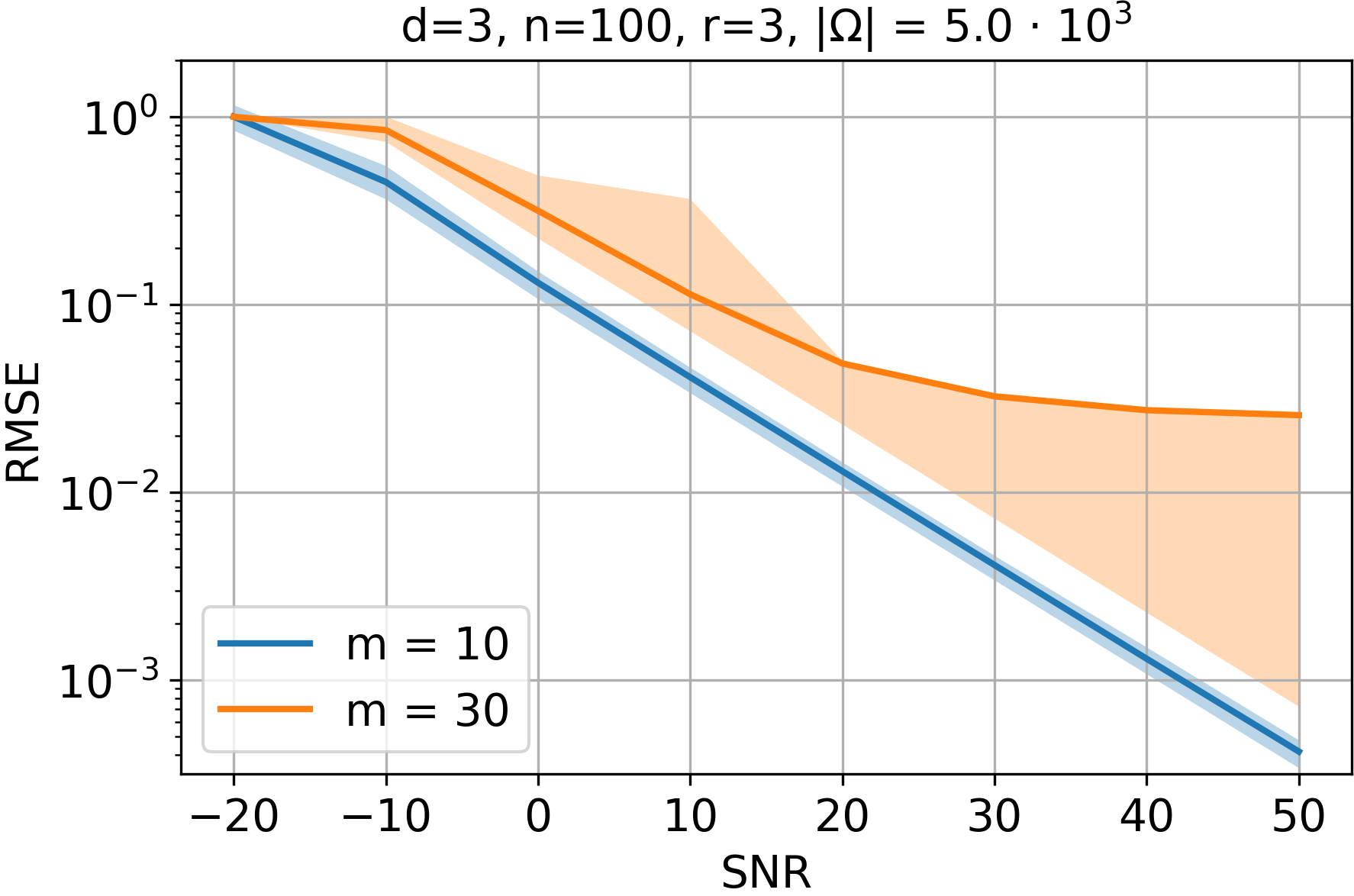}
	\caption{}
\end{subfigure}\hfill%
\begin{subfigure}[b]{0.5\linewidth}
\centering
	\includegraphics[width=0.7\linewidth]{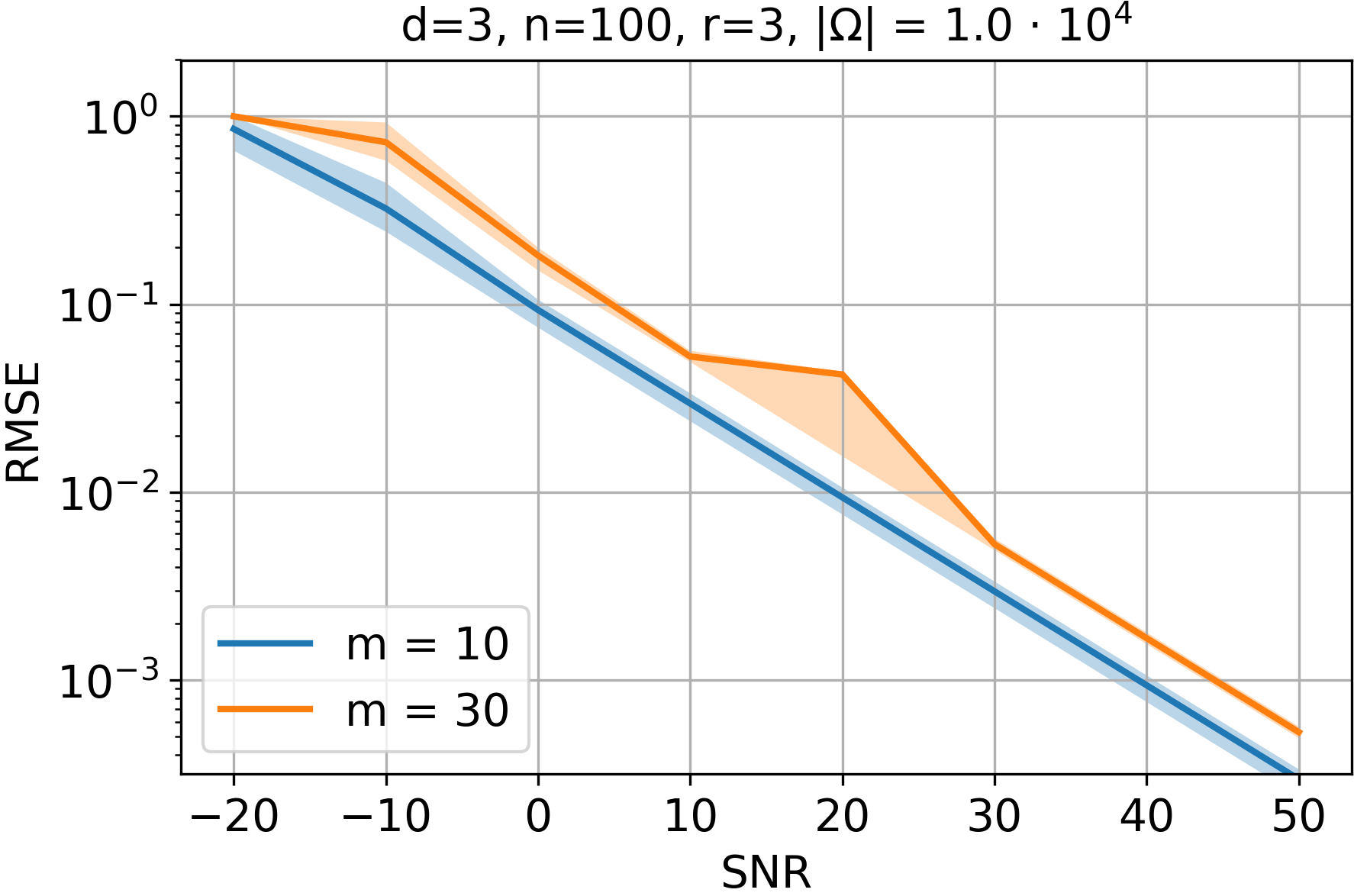}
	\caption{}
\end{subfigure}
\caption{Attainable RMSE for 3-dimensional CP completion with size $n = 100$, rank $r = 3$, perfect rank prediction $k = 3$, varying levels of noise and different numbers of samples: $|\Omega| = 1 \cdot 10^3$~(a), $|\Omega| = 2 \cdot 10^3$~(b), $|\Omega| = 5 \cdot 10^3$~(c), and $|\Omega| = 1 \cdot 10^4$~(d). The curves show the averaged RMSE together with the 5th and 95th percentiles for side information sizes $m = 10$ and $m = 30$.}
\label{num:fig:si_d3_snr}
\end{figure}

Another important aspect is how FBCP-SI performs in the presence of noise. In Fig.~\ref{num:fig:si_d3_snr} we present the results of experiments with random rank-3 CP tensors of sizes $100 \times 100 \times 100$ for different values of $|\Omega|$ and $m$ (the rank is assumed to be known). For varying levels of noise, we plot the RMSE on the test samples after $N_{iter} = 100$ iterations, averaged over $N_{trial} = 20$ trials with $N_{ic} = 1$. The results show that the error is proportional to the standard deviation of additive white Gaussian noise and that smaller $m$ (i.e. more informative side information) leads to lower errors. Notably, the signals are recovered from noise as high as -10dB.

\end{document}